\documentclass[pdflatex, iicol,sn-mathphys-num]{sn-jnl}

\usepackage{graphicx}%
\usepackage{multirow}%
\usepackage{amsmath,amssymb,amsfonts}%
\usepackage{amsthm}%
\usepackage{mathrsfs}%
\usepackage[title]{appendix}%
\usepackage{xcolor}%
\usepackage{textcomp}%
\usepackage{manyfoot}%
\usepackage{booktabs}%
\usepackage{algorithm}%
\usepackage{algorithmicx}%
\usepackage{algpseudocode}%
\usepackage{listings}%

\usepackage{microtype}%
\usepackage{subcaption}%
\usepackage{hyperref}%
\usepackage{mathtools}%
\usepackage[capitalize,noabbrev]{cleveref}%
\usepackage{longtable}%
\usepackage{stfloats}%
\usepackage{tcolorbox}%
\usepackage{todonotes}
\usepackage[cachedir=minted-cache]{minted}
\setminted{
    frame=lines,
    framesep=2mm,
    baselinestretch=1.2,
    fontsize=\tiny,
    breaklines=true,
    breakanywhere=true,
    breaksymbol={},
    breaksymbolleft={}
}

\makeatletter
\renewcommand\scriptsize{\@setfontsize\scriptsize{7}{8}}
\makeatother

\theoremstyle{thmstyleone}%
\theoremstyle{thmstyletwo}%

\theoremstyle{thmstylethree}%

\begin{document}

\title[Atom-anchored LLMs speak Chemistry]{Atom-anchored LLMs speak Chemistry: A Retrosynthesis Demonstration}


\author*[1,2]{\fnm{Alan Kai} \sur{Hassen}}
\equalcont{These authors contributed equally to this work.}

\author[2,3]{\fnm{Andrius} \sur{Bernatavicius}}
\equalcont{These authors contributed equally to this work.}

\author[4]{\fnm{Antonius P. A.} \sur{Janssen}}

\author[2]{\fnm{Mike} \sur{Preuss}}

\author[3]{\fnm{Gerard J. P.} \spfx{van} \sur{Westen}}

\author[1]{\fnm{Djork-Arn\'{e}} \sur{Clevert}}

\affil*[1]{\orgdiv{Machine Learning Research}, \orgname{Pfizer Research and Development}, \orgaddress{\city{Berlin}, \country{Germany}}}

\affil[2]{\orgdiv{Leiden Institute of Advanced Computer Science}, \orgname{Leiden University}, \orgaddress{\city{Leiden}, \country{The Netherlands}}}

\affil[3]{\orgdiv{Leiden Academic Centre for Drug Research}, \orgname{Leiden University}, \orgaddress{\city{Leiden}, \country{The Netherlands}}}

\affil[4]{\orgdiv{Leiden Institute of Chemistry}, \orgname{Leiden University}, \orgaddress{\city{Leiden}, \country{The Netherlands}}}


\abstract{Applications of machine learning in chemistry are often limited by the scarcity and expense of labeled data, restricting traditional supervised methods. In this work, we introduce a framework for molecular reasoning using general-purpose Large Language Models (LLMs) that operates without requiring task-specific model training. Our method anchors chain-of-thought reasoning to the molecular structure by using unique atomic identifiers. First, the LLM performs a zero-shot task to identify relevant fragments and their associated chemical labels or transformation classes. In an optional second step, this position-aware information is used in a few-shot task with provided class examples to predict the chemical transformation.
We apply our framework to single-step retrosynthesis, a task where LLMs have previously underperformed. 
Across academic benchmarks and expert-validated drug discovery molecules, our work enables LLMs to achieve high success rates in identifying chemically plausible reaction sites ($\geq90\%$), named reaction classes ($\geq40\%$), and final reactants ($\geq74\%$).
Ultimately, our work establishes a general blueprint for applying LLMs to challenges where molecular reasoning and molecular transformations are key, positioning atom-anchored LLMs as a powerful solution for data-scarce chemistry domains.}






\maketitle

\section{Introduction}\label{sec1}
General-purpose large language models (LLMs) have advanced rapidly in recent years, finding increasing application in the domain of chemistry. A prominent example of this trend is the use of LLMs like GPT-4 \cite{openaiGPT4TechnicalReport2024} as high-level reasoning agents that leverage specialized chemistry tools to automate complex tasks \cite{boikoAutonomousChemicalResearch2023, m.branAugmentingLargeLanguage2024}. In this paradigm, the LLM orchestrates tool calls that encapsulate chemical logic and subsequently reasons over the tool outputs.
\\
Beyond the use of general-purpose models, prevailing approaches either train specialized chemistry LLMs or adapt general-purpose LLMs to the chemical domain, where molecular data is represented in the Simplified Molecular Input Line Entry System (SMILES) format \cite{weiningerSMILESChemicalLanguage1988, weiningerSMILES2Algorithm1989}, a chemical notation for representing chemical graph structures as computer-readable strings. Examples of specialized chemistry LLMs include models that are solely pre-trained on SMILES data and then either fine-tuned for a specific downstream task (e.g., \cite{rossLargeScaleChemicalLanguage2022a, irwinChemformerPretrainedTransformer2022a}) or used to extract molecular embeddings for downstream tasks (e.g., \cite{rossLargeScaleChemicalLanguage2022a, sadeghiCanLargeLanguage2024, masoodMolecularPropertyPrediction2025}).
Alternatively, general-purpose LLMs are adapted to the chemical domain through methods such as supervised fine-tuning (SFT) \cite{kimLargeLanguageModels2024, cavanaghSmileyLlamaModifyingLarge2024}, preference optimization (PO) \cite{cavanaghSmileyLlamaModifyingLarge2024}, or the direct extraction of task-specific embeddings from general-purpose LLMs \cite{sadeghiCanLargeLanguage2024}. 
Finally, recent work adapts Chain-of-Thought (CoT) \cite{weiChainofThoughtPromptingElicits2023} chemistry reasoning models following the Deepseek-R1 \cite{deepseek-aiDeepSeekR1IncentivizingReasoning2025} paradigm, e.g., Ether0 \cite{narayananTrainingScientificReasoning2025} fine-tunes Mistral-Small-24B-Instruct \cite{MistralaiMistralSmall24BInstruct2501Hugging} using SFT on Deepseek-R1 reasoning traces and PO on chemistry tasks.
\\
However, a central challenge in chemical machine learning is the scarcity and high cost of labeled data. This presents a significant limitation, as the aforementioned approaches all rely on labeled data for model training. Nevertheless, recent studies have shown that general-purpose LLMs are capable of reasoning over chemical structures, yet this capability is often exercised indirectly.
For instance, general-purpose LLMs have been used to enrich SMILES with text descriptions to fine-tune smaller models \cite{qianCanLargeLanguage2023}, address diverse chemistry tasks via zero-shot and few-shot prompting with varying success \cite{guoWhatCanLarge2023}, and solve chemical mathematical calculations by generating and refining code-based solutions \cite{ouyangStructuredChemistryReasoning2024}.
A final category of applications addresses synthesis planning, the task of identifying viable synthetic routes by deconstructing a target molecule into smaller precursors using reactions until a set of commercially available starting materials is found \cite{seglerPlanningChemicalSyntheses2018, coreyLogicChemicalSynthesis1989}. 
In this context, LLMs can reason about chemical structures to guide and evaluate the synthesis planning process itself based on a desired provided route outcome prompt, without directly manipulating the structures \cite{branChemicalReasoningLLMs2025a}. 
As LLMs tend to struggle with generating high-quality reaction predictions directly, they can be paired with an evolutionary algorithm to reason over and evolve a population of full synthesis routes \cite{wangLLMAugmentedChemicalSynthesis2025}. To ensure chemical validity, this process uses a database of known reactions and molecule routes, which are queried via a nearest-neighbor search in an embedding space to identify structurally similar precedents for chemical grounding.
\\
In this work, we build on these insights to introduce a framework that enables general-purpose LLMs to successfully reason directly over molecular structures. Our method works by anchoring the reasoning process to a molecule's atom-maps, which are unique identifiers for each atom in a molecular SMILES. This approach mirrors a chemist's workflow, \textbf{operates without labeled training data or task-specific model training}, and consists of two stages. First, in a zero-shot task, the model performs a chemical analysis on the chemical structure to identify the atom-maps of relevant fragments for the task and assigns structural labels for these fragments solely based on chemical reasoning. Second, in an optional few-shot task, it transforms the chemical structure based on these identified fragments, guided by examples from a specific chemical transformation class (e.g., a particular reaction or other defined chemical transformation).
\\
We apply this framework to single-step retrosynthesis, where the goal is to identify, given a product molecule, a set of plausible reactant molecules (precursors) that can form the product in a single reaction step \cite{torren-peraireModelsMatterImpact2024}. Formally, the goal is to learn a function $f(P) \rightarrow [R_1, R_2, \dots, R_n]$ that maps a product molecule $P$ to a ranked list of plausible reactant sets, $[R_1, R_2, \dots, R_n]$, where each $R_i$ is a set of one or more reactant molecules, $\{r_1, r_2, \dots\}$, proposed to synthesize $P$.
In this task, prior research shows that general-purpose LLMs are not competitive with specialized models as they underperform their specialized counterparts by more than 40 percentage points in top-1 accuracy \cite{guoWhatCanLarge2023} or solve only one out of five test examples correctly \cite{liChemicalQAEvaluating2025}. 
Our approach marks a shift from conventional supervised methods, which either (1) directly map products to reactants using Transformers \cite{irwinChemformerPretrainedTransformer2022a, tetkoStateoftheartAugmentedNLP2020}, Graph Neural Networks \cite{chenDeepRetrosyntheticReaction2021a, zhongRetrosynthesisPredictionUsing2023}, Markov Bridges \cite{igashovRetrobridgeModelingRetrosynthesis2024}, or fine-tuned LLMs \cite{yangBatGPTChemFoundationLarge2024, nguyen-vanAdaptingLanguageModels2024}, or (2) use a two-step, disconnection-aware paradigm where a model first learns to identify a bond disconnection site and second applies a transformation afterward.
Our approach evolves the second paradigm. Whereas these supervised methods apply a learned mapping by selecting a site either automatically \cite{thakkarUnbiasingRetrosynthesisLanguage2023, kreutterMultistepRetrosynthesisCombining2023} or with human guidance \cite{thakkarUnbiasingRetrosynthesisLanguage2023, westerlundHumanguidedSynthesisPlanning2025a}, our work introduces explicit chemical reasoning as the core mechanism for both steps, leading to the following key contributions:
\begin{enumerate}
    \item We introduce a novel reasoning framework that enables LLMs to zero-shot analyze and few-shot transform molecular structures without task-specific training by anchoring their reasoning process directly to the molecule's SMILES atom maps, thereby eliminating the need for labeled training data or task-specific model training.
    \item We demonstrate the framework's effectiveness in single-step retrosynthesis on both academic benchmarks and expert-validated real drug discovery molecules, where it successfully identifies strategic disconnections, executes the corresponding transformation to predict reactant structures, and provides a chemically-grounded, explainable rationale for its predictions.
    \item We establish a general blueprint for applying LLMs to challenges requiring molecular reasoning and molecular transformations, positioning atom-anchored LLMs as a powerful, data-efficient alternative to supervised learning in low-data chemistry regimes.
\end{enumerate}

\section{Methods}\label{sec2}

\begin{figure*}[t]
\centering
\includegraphics[width=1\textwidth]{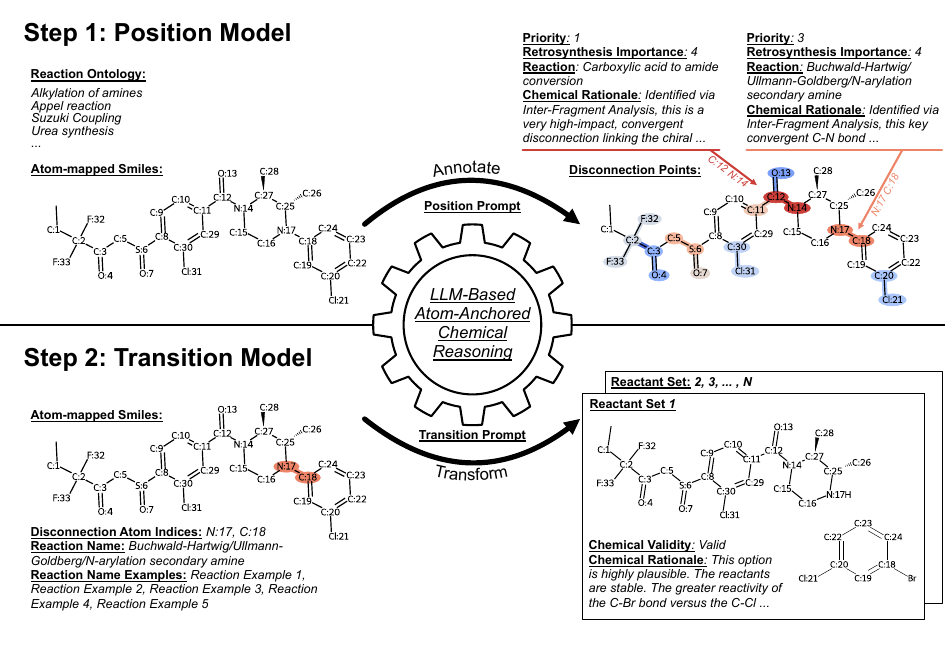}
\caption{Adaptation of our general framework to retrosynthesis. The Zero-Shot Position Model ($f_{\text{position\_retro}}$) guided by Position Prompt ($r_{\text{position}}$) analyzes an atom-mapped product ($m_0$) together with the reaction ontology ($O$) to identify and rank disconnection candidates each containing disconnection points ($S_i$), predicted reactions ($\beta_i$), retrosynthesis importance rankings ($\iota_i$) and a chemical rationale ($\rho_i$). The (optional) Few-Shot Transition Model ($f_{\text{transition\_retro}}$) guided by Transition Prompt ($r_{\text{transition}}$) and a library $L_{\text{retro}}$ of $\beta_i$ reaction examples applies the selected named reaction ($\beta_i$) at the disconnection site ($S_i$) to generate plausible reactant molecules ($R_k$) with validity assessment ($\gamma_k$) and chemical rationale ($\omega_k$).}
\end{figure*}
\subsection{Framework}
Conventional drug discovery models learn a direct mapping $f: \mathcal{X} \rightarrow \mathcal{Y}$, treating molecular representations $x \in \mathcal{X}$ as abstract data points to predict properties $y \in \mathcal{Y}$. This paradigm disregards the underlying chemical knowledge that could govern the relationship $r$ between a molecule's structure and its properties. In contrast, our approach circumvents this data-driven mapping by leveraging the emergent reasoning capabilities of a pre-trained LLM. Guided by a natural language prompt, the LLM performs a detailed chemical analysis with its reasoning explicitly anchored to the molecule's SMILES atom maps, ensuring a precise linkage to specific structural locations. This structurally-grounded analysis enables the direct inference of chemical properties, eliminating the need for task-specific fine-tuning. Our approach operates in two stages:
\\
\textbf{Zero-Shot Structural Analysis and Property Prediction (Position Model):} Guided by a natural language prompt $r_{position}$ that encodes domain knowledge about the task, the LLM analyzes an atom-mapped molecule SMILES $m$ to identify relevant substructures. Based on this prompt-guided reasoning, which is explicitly linked to atom map indices, the position model $f_{position}(m)$ predicts a set of properties $P = \{p_1, \dots, p_n\}$. Each prediction $p_i$ is a tuple $p_i = (S_i, A_i)$, where $S_i \subseteq V(m)$ is a set of atom indices from the molecule $m$ (the structural label), and $A_i = (a_1, a_2, ..., a_k)$ is an ordered tuple of inferred chemical attributes relevant to the task (e.g., "toxic," "reaction"). Each individual attribute $a_i$ in this tuple can be a passive descriptor or an actionable transformation.
\\
\textbf{Prompt-Guided Molecular Transformation (Transition Model):} In an optional second phase, predictions $p_i = (S_i, A_i)$ containing an actionable transformation in their attribute tuple $A_i$ are executed. 
For each general chemical task, a transformation function $f_{transition}$ is defined by a second natural language prompt $r_{transition}$. This transition function executes an actionable attribute $a_j \in A_i$ by applying $f_{transition}$ to an initial molecule $m_0$ at the location $S_i$ to yield a new molecule $m_1$, such that $m_1 = f_{transition}(m_0, S_i,a_j, L)$.
Here, $L$ is a context library providing examples or any relevant information for the established chemical operations identified by the actionable attribute $a_i$ from the tuple $A_i$. This is feasible because many chemical transformations are discrete, well-established operations, allowing in-context learning to ensure chemical validity.
\subsection{A Position Model for Retrosynthesis}
The Position Model emulates a human chemist's analytical workflow to identify and rank potential disconnection sites in a product molecule. Formally, given an atom-mapped product molecule $m$, the Position Model is a function $f_{position\_retro}(m)$ that predicts a set of potential retrosynthetic disconnection candidates, $D = \{d_1, d_2, \dots, d_N\}$. Each candidate  $d_i = (S_i, \beta_i, \iota_i, \rho_i)$, which instantiates the general property prediction $p_i = (S_i, A_i)$ for retrosynthesis, is generated by the function:
$$
D = \{ (S_i, \beta_i, \iota_i, \rho_i) \}_{i=1}^N = f_{position\_retro}(m_0, O)
$$
This function maps a single set of inputs comprising $m_0$ (the atom-mapped target product molecule canonicalized SMILES) and $O$ (a reaction ontology containing reaction names corresponding to a library of executable transformations $L$, providing a bridge to the optional transformation phase) to a set of $N$ distinct output tuples: 
$S_i \subseteq V(m_0)$, the structural label (a set of atom indices defining the disconnection point); $\beta_i$, the predicted reaction name (a chemical attribute identifying a suitable transformation, e.g., "Suzuki Coupling"), grounded in the reaction ontology ($O$) to make reactions actionable (permitting additional flagged reactions outside $O$); $\iota_i \in \mathbb{R}$, the retrosynthesis importance ranking the strategic value of the disconnection (prioritizing the most promising reactions, e.g., major ring-forming reactions, core scaffold construction); and $\rho_i$, the chemical rationale (a text-based justification tied to primary strategic goals of retrosynthesis, e.g., structural simplification, reaction robustness, and stereochemical control).
\\
The entire reasoning process of $f_{position\_retro}$ is defined by a natural language prompt $r_{position}$ (see Prompt \ref{appendix:position_prompt}). Crucially, $r_{position}$ does not contain explicit transformation rules (e.g., SMARTS patterns) or any other reaction-specific rules. Instead, it instructs the LLM to emulate a chemist's analytical workflow. 
Reframing the retrosynthesis task necessitates a shift in evaluation, moving beyond classical top-n performance based on product-reactant replication. Our evaluation instead measures the model's ability to correctly identify the ground-truth disconnection site and reaction type, for which the following metrics are defined:
\begin{itemize}
    \item \textsc{Partial Match Accuracy:} An indicator metric that is true if any predicted disconnection $S_i \in D$ has a non-empty intersection with the ground truth $S_{gt}$.
    \item \textsc{Best Match Jaccard:} The highest Jaccard similarity between any predicted structural label $S_i \in D$ and the ground truth set $S_{gt}$.
    \item \textsc{Exact Match Accuracy:} A stricter metric that is true if the best-matching predicted disconnection site (by Jaccard score) is identical to the ground truth $S_{gt}$.
    \item \textsc{Conditional Reaction Accuracy:} Conditional on a partial match and the highest Jaccard similarity in $D$, this metric evaluates the reaction name(s) $\beta_i$ from the disconnection candidate(s) $d_i$. The metric is true if any of these $\beta_i$ match the ground truth reaction name, $\beta_{gt}$.
\end{itemize}

\subsection{A Transition Model for Retrosynthesis}
To complete the retrosynthesis workflow, we define the Transition Model as $f_{\text{transition\_retro}}$. This model uses a disconnection candidate $d_i$ and a target product $m_0$ to generate a set of plausible reactants $R$. To simulate a chemist's literature lookup for a reaction, the reaction name $\beta_i \in O$ is used to sample up to five reaction examples from a training dataset to create the task-specific, in-context library $L_{\text{retro}}$. The one-to-many Transition Model is then defined as:
$$
\begin{aligned}
&\{ (R_k, \gamma_k, \omega_k) \}_{k=1}^N =\\
&\quad f_{transition\_retro
}(m_0, S_i, \beta_i, L_{\text{retro}})
\end{aligned}
$$
This function maps a single set of inputs comprising $m_0$ (the atom-mapped target product molecule canonicalized SMILES); $S_i$ (the set of disconnection point atom indices); $\beta_i$ (the reaction name, serving as the actionable attribute $a_j$); and $L_{\text{retro}}$ (the context library containing examples of the reaction $\beta_i$) to a set of $N$ distinct tuples: $R_k$, the $k$-th predicted set of reactant molecules $\{r_1, r_2, \dots, r_n\}$; $\gamma_k$, the specific chemical validity assessment (stability, chemoselectivity, stereochemical consistency) for the transformation leading to $R_k$; and $\omega_k$, the specific chemical rationale that justifies the validity of the $k$-th outcome.
\\
The transition function $f_{transition\_retro}$ is defined by prompt $r_{transition}$ (see Prompt \ref{appendix:transition_prompt}), which emulates a chemist's reasoning and avoids explicit reaction rules. Beyond reactant prediction, the model can also generalize transformations by abstracting a reaction template $R_t$, which is flagged accordingly. This template can handle complex cases, such as multiple atoms being viable for reaction side or added reagents, thereby preventing exhaustive iteration. We evaluate performance by comparing the predicted reactant sets, $R_{\text{pred}} = \{R_1, \dots, R_N\}$, against the ground-truth reactants, $R_{\text{gt}}$. As multiple reactant sets can be chemically valid, our goal is to assess the model's ability to recover the known, ground-truth transformation without ranking. The following metrics are calculated per-prediction and averaged across the dataset:

\begin{itemize}
    \item \textsc{Template Accuracy:} Measures if any predicted reactant template set $R_t \in R_{\text{pred}}$ correctly identifies the core structure of the ground-truth reactants $R_{gt}$. A prediction is considered a match if for every ground-truth reactant $r_{gt} \in R_{gt}$ there is a corresponding predicted reactant template $r_t \in R_t$ sharing at least 75\% of its atoms and having a direct substructure match.
    \item \textsc{Reactant Accuracy:} Measures if any predicted reactant set $R_k$ is an exact, non-template match for the ground-truth set $R_{gt}$. 
    \item \textsc{Combined Accuracy:} Measures if a prediction meets either the Template or Reactant Accuracy criterion. 
\end{itemize}

\subsection{Experimental Setup}
We evaluate the Position ($f_{position\_retro}$) and Transition ($f_{transition\_retro}$) models across a diverse set of LLMs to assess the scaling of reasoning capabilities. Our selection includes various open-source models (Qwen3-2507 4B, 30B, 235B \cite{yangQwen3TechnicalReport2025}, DeepSeek-R1-0528 \cite{deepseek-aiDeepSeekR1IncentivizingReasoning2025}), several closed-source models (Gemini 2.5 Flash/Pro \cite{comaniciGemini25Pushing2025}, Claude Sonnet 4 \cite{anthropicClaude4System2025}, GPT5 \cite{openaiGPT5SystemCard2025}), and a chemistry-specialized model, Ether0 \cite{narayananTrainingScientificReasoning2025}. For efficiency, the largest open-source models were quantized for inference on an 8x H100 DGX node and used default inference parameters (see Table \ref{tab:method_llms}). 
\\
We use two public reaction datasets: USPTO50k \cite{loweExtractionChemicalStructures2012, schneiderWhatsWhatNearly2016} and PaRoutes \cite{genhedenPaRoutesFrameworkBenchmarking2022}. For USPTO50k ($n \approx 5 \times 10^4$), we use an adjusted version that corrects a known atom-mapping bias \cite{somnathLearningGraphModels2021}. For PaRoutes ($n \approx 1 \times 10^6$), we use the provided data splits \cite{torren-peraireModelsMatterImpact2024}.
For all datasets, we preprocess the data to generate structural labels ($S_i$), reaction names ($\beta_i$) and reaction ontology ($O$). The labels ($S_i$) define the reaction center by annotating atoms of bonds that are broken, formed, or changed in type from the product's perspective. We prioritize changes in connectivity (bonds breaking or forming) over bond type changes, where the atom structure itself remains unchanged, unless no connectivity change occurs. The reaction names ($\beta_i$) and their reaction classes are extracted using the open-source rxn-insight package \cite{dobbelaereRxnINSIGHTFastChemical2024c}, allowing the release of our labeled data.
The ontology ($O$) is constructed from unique reaction names ($\beta_i$) in the respective training data. 
To mitigate the skewed distribution of reaction names in the USPTO50k test set ($n=5 \times 10^3$) and prevent redundant evaluation, we create a subsampled version, USPTO50k-LLM (see Figure \ref{fig:appendix_uspto50k_test_reaction_name_distribution}). This 541-point evaluation set contains up to five examples per unique reaction name, preserving the original proportion of unclassified reactions. Unless specified otherwise, we use this set with a reaction ontology ($n=136$) derived from the USPTO50k training data.

\section{Results}\label{sec3}

\begin{figure*}[b]
\centering
\includegraphics[width=1\textwidth]{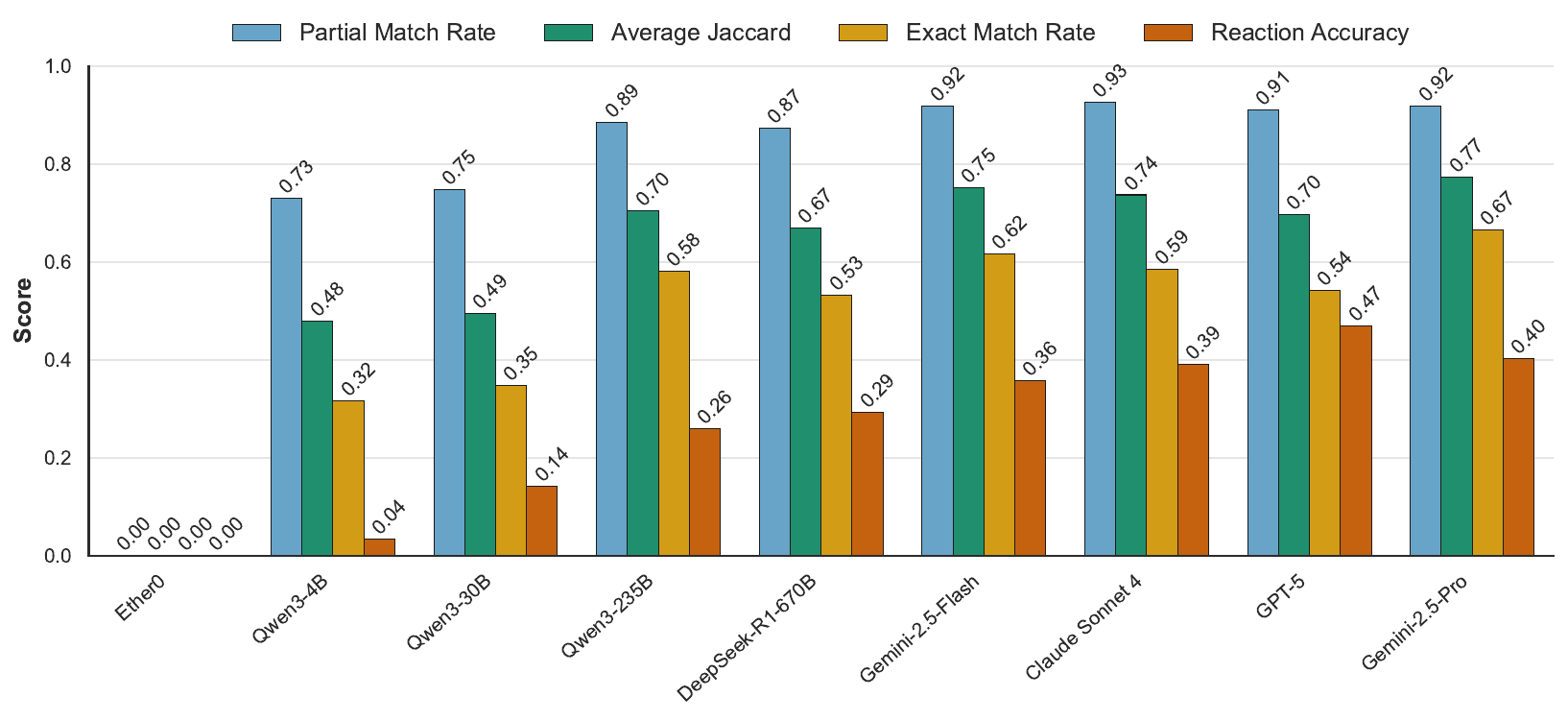}
\caption{Position model performance on USPTO-LLM. The plot compares various foundation models on the task of reaction position prediction, measured by four evaluation metrics: achieving a partial positional match, maximizing the Jaccard metric, identifying the exact position, and predicting the correct reaction (conditional on a partial match).}
\label{fig:position_model_perforamce}
\end{figure*}
\subsection{Position Model}
\begin{figure}[t]
\centering
\includegraphics[width=1.0\columnwidth]{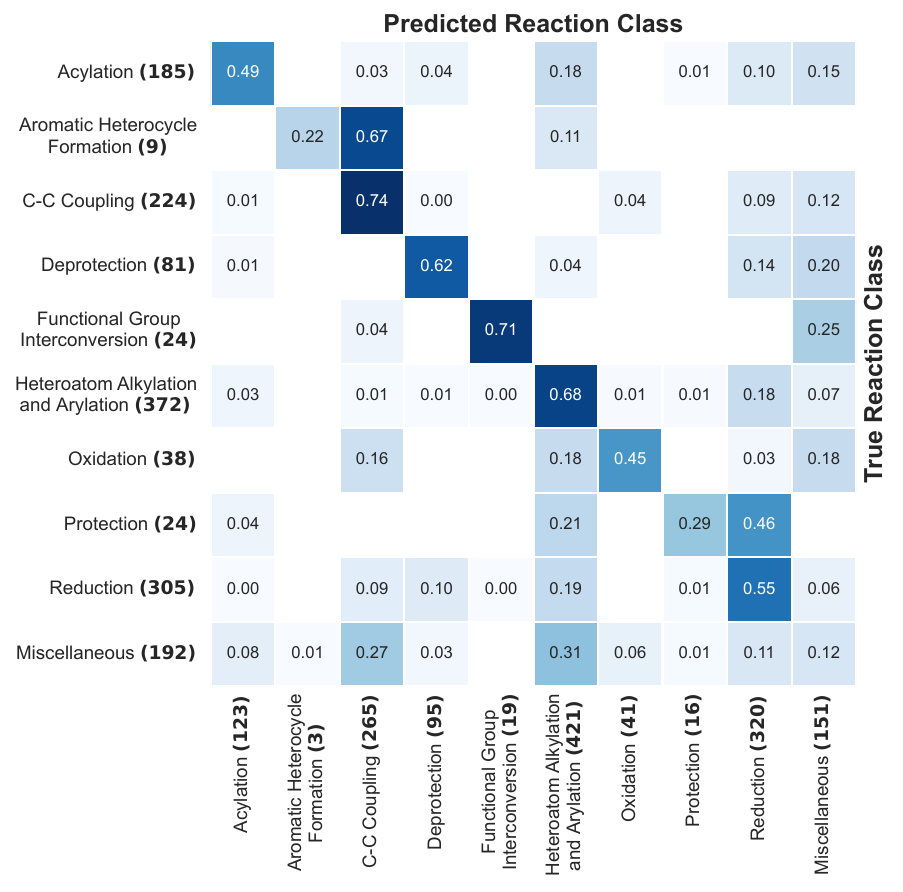}
\caption{Confusion matrix of predicted versus ground-truth reaction classes for the Gemini 2.5 Pro model on USPTO-LLM. The analysis is conditional, including only predictions where the model successfully identified at least a partial positional match. For this visualization, reactions outside the defined reaction ontology were excluded. The matrix was generated using the original class-to-name mappings from the ground-truth data, with any unassigned reactions grouped into the 'Miscellaneous' category.}
\label{fig:confusion_matrix_gemini_2_5}
\end{figure}

Our analysis of structural chemical reasoning shows performance scales with model size, with large closed-source models such as the top-performing Gemini 2.5 Pro required for the best results (see Figure \ref{fig:position_model_perforamce}). We evaluated models on four tasks of increasing difficulty: partial position match, maximizing Jaccard overlap, exact position match, and correct reaction prediction given a partial match. A consistent pattern emerged, where performance increased with the size of the model. For instance, partial match scores jumped from 73\% for 4B models to 87\% for 235B+ models. This trend held across all tasks, with the performance gap becoming most stark on the reaction prediction task, where smaller models scored just 4\%. In contrast, only the largest proprietary models achieved a moderate success rate of $40\text{-}47\%$, showing a trade-off between higher accuracy and lower prediction efficiency (i.e., more predictions per success; see Table \ref{tab:position_model_comparison_compact}). While performance depends on model size, disconnection prediction success is effectively decoupled from molecular size (see Figure \ref{appendix:fig_position_molsize_vs_performance}).
\\
Three models warrant a specific discussion. First, the Ether0 model, a Mistral-24B variant fine-tuned for chemistry, fails to produce any valid predictions, generating neither valid outputs nor chemically valid positions, unlike other models that fail only occasionally (see Table \ref{tab:position_model_comparison_compact}). This total failure suggests that its specialized training, which utilizes chemistry reasoning traces and GRPO on chemical tasks, hindered generalizability to our problem. Second, an ablation of Qwen-235B-Instruct reveals a trade-off with its thinking counterpart. Despite a comparable partial match score, the instruct model showed poor prediction efficiency, generating far more candidate positions, and was only half as effective at identifying the correct reaction (see Table \ref{tab:position_model_comparison_compact}), highlighting the importance of CoT reasoning. Interestingly, this pattern does not appear for Gemini 2.5 Flash, where its thinking and non-thinking versions perform comparably with high reaction accuracy and low prediction efficiency.
\\
Our problem involves a one-to-many relationship in which a chemical position can have multiple valid reactions. To evaluate one of the best performing models, Gemini 2.5 Pro, we mapped its predictions to broader reaction classes using the reaction class mapping from rxn-insight on the ground truth data (see Figure \ref{fig:confusion_matrix_gemini_2_5}). The model often suggests alternative reactions from the correct class rather than predicting a reaction from a different class. However, some exceptions represented chemically plausible alternative strategies: for 'Aromatic Heterocycle Formation', the model often predicted 'C-C couplings', and for 'Protection' reactions, it suggested 'Reductions'. The 'Heteroatom Alkylation and Arylation' class was a notable outlier, being proposed for most other categories except 'FGI' and 'C-C couplings'. This predictive pattern of staying within-class and these specific exceptions also holds at the individual reaction-name level (see Figure \ref{fig:appendix_position_model_confusion_matrix_reaction_names}).
\subsection{Transition Model}
\begin{figure*}[t]
\centering
\includegraphics[width=1\textwidth]{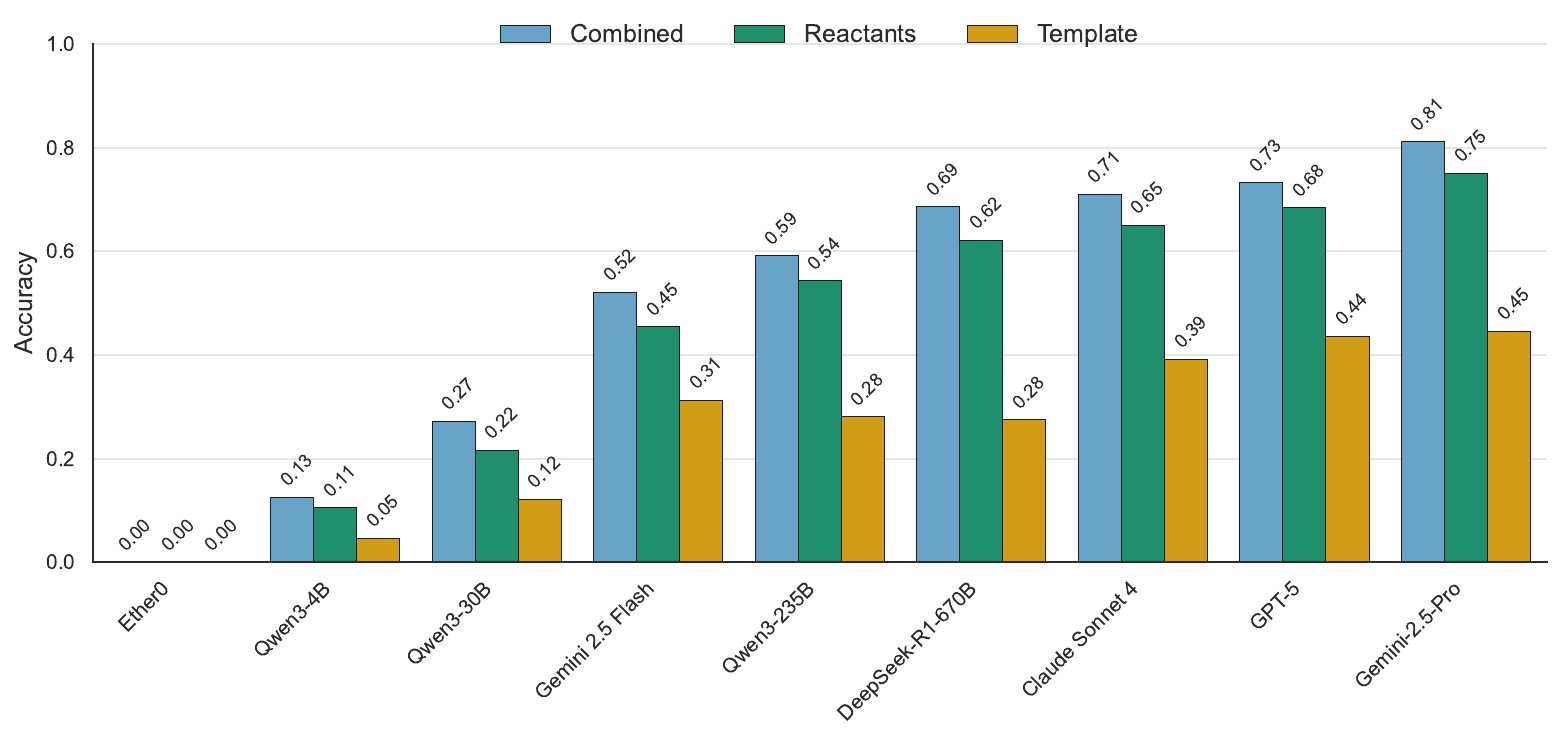}
\caption{Transition Model Performance on USPTO-LLM. The plot evaluates various LLMs on their ability to predict chemical transformations. Accuracy is measured using three metrics: direct reactant prediction ('Reactants'), valid template generation ('Template'), and a combined approach where either is considered a success ('Combined').}
\label{fig:results_transition_model}
\end{figure*}

We evaluated various LLMs on their ability to predict ground-truth transformations using the reaction's position, name, and up to five examples (see Figure \ref{fig:results_transition_model}). Model performance scales logarithmically with size before plateauing at the scale of Deepseek-R1. Gemini 2.5 Pro is the top performer, excelling both at direct reactant prediction ("Reactant"; see example Figure \ref{appendix:application_lei_515_correct}) and in combination with a reaction template ("Combined"). This template generation ("Template"; see example Figure \ref{appendix:application_lei_515_correct_template}), which is a proxy for chemical understanding, is strongest in proprietary models, such as GPT-5 and Gemini 2.5 Pro ($45\%$ accuracy). In contrast, Deepseek-R1 performs worse than its smaller open-source peers in template prediction, while Ether0 fails again at this task.
\\
In a first ablation study, our results reveal the critical importance of a defined reaction name to act as a \textit{chemical anchor} (see Figure \ref{appendix:transition_known_unknown}). Performance dropped by approx. $50\%$ for unknown reactions in a zero-shot setting (no examples provided) compared to known ones in a few-shot setting (up to five examples). The decline was particularly severe for the prediction of direct reactants, with accuracy falling from approximately $83\%$ for known reactions to $25\%$ for unknown reactions (Gemini 2.5 Pro). In a second ablation study on Gemini 2.5 Pro, we further isolate the contributions of prompt detail versus few-shot examples on overall ("combined") performance (see Figure \ref{appendix:fig_transition_prompt_applcation}). Although the model achieved baseline combined accuracy using a minimal prompt (49\%), and the detailed prompts offered some improvement through the reaction template (57\%), the inclusion of examples was the dominant factor (65\%); a simple prompt with examples was much more effective than a detailed prompt without them. The best performance required both (81\%). Finally, CoT reasoning improves reactant and combined accuracy, but it underperforms non-reasoning models on reaction template prediction, at the cost of lower prediction efficiency (see Qwen3-235B in Table \ref{tab:ablation_transition_model_table}).
\\
With performance again independent of molecular size (see Figure \ref{appendix:fig_transition_molsize_vs_performance}), analyzing LLM failure modes reveals two distinct error types. First, reaction class-specific performance variations among the top-performing models indicate that no single model is universally superior, suggesting solutions such as multi-model ensembles or best-of-n sampling (see Figure \ref{fig:appendix_transition_model_confusion_matrix}). Second, all models consistently fail on a small set of reaction classes with few data points (e.g., Wohl-Ziegler bromination). This systemic failure likely stems from data deficiencies, such as incorrect labeling and poor examples that make the task ill-posed, rather than fundamental mechanistic reasoning challenges for current LLM architectures.

\subsection{Real-World Application}
\begin{figure*}[h]
\centering
\includegraphics[width=1\textwidth]{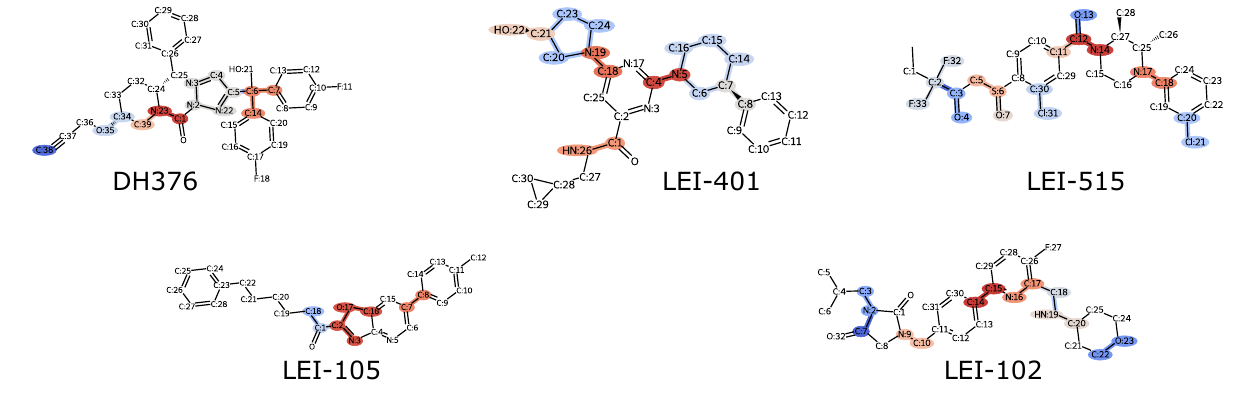}
\caption{Zero-shot position model prediction for five real-world drug discovery molecules used in our case study using the PaRoutes reaction ontology highlighting suggested disconnection sites: DH376 \cite{dengTriazoleUreasAct2017}, LEI-102 \cite{liStructuralBasisSelective2023}, LEI-105 \cite{baggelaarHighlySelectiveReversible2015}, LEI-401 \cite{mockDiscoveryNAPEPLDInhibitor2020}, LEI-515 \cite{jiangMonoacylglycerolLipaseInhibitor2023}.}
\label{fig:five_mols}
\end{figure*}

While LLMs demonstrated strong performance on USPTO50k, such academic tests risk data contamination for models pre-trained on vast data corpora. To conduct a more rigorous, real-world validation, we evaluated our approach on five molecules that were previously synthesized and published in high-impact journals (see Figure \ref{fig:five_mols}), for which we were able to discuss the experimental procedures with the respective lab chemists. 
For this evaluation, we used one of our top-performing LLMs (Gemini 2.5 Pro) with the PaRoutes reaction ontology (n=335) and annotated atom-maps by sequentially counting the atoms in a canonicalized SMILES. Our position model first proposed potential disconnection points, which the respective lab chemist of the molecule then curated for chemical relevance and to avoid redundancy for the transition model evaluation (an example for LEI-515 is provided in Table \ref{table:appendix_disconnections-lei-515}). This process yielded 63 distinct position predictions for assessment and 19 selected positions with a total of 98 transitions. Afterwards, the chemist assessed these predictions against predefined questions, and we calculated accuracy as the percentage of correct model responses (see Table \ref{tab:application_real_molecules}).
\begin{table}[h]
  \caption{Chemist evaluation results for the Position model (P) and Transition Model (T). $n$ indicates the number of data points evaluated, and Acc. denotes the accuracy ($\%$ of correct predictions). Actionable refers here to non-template and not to chemically invalid predicted reactant sets. We provide a full overview of questions and results in the appendix (see Table \ref{appendix:full_questions} \& \ref{appendix:application_questionaire_responses})}
  \label{tab:application_real_molecules}
  \begin{center}
    \begin{small}
        \begin{tabular}{lcc}
          \toprule
          \textbf{Metric} & $\textbf{n}$ & \textbf{Acc.} (\%) \\
          \midrule
          \multicolumn{3}{l}{\textit{Position Model (P)}} \\
          P1: Position Plausibility & 63 & 90.5 \\
          P2: Reaction Name Correctness & 63 & 85.7 \\
          P3: Chemical Reasoning & 63 & 73.0 \\
          P4: Lab Feasibility & 63 & 77.8 \\
          P5: Lab Precedent & 63 & 25.4 \\
          P6: Missing Disconnections for Mol. & 5 & 80.0 \\
          \midrule
          \multicolumn{3}{l}{\textit{Transition Model (T)}} \\
          T1: Template Validity & 16 & 81.3 \\
          T2: Template Chemical Reasoning & 16 & 87.5 \\
          T3: Any Valid Reactants Found & 19 & 89.5 \\
          T4: Reactant Reasoning Correct & 19 & 89.5 \\
          T5: Lab Reactant Match & 15 & 73.3 \\
          T6: Invalidity Reasoning Correct & 7 & 100.0 \\
          T7: Actionable Reactant Accuracy & 98 & 74.5 \\
          \bottomrule
        \end{tabular}
    \end{small}
  \end{center}
  \vskip -0.1in
\end{table}
\\
The case study results were highly encouraging. The model's suggested disconnection points (P1: 90.5\%) and associated reaction names (P2: 85.7\%) were overwhelmingly judged as chemically plausible, with the latter often providing non-obvious alternatives to our expert chemists. While the correctness of chemical reasoning over both positions and reactions was lower (P3: 73.0\%), a majority of all suggestions (P4: 77.8\%) were deemed applicable in a real-world laboratory setting. Notably, the model rediscovered 25.4\% of the experimentally validated disconnections (P5). This figure is lower because the model often proposes multiple valid reactions for a single position, where only one would be used in practice.
However, the system has limitations. For four of the five molecules evaluated, the model missed disconnections anticipated by our chemists (P6). It might, for example, propose a feasible reaction (e.g., Buchwald-Hartwig coupling) where an expert would prefer an alternative (e.g., an S$_{N}$Ar reaction). Our analysis indicates that errors typically originate from the LLM's misinterpretation of the molecular structure (e.g., the misidentified Cl position in Table \ref{table:appendix_disconnections-lei-515}, position 10). This initial error then propagates through the prediction, ultimately leading to an incorrect suggestion for the position, reaction, or reasoning.
Conversely, a key strength of the position model is its ability to provide a comprehensive set of plausible disconnections for an entire synthetic route, not just a single retrosynthetic step. Our chemists considered these predictions valid if the proposed disconnection could occur at any stage of the synthesis route. Importantly, the position model demonstrates the capacity to suggest advanced chemical concepts, such as stereoselective reactions (see Table \ref{table:appendix_disconnections-lei-401}, positions 5 and 6).
\\
The transition model also demonstrated strong performance. It achieved 81.3\% accuracy for predicting reaction templates (T1) and 87.5\% for the associated reasoning (T2), although chemists noted it worked mainly for standard reactions and is less reliable for complex ones (see Figure \ref{appendix:application_lei_515_correct_template}). In 89.5\% of cases, the model generated at least one chemically valid reactant set with sound reasoning (T3 \& T4, see Figure \ref{appendix:application_lei_515_correct}), a reasoning quality judged comparable to that of a master's or PhD-level chemist. Furthermore, it successfully identified 73.3\% of reactants previously conducted in the lab (T5). A key strength was its perfect (100\%) accuracy in identifying non-viable reactions (T6, see Figure \ref{appendix:application_lei_515_correct_chemically_invalid}), correctly explaining why a proposed reaction would fail (e.g., identifying that a specific atom cannot exist at a given position). This highlights its role as a filter, as it sometimes corrected position model suggestions by proposing more intuitive reactions or filtering out disconnections that were invalid without prerequisite synthesis steps.
The model achieved a 74.5\% overall accuracy in predicting reactants after excluding predictions that were reaction template-based or flagged as chemically invalid (T7). Failures typically occurred in one of two ways: the model either failed to return any valid reactant set (accounting for 15/29 failures in our evaluation), or it failed due to incorrect SMILES parsing (see Figure \ref{appendix:application_lei_515_correct_smiles_broken}), even when the underlying chemical reasoning was correct.

\section{Discussion}\label{sec4}


\begin{figure*}[t]
\centering
\begin{tcolorbox}[colback=white!5, colframe=blue!70, fontupper=\small, title={\small\sffamily Deepseek-R1 reasoning trace excerpt for LEI-515 (verbatim, abridged)}, left=4pt, right=4pt, top=4pt, bottom=4pt]
\raggedright
\textit{[...Step 1: Identify All Candidate Transformations...]}\\[4pt]
\textbf{3.} Disconnection at the amide bond: C:12-N:14. This would give a carboxylic acid (from C:12) and the amine (from N:14). The amine is part of the piperazine ring. The transformation is amide formation.\\[4pt]
\quad Disconnection: breaking C:12-N:14. Atoms: C:12 and N:14.\\[6pt]
\textbf{4.} Disconnection at the bond between the piperazine ring and the aryl group attached to N:17. The bond is N:17-c:18. This is an N-aryl bond. Disconnection would give an amine (the piperazine) and an aryl halide? So the transformation is N-arylation.\\[4pt]
\quad Disconnection: breaking N:17-c:18. Atoms: N:17 and c:18 (specifically, the carbon of the aryl ring that is attached? We can use the atom mapping: c:18 is the carbon attached to N:17? So atoms: N:17 and c:18.\\[6pt]
\rule{\linewidth}{0.4pt}\\[4pt]
\textit{[...Step 2: Assign Candidate Reactions...]}\\[4pt]
\textbf{4.} ``N:17 c:18'':Disconnection of the N-aryl bond. Forward reactions:\\
-- N-arylation of the amine (piperazine) with an aryl halide?\\
-- Reactions:\\
\quad -- ``Buchwald-Hartwig/Ullmann-Goldberg/N-arylation secondary amine'' $\rightarrow$ because the amine (N:17) is tertiary? Wait, in the product, N:17 is attached to the aryl group and to two carbons? So its a tertiary amine? But the arylation of a tertiary amine?\\[4pt]
\quad -- Actually, the amine in the forward reaction would be secondary? Because we disconnect the aryl group? So the amine fragment would be: the piperazine with N:17 being secondary (attached to two carbons: C:16 and C:25, and H).\\[4pt]
\quad -- So its a secondary amine?\\[4pt]
\quad -- Then we have: ``Buchwald-Hartwig/Ullmann-Goldberg/N-arylation secondary amine''
\end{tcolorbox}
\caption{Verbatim excerpt from the Deepseek-R1 reasoning trace for LEI-515, illustrating atom-anchored retrosynthetic reasoning. The model identifies the amide bond (C:12--N:14) and the N-aryl bond (N:17--c:18) as strategic disconnections by referencing atom-mapped positions, then self-corrects its initial assumption about amine substitution to correctly assign \textit{Buchwald-Hartwig/Ullmann-Goldberg/N-arylation secondary amine}. Full annotated trace by expert chemists in Section \ref{appendix:reasoning_trace}.}
\label{fig:reasoning_trace_excerpt}
\end{figure*}

Our results demonstrate that grounding chain-of-thought reasoning to molecular structures via atom-anchors enables general-purpose LLMs to perform retrosynthesis. 
Here, atom-anchors allow LLMs to analyze the molecular structure in depth, identify functional groups, and transfer chemical reaction knowledge from the pre-trained LLM to the molecular structure without task-specific fine-tuning (see Figure \ref{fig:reasoning_trace_excerpt} for an excerpt of the Deepseek-R1 reasoning trace for LEI-515; full trace annotated by expert chemists in Section \ref{appendix:reasoning_trace}).
\\
Given that the performance of both models is independent of molecular size (Figures \ref{appendix:fig_position_molsize_vs_performance}, \ref{appendix:fig_transition_molsize_vs_performance}), we additionally explored whether our framework generalizes to complex, data-scarce modalities beyond standard small molecules. 
These modalities pose a particular challenge as their complex scaffolds are underrepresented as products in public reaction datasets, limiting the ability of supervised models to generalize to them.
For molecular glues, the position model identifies strategic disconnections for TRAP-1 \cite{zhuActivatingP53Y220CMutantSpecific2024} consistent with the originally reported synthesis (Figure \ref{appendix:trap_1}). Similarly, it successfully navigates the complex topology of macrocycles, correctly predicting the strategic ring-closing reaction for the MCL-1 inhibitor compound 25 \cite{tarrDiscoveryMacrocyclicMyeloid2025} (Figure \ref{appendix:fig_mcl_1}). These results confirm that our atom-anchored reasoning is not limited to common benchmark scaffolds but extends to novel, high-value modalities where traditional supervised models often fail due to the absence of training data.
\\
Furthermore, we observe that the atom-anchored reasoning traces and chemical rationale are not strictly limited to retrosynthesis as the LLMs reason over adjacent tasks like forward synthesis (see Section \ref{appendix:reasoning_trace} for the Deepseek-R1 reasoning trace of LEI-515) and reagent prediction (e.g., Gemini 2.5 Pro flags the MCL-1 disconnection 3 as unfavorable because of "hazardous reagents", see Table \ref{table:appendix_mcl_1_cmpd_25}).
For multi-step synthesis planning, the position model analyzes \textit{all} strategic disconnections in the molecule holistically (see Table \ref{table:appendix_disconnections-lei-515} for LEI-515, Table \ref{table:appendix_trap_1} for TRAP-1, and Table \ref{table:appendix_mcl_1_cmpd_25} for MCL-1). This output effectively provides a strategic synthesis plan for all possible disconnections in a molecule. Although we do not ask LLMs to provide an ordering for creating a synthesis route, they exhibit inherent multi-step logic. For example, Deepseek-R1 explicitly reasons over multiple reaction steps (see Section \ref{appendix:reasoning_trace}). These holistic multi-step predictions have two important consequences: First, the generated positions constrain the search space for a synthesis planning algorithm (e.g., \cite{hassenSynthesisPlanningReaction2025}), streamlining the identification of an optimal reaction sequence \cite{westerlundHumanguidedSynthesisPlanning2025a, kreutterMultistepRetrosynthesisCombining2023}. Second, these predictions highlight vectors for molecular modification, proving invaluable for guiding and accelerating medicinal chemistry campaigns by providing a strategic blueprint for replacing molecular cores or side-chains, while using a user-defined reaction ontology for robotic or parallel chemistry (e.g., \cite{dombrowskiChosenFewParallel2022}).
\\
From a practical standpoint, we contrast the costs and real-world value our approach provides in comparison to contemporary approaches. While methods like the single-step retrosynthesis model in AiZynthFinder \cite{saigiridharanAiZynthFinder40Developments2024} run locally with negligible cost, our approach requires one LLM call per position model to identify all possible disconnections for a molecule, and then one call per transition model evaluation for each disconnection. With Gemini 2.5 Pro, these individual calls cost on average \$0.07 each (see Table \ref{tab:costs}). However, traditional single-step models output a list of disconnection reactions without an underlying chemical reasoning process. These are essentially "raw reaction ideas" that require significant human time to validate and offer no control over either the selected reaction or its position. Thus, the free local inference is offset by the high labor cost of an expert-level chemist needed to filter and rationalize these predictions. In comparison, our LLM framework performs selected reactions at a specified molecular position while providing expert-level chemical rationale, a process that is parallelizable at scale beyond singular structures and requires minimal human intervention.

\section{Conclusion}\label{sec5}

In this work, we introduced a molecular reasoning framework that leverages the chemical knowledge in general-purpose LLMs to address data scarcity in computational chemistry without requiring labeled training data or task-specific model training.
\\
Our framework grounds chain-of-thought reasoning to molecular structures using atom-mapped SMILES as chemical anchors and operates in two stages: a zero-shot position model that identifies disconnection sites and reaction classes, and a position-aware few-shot transition model that predicts reactant structures. Across academic benchmarks and expert-validated drug discovery molecules, this enables our best-performing LLM to achieve high success rates in identifying chemically plausible reaction sites ($\geq90\%$), named reaction classes ($\geq40\%$), and final reactants ($\geq74\%$) without task-specific training.
Beyond direct retrosynthesis prediction, this work has broader utility as our approach already shows generalization to larger drug-like modalities, adjacent tasks such as forward synthesis and reagent prediction, but also provides an explainable chemical rationale for its decisions. 
\\
In the future, by treating the outputs of our position model as the result of a zero-shot data labeling process, our framework could be used to generate realistic synthetic datasets in data-scarce chemistry domains. This would be achieved by mapping high-level chemical concepts, such as reactions, directly from the intrinsic chemistry knowledge of an LLM to molecular structures, which could enable LLM-based downstream applications, such as the generation of novel, synthetically feasible candidates in de novo drug design.
\\
There are limitations to our work. Currently, our method should not be used as a substitute for a single-step retrosynthesis model in multi-step synthesis planning, as it would be cost- and time-prohibitive to call the model repeatedly, and SMILES structure generation from the transition model can be brittle. Furthermore, the performance of our framework hinges on the capabilities of the underlying LLM, where performance may change over time due to model updates. Finally, strong performance currently requires paid frontier-scale models, and errors in the LLM's interpretation of molecular structure can propagate through the prediction pipeline, leading to incorrect suggestions for the position, reaction, or reasoning.
\\
Summarizing, our work presents a general blueprint for applying LLMs to challenges where molecular reasoning and molecular transformations are key, establishing atom-anchored LLMs as a powerful and data-efficient addition to the modern drug discovery toolbox.

\backmatter

\bmhead{Supplementary information}
Additional information is available in the appendix.


\bmhead{Acknowledgements}
This study was partially funded by the European Union's Horizon 2020 research and innovation program under the Marie Skłodowska-Curie Innovative Training Network European Industrial Doctorate grant agreement No. 956832 “Advanced machine learning for Innovative Drug Discovery”. 
Large Language Models (LLMs) were used throughout the creation of this manuscript to improve spelling mistakes, grammar, and the overall reading flow. All LLM suggestions were carefully checked for correctness and refined by the authors of this work. The LLM was not used for any research-related tasks.


\section*{Declarations}


\begin{itemize}
\item Code/Data availability: The code for this work can be found at \url{https://github.com/AlanHassen/atom-anchored-llms-speak-chemistry}. The datasets and raw LLM response files can be found in the \textsc{data/} directory. Figures and tables used in this manuscript can be reproduced via Jupyter notebooks included in the \textsc{notebooks/} directory.
\end{itemize}

%

\onecolumn
\begin{appendices}

\section{Appendix}\label{secA1}

\subsection{Experimental Setup}

\begin{table}[h]
\caption{A summary of the Large Language Models (LLMs) evaluated in this work. The table specifies whether the model is open-source, its status as a reasoning-optimized ("Thinking") variant, and its thinking budget allocation (in number of tokens) for closed-source models along with other parameters. }
\label{tab:method_llms}
\scriptsize
\begin{tabular}{llccccc}
\toprule
{\bf Source} & {\bf Model Name} & \bfseries \shortstack{Thinking \\ model} & \bfseries \shortstack{Open- \\ Source} & \bfseries \shortstack{Model \\ quantization} & \bfseries \shortstack{Max output \\ length}& \bfseries \shortstack{Thinking \\ budget} \\ 
\midrule \\
\cite{yangQwen3TechnicalReport2025} & Qwen3-4B-Thinking-2507 & yes & yes & &  32768 & -\\
\cite{narayananTrainingScientificReasoning2025} & Ether0 (24B) & yes & yes & &  32768 & -\\
\cite{yangQwen3TechnicalReport2025} & Qwen3-30B-A3B-Thinking-2507 & yes & yes & 8bit &  32768 & -\\
\cite{yangQwen3TechnicalReport2025} & Qwen3-235B-A22B-Instruct-2507-FP8 & no & yes & 8bit & 32768 & -\\
\cite{yangQwen3TechnicalReport2025} & Qwen3-235B-A22B-Thinking-2507-FP8 & yes & yes & 8bit & 32768 & -\\
\cite{deepseek-aiDeepSeekR1IncentivizingReasoning2025} & RedHat-DeepSeek-R1-0528-w4a16 (670B) & yes & yes & 4bit & 32768 & - \\
\cite{comaniciGemini25Pushing2025} & Gemini 2.5 Flash & yes & no & API & 65536 & 30000  \\
\cite{comaniciGemini25Pushing2025} & Gemini 2.5 Pro & yes & no & API & 65536 & 30000 \\
\cite{anthropicClaude4System2025} & Claude Sonnet 4 & yes & no & API & 64000 & 30000 \\
\cite{openaiGPT5SystemCard2025} & GPT5 & yes & no & API & 128000 & 'High' \\
\bottomrule
\end{tabular}
\end{table}

\begin{table}[h]
\caption{
Cost per model call derived from official provider API pricing. Variations in input/output token counts are attributed to differences in tokenizer architectures and model verbosity. Costs for open-source models are excluded, as they rely on variable hardware configurations for inference.
}
\label{tab:costs}
\scriptsize
\begin{tabular}{lccccc}
\toprule
\bf{Model}              & \bf{Task} & \bf{Molecules} & \bfseries \shortstack{Avg. Input \\ Tokens} & \bfseries \shortstack{Avg. Output\\ Tokens} & \bf{Avg. Cost Mol.} \\ 
\hline
\\
Gemini 2.5 Pro     & Position      & 541                & 9202.7                                                               & 5917.3                                                                & 0.071\$            \\ 
GPT5              & Position      & 538                & 4244.64                                                              & 6952.1                                                                & 0.075\$            \\ 
Claude Sonnet 4    & Position      & 538                & 4926.1                                                               & 9123.0                                                                & 0.152\$            \\ 
Gemini 2.5 Flash   & Position      & 539                & 11604.9                                                              & 6572.2                                                                & 0.020\$            \\ 
RedHat-DeepSeek-R1-0528-w4a16 (670B)   & Position      & 541                & 4183.1                                                               & 12232.0                                                               & -                  \\ 
Qwen3-235B-A22B-Thinking-2507-FP8 & Position      & 541                & 4365.0                                                               & 16518.0                                                               & -                  \\ 
Qwen3-30B-A3B-Thinking-2507           & Position      & 541                & 4365.0                                                               & 13287.8                                                               & -                  \\ 
Qwen3-4B-Thinking-2507            & Position      & 541                & 4365.0                                                               & 13410.4                                                               & -                  \\ 
Ether0             & Position      & 541                & 4229.3                                                               & 739.5                                                                 & -                  \\
\midrule
\\
Gemini 2.5 Pro     & Transition    & 540                & 9301.1                                                               & 6226.9                                                                & 0.074\$            \\ 
GPT5              & Transition    & 538                & 3766.2                                                               & 14288.6                                                               & 0.148\$            \\ 
Claude Sonnet 4    & Transition    & 541                & 4059.6                                                               & 5056.5                                                                & 0.103\$            \\ 
Gemini 2.5 Flash   & Transition    & 537                & 10327.8                                                              & 4579.15                                                               & 0.015\$            \\ 
RedHat-DeepSeek-R1-0528-w4a16 (670B)   & Transition    & 528                & 4408.7                                                               & 10847.9                                                               & -                  \\ 
Qwen3-235B-A22B-Thinking-2507-FP8 & Transition    & 537                & 4435.5                                                               & 17550.0                                                               & -                  \\ 
Qwen3-30B-A3B-Thinking-2507           & Transition    & 535                & 4435.5                                                               & 15002.7                                                               & -                  \\ 
Qwen3-4B-Thinking-2507            & Transition    & 529                & 4435.5                                                               & 15311.0                                                               & -                  \\ 
Ether0             & Transition    & 541                & 4542.6                                                               & 5512.3                                                                & -                  \\ 
\bottomrule
\end{tabular}
\end{table}

\newpage
\begin{figure}[t!]
\centering
\includegraphics[width=1\textwidth]{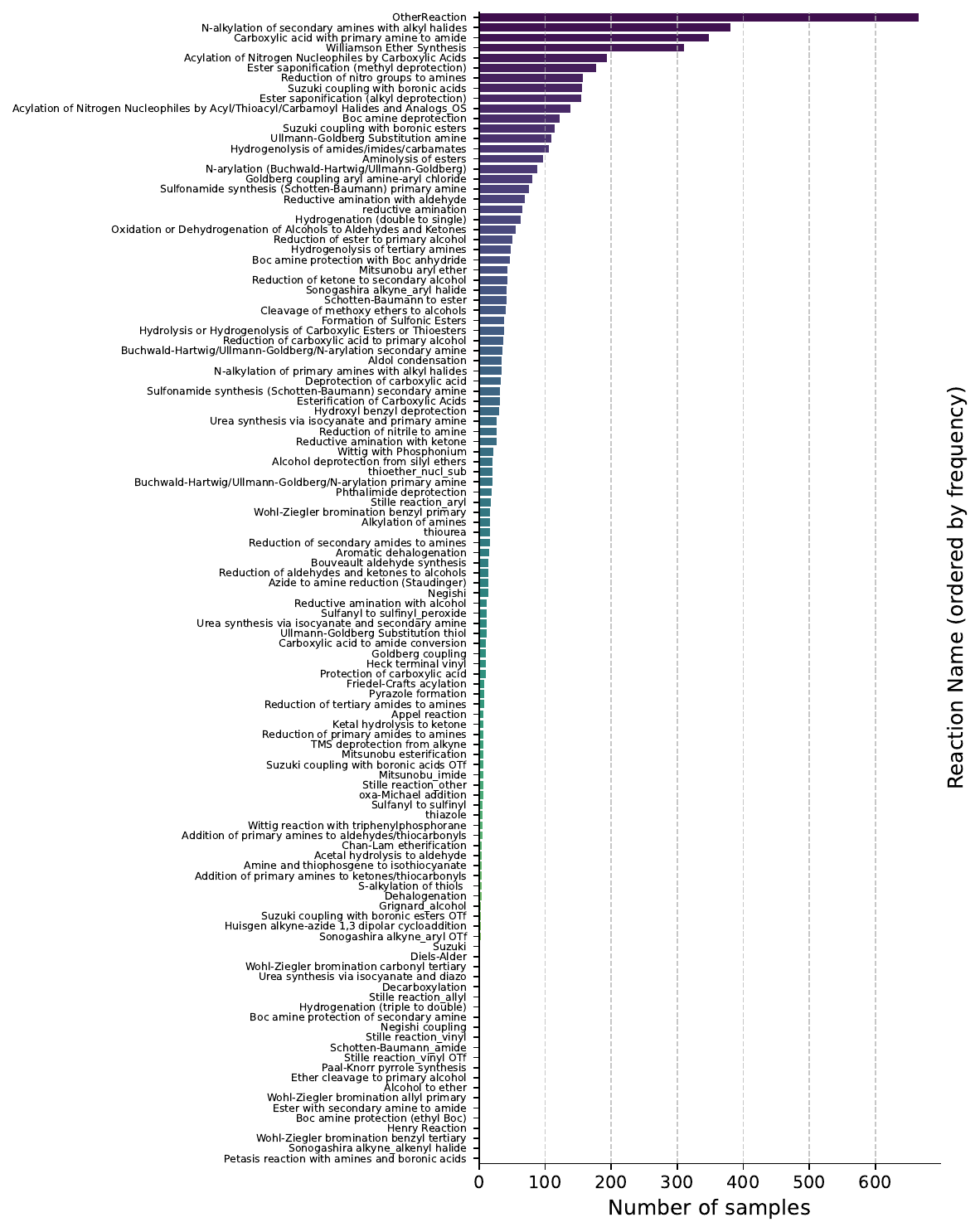}
\caption{Distribution of reaction names in the USPTO-50k test set. From this dataset, we created a balanced subsample (USPTO-LLM) for evaluation by selecting up to five examples per named reaction class, while maintaining the original proportion of the 'otherReaction' class.}
\label{fig:appendix_uspto50k_test_reaction_name_distribution}
\end{figure}

\clearpage
\subsection{Position Model}

\begin{table}[h]
\caption{\small A comprehensive comparison of various models based on several key performance metrics. The table highlights the average number of predictions (number of proposed disconnection points per model call), partial and exact match percentages, reaction accuracy, average Jaccard similarity metric and the total number of successes and failures for each model. The best performance in each column is highlighted in bold.}
\label{tab:position_model_comparison_compact}
\centering
\scriptsize
\begin{tabular}{lccccccc}
\toprule
\bfseries Model & \bfseries \shortstack{Avg. number \\ of predictions} & \bfseries \shortstack{Partial \\ match (\%)} & \bfseries \shortstack{Exact \\ match (\%)} & \bfseries \shortstack{Reaction \\ acc. (\%)} & \bfseries \shortstack{Avg. \\ Jaccard} & \bfseries \shortstack{Total \\ predictions} & \bfseries \shortstack{Failed \\ predictions} \\
\midrule
Qwen3-4B & 4.0 & 73.01 & 31.61 & 3.51 & 0.48 & 541 & 0 \\
Qwen3-30B & 3.8 & 74.86 & 34.75 & 14.23 & 0.49 & 541 & 0 \\
Qwen3-235B & 5.9 & 88.50 & 58.07 & 25.97 & 0.70 & 539 & 2 \\
DeepSeek-R1-670B & 7.3 & 87.25 & 53.23 & 29.21 & 0.67 & 541 & 0 \\
Gemini-2.5-Flash & 16.3 & 91.84 & 61.60 & 35.81 & 0.75 & 539 & 2 \\
Claude Sonnet 4 & 10.0 & \textbf{92.57} & 58.55 & 39.03 & 0.74 & 538 & 3 \\
GPT-5 & 15.1 & 91.08 & 54.28 & \textbf{47.03} & 0.70 & 538 & 3 \\
Gemini-2.5-Pro & 11.1 & 91.87 & \textbf{66.54} & 40.30 & \textbf{0.77} & 541 & 0 \\
\bottomrule
\end{tabular}
\end{table}

\begin{figure}[h]
\centering
\includegraphics[width=1\textwidth]{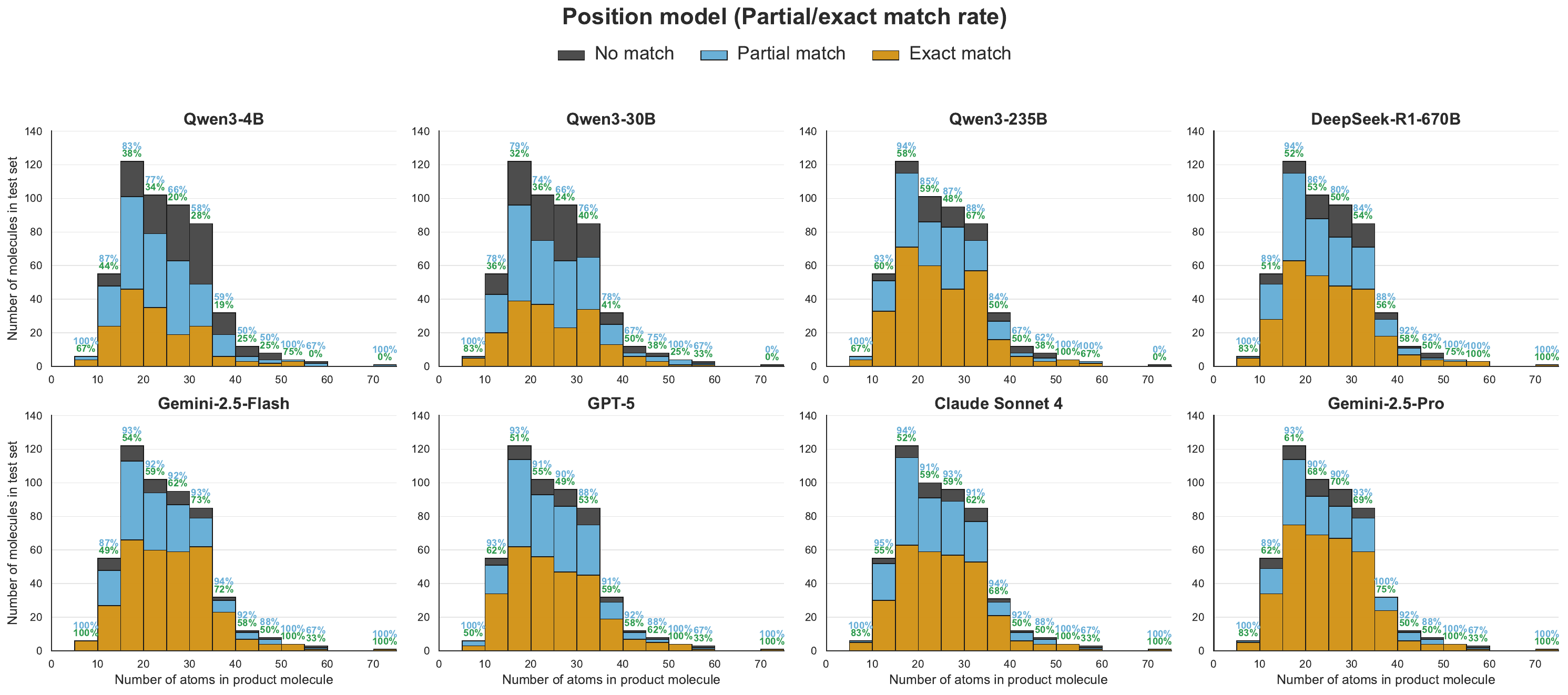}
\caption{Impact of molecule size on position model performance. The figure displays exact and partial match accuracy for predicted disconnection positions, stratified by the number of atoms in product molecules (bin size = 5) across all tested LLMs.}
\label{appendix:fig_position_molsize_vs_performance}
\end{figure}

\begin{figure}[h]
\centering
\includegraphics[width=1.0\textwidth]{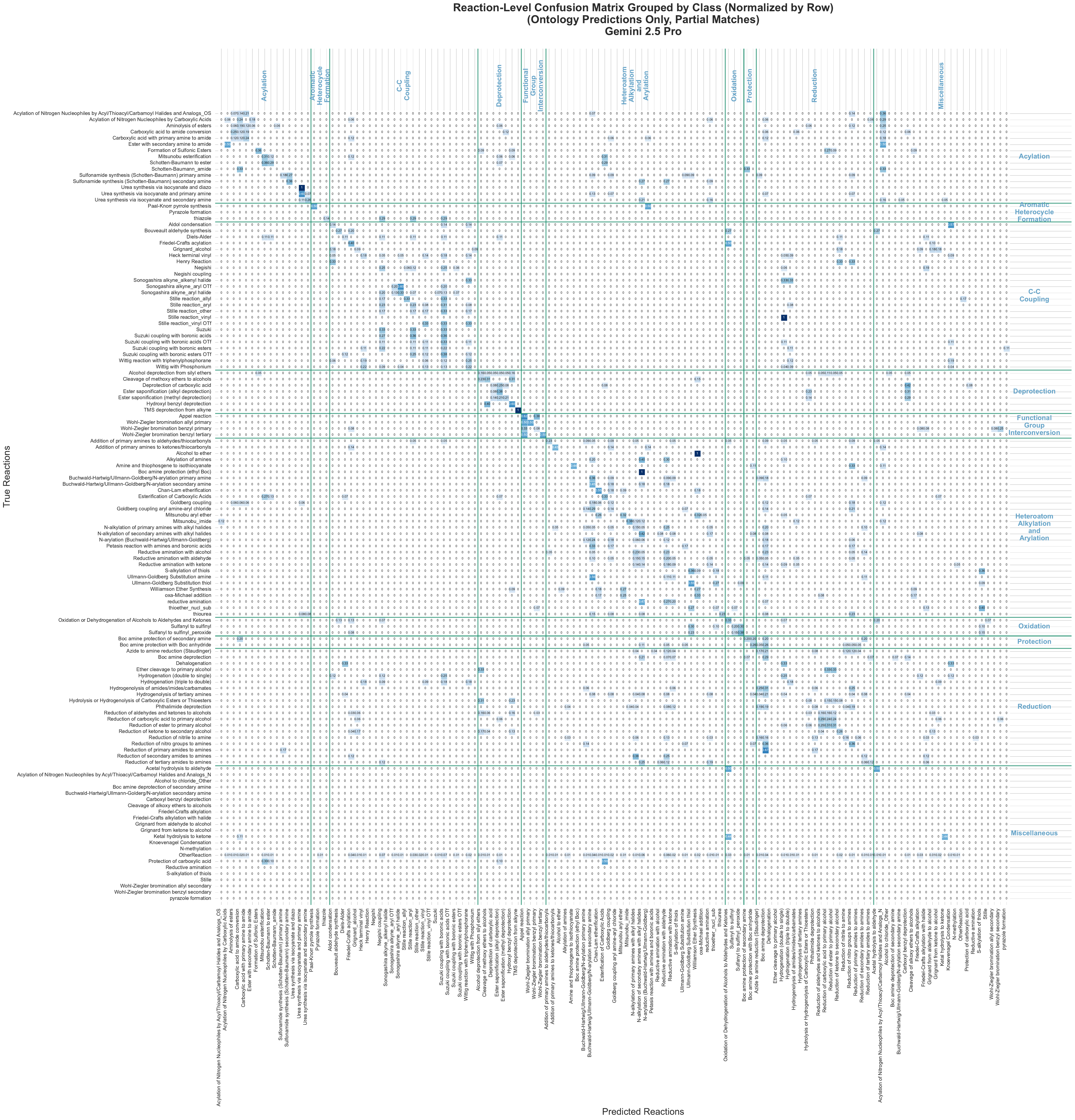}
\caption{Confusion matrix of predicted versus ground-truth reaction names for the Gemini 2.5 Pro model. The analysis is conditional, including only predictions where the model successfully identified at least a partial positional match. For this visualization, reactions outside the defined reaction ontology were excluded. The matrix was generated using the original class-to-name mappings from the ground-truth data, with any unassigned reactions grouped into the 'Miscellaneous' category.}
\label{fig:appendix_position_model_confusion_matrix_reaction_names}
\end{figure}

\clearpage
\subsection{Transition Model}


\begin{table}[h!]
\caption{\small A comparison of model performance on the transition task (reactant prediction). This table presents the total successful predictions, along with accuracy scores for reactants, templates, and the combined category. The best performance in each column is highlighted in bold.}
\label{tab:ablation_transition_model_table}
\centering
\scriptsize
\begin{tabular}{lcccccc}
\toprule
\bfseries Model & \bfseries \shortstack{Avg. number \\ of  predictions}  & \bfseries \shortstack{Reactants \\ accuracy} & \bfseries \shortstack{Template \\ accuracy} & \bfseries \shortstack{Combined \\ accuracy} & \bfseries \shortstack{Total \\ predictions} & \bfseries \shortstack{Failed \\ predictions} \\
\midrule
Ether0 & 0.0 & 0.0 & 0.0 & 0.0 & 0 & 541 \\
Qwen3-4B & 2.9 & 0.11 & 0.05 & 0.13 & 529 & 12 \\
Qwen3-30B & 3.6 & 0.22 & 0.12 & 0.27 & 535 & 6 \\
Qwen3-235B-thinking & 4.3 & 0.54 & 0.28 & 0.59 & 522 & 19 \\
Qwen3-235B-instruct & 6.6 & 0.40 & 0.39 & 0.48 & 537 & 4 \\
Gemini 2.5 Flash & 4.3 & 0.45 & 0.31 & 0.52 & 537 & 4 \\
DeepSeek-R1-670B & 4.3 & 0.62 & 0.28 & 0.69 & 528 & 13 \\
Claude Sonnet 4 & 4.8 & 0.65 & 0.39 & 0.71 & 519 & 0 \\
GPT-5 & 10.2 & 0.68 & 0.44 & 0.73 & 538 & 3 \\
Gemini-2.5-Pro & 5.7 & \textbf{0.75} & \textbf{0.45} & \textbf{0.81} & 540 & 1 \\
\bottomrule
\end{tabular}
\end{table}

\begin{figure*}[h]
\centering
\includegraphics[width=1\textwidth]{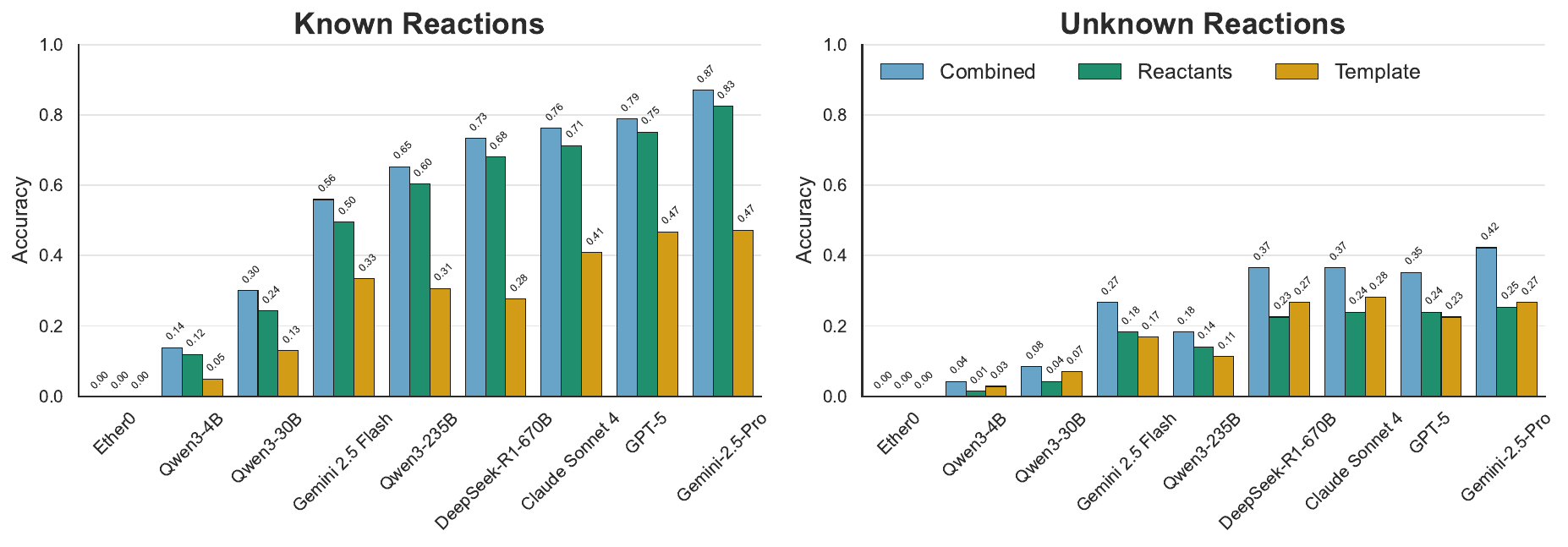}
\caption{\small Performance difference between known and unknown reaction names. For unknown reactions, no equivalent name reaction examples within the \textit{USPTO50k} training dataset are provided, illustrating the importance of the reaction name as a chemical anchor for retrieving reaction examples and chemical reasoning.}
\label{appendix:transition_known_unknown}
\end{figure*}

\begin{figure}[h]
\centering
\includegraphics[width=0.8\textwidth]{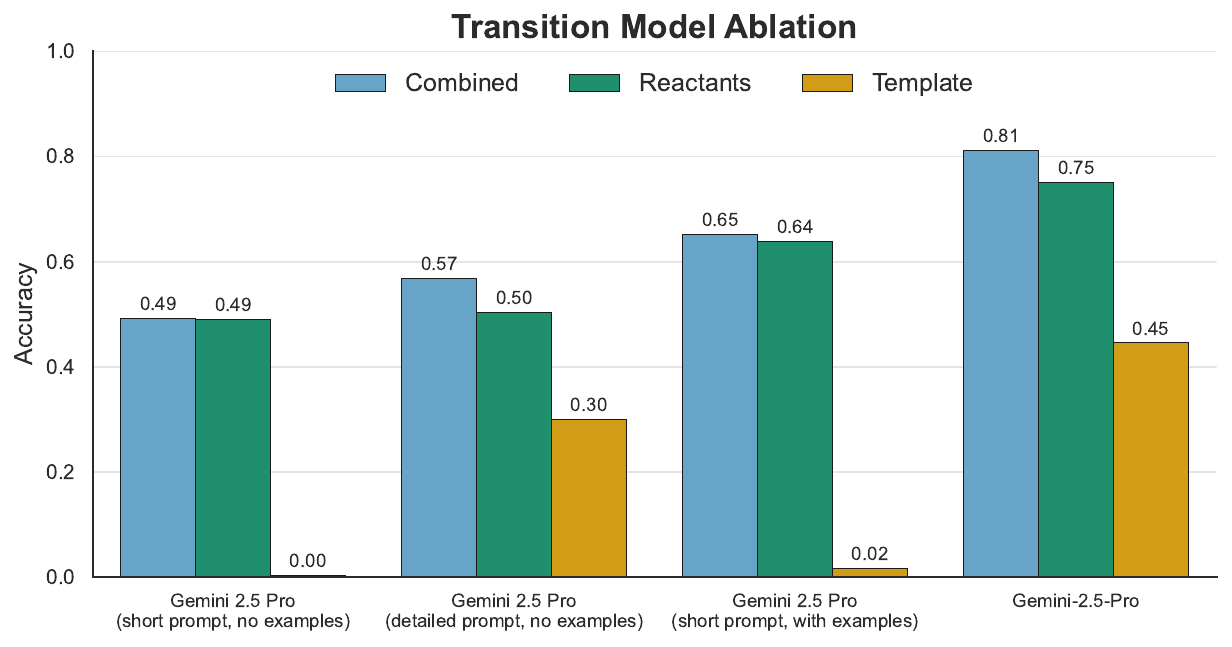}
\caption{\small An ablation study on the impact of prompt instruction detail and the inclusion of in-context examples on the performance of the Gemini 2.5 Pro transition model. We evaluate four settings: 1) a simple prompt without examples (see Prompt \ref{appendix:short_transition_prompt}); 2) a detailed prompt without examples (see Prompt \ref{appendix:transition_prompt}); 3) a simple prompt with examples of molecules synthesized using the same reaction; and 4) a detailed prompt with examples of relevant reactions.}
\label{appendix:fig_transition_prompt_applcation}
\end{figure}

\begin{figure}[h]
\centering
\includegraphics[width=1\textwidth]{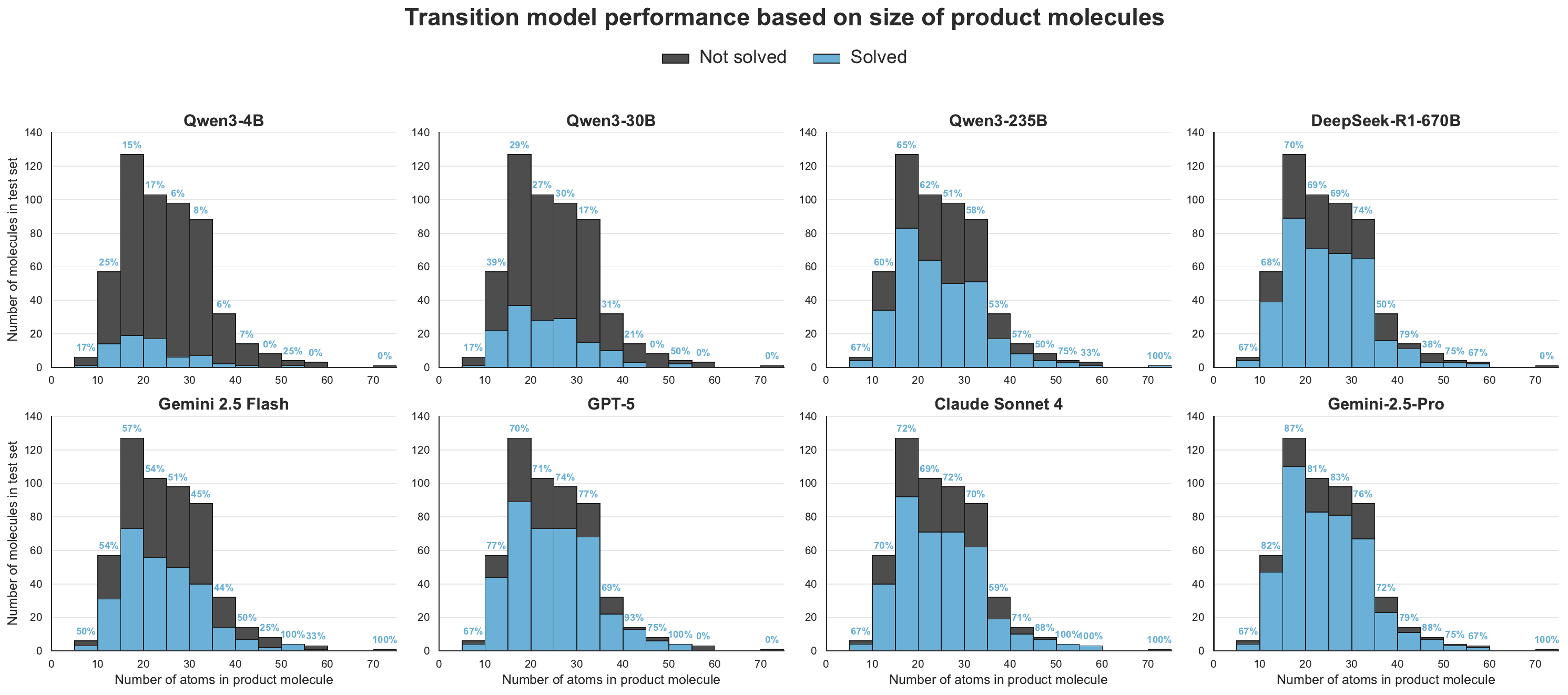}
\caption{Impact of molecule size on transition model performance. The figure displays Reactant Accuracy, stratified by the number of atoms (bin size = 5) across all evaluated LLMs.}
\label{appendix:fig_transition_molsize_vs_performance}
\end{figure}

\begin{figure}[h]
\centering
\includegraphics[width=0.8\textwidth]{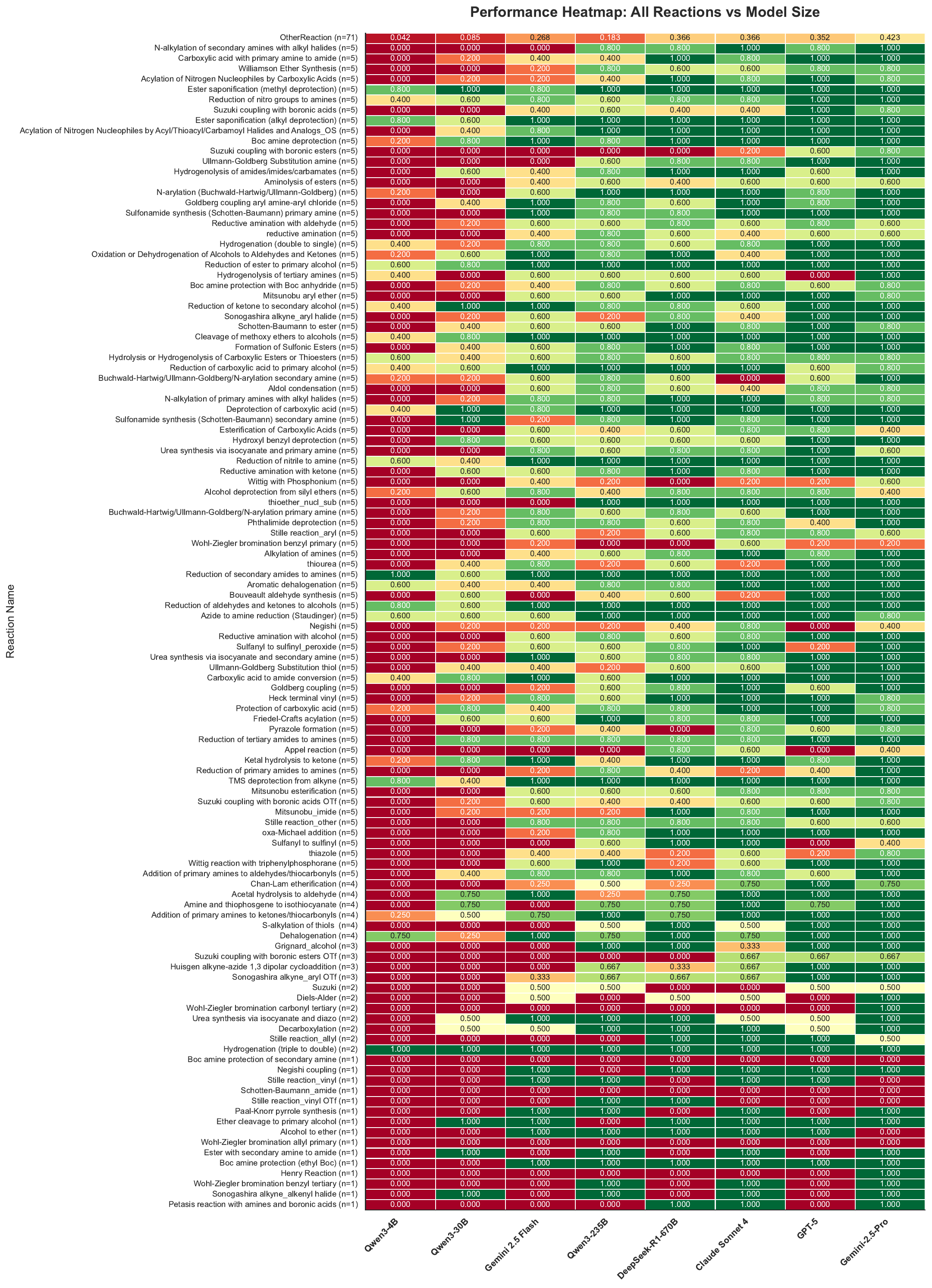}
\caption{Confusion matrix highlighting the performance of different Transition Models on respective reaction name classes using either template or reactant accuracy. The reactions are sorted by the number of reaction examples available in the set (high-to-low).}
\label{fig:appendix_transition_model_confusion_matrix}
\end{figure}

\clearpage
\subsection{Application Examples}

{
\scriptsize
\begin{longtable}{c c p{2cm} c c p{8cm}}
\caption{Predicted Disconnection Sites for LEI-515 \cite{jiangMonoacylglycerolLipaseInhibitor2023}. Header descriptions are as follows: \textbf{Prio.}: Priority Ranking of the Disconnections; \textbf{Position}: The position where the disconnection is; \textbf{Reaction}: The forward reaction; \textbf{Ontology}: If the reaction is in the reaction ontology for which examples are available; \textbf{Imp.}: Retrosynthesis Importance, alignment with retrosynthesis goals; \textbf{Rationale}: The chemical rationale.}
\label{table:appendix_disconnections-lei-515} \\
\toprule
\multicolumn{1}{c}{\textbf{Prio.}} & \multicolumn{1}{c}{\textbf{Position}} & \multicolumn{1}{c}{\textbf{Reaction}} & \multicolumn{1}{c}{\textbf{Ontology}} & \multicolumn{1}{c}{\textbf{Imp.}} & \multicolumn{1}{c}{\textbf{Rationale}} \\
\midrule
\endfirsthead
\toprule
\multicolumn{1}{c}{\textbf{Prio.}} & \multicolumn{1}{c}{\textbf{Position}} & \multicolumn{1}{c}{\textbf{Reaction}} & \multicolumn{1}{c}{\textbf{Ontology}} & \multicolumn{1}{c}{\textbf{Imp.}} & \multicolumn{1}{c}{\textbf{Rationale}} \\
\midrule
\endhead
\bottomrule
\endfoot
1 & C:12 N:14 & Carboxylic acid to amide conversion & Yes & 4 & Identified via Inter-Fragment Analysis, this is a very high-impact, convergent disconnection linking the chiral piperazine amine and the aryl carboxylic acid core, drastically simplifying the molecule (Goals a, c). The forward reaction (amide coupling) is robust and high-yielding (Goal b). No major chemoselectivity issues are expected with modern coupling reagents. \\
2 & C:12 N:14 & Acylation of Nitrogen Nucleophiles by Acyl/Thioacyl/Carbamoyl Halides and Analogs\_N & Yes & 4 &  Identified via Inter-Fragment Analysis, this is a very high-impact, convergent disconnection that simplifies the molecule into two key fragments (Goals a, c). Activation of the carboxylic acid as an acyl chloride is a classic, robust method for amide formation (Goal b), though it may require protection of other nucleophilic sites. \\
3 & N:17 c:18 & Buchwald-Hartwig/Ullmann-Goldberg/N-arylation secondary amine & Yes & 4 &  Identified via Inter-Fragment Analysis, this key convergent C-N bond formation builds the N-arylpiperazine motif (Goal c). The forward reaction, a Buchwald-Hartwig amination, is a powerful and reliable method for this transformation, simplifying the chiral amine precursor (Goals a, b). Potential for competitive reaction at N:14 necessitates a protecting group strategy. \\
4 & N:17 c:18 & N-arylation (Buchwald-Hartwig/Ullmann-Goldberg) & Yes & 4 &  Identified via Inter-Fragment Analysis, this strategic disconnection simplifies the chiral piperazine component by removing the aryl group (Goal a). The forward N-arylation is a powerful and reliable reaction for constructing this key bond (Goals b, c). Regioselectivity between the two piperazine nitrogens is a key issue that must be controlled, likely with a protecting group on N:14. \\
5 & C:5 S:6 & S-alkylation of thiols & Yes & 3 &  Identified via Strategic Bond Analysis. This disconnects the beta-keto sulfoxide side chain from the aryl core (Goal c). The forward synthesis involves S-alkylation of the corresponding thiophenol with an alpha-halo ketone precursor, followed by oxidation. This is a robust way to build this key C-S bond, but the electrophile itself requires separate synthesis (Goals a, b). \\
6 & c:11 C:12 & Grignard with CO2 to carboxylic acid & Yes & 3 &  Identified via FGI Analysis. Strategic C-C bond formation to install the carboxyl group on the central aromatic ring (Goal c). The forward synthesis via carboxylation of an organometallic (e.g., Grignard) from an aryl halide is classic and reliable (Goals a, b). Chemoselectivity is a concern, as the formation of the organometallic must be compatible with the sulfide/sulfoxide group. \\
7 & S:6 O:7 & Sulfanyl to sulfinyl\_H2O2 & Yes & 2 &  Identified via FGI Analysis. Standard FGI to install the sulfoxide from a more stable and easier to handle sulfide precursor. The oxidation can be performed late-stage, but requires careful control of conditions to prevent over-oxidation to the sulfone, which presents a chemoselectivity challenge (Goal b). \\
8 & C:2 F:32 F:33 & Fluorination & Yes & 2 &  Identified via FGI Analysis. This disconnection corresponds to a forward reaction installing the difluoro moiety. Electrophilic fluorination of the beta-keto sulfoxide enolate is a viable route (Goal c). Chemoselectivity could be an issue due to multiple acidic protons (at C:5) and potential for mono- vs di-fluorination, requiring kinetic control. \\
9 & c:30 Cl:31 & Aromatic chlorination & Yes & 2 &  Identified via FGI Analysis. This FGI installs the chloro substituent via electrophilic aromatic substitution (Goal c). The regioselectivity of the chlorination would be directed by the existing sulfoxide/sulfide and carboxylate/amide groups. Predicting and controlling the outcome relative to other open positions on the ring requires careful consideration of the combined directing effects. \\
10 & c:20 Cl:21 & Aromatic chlorination & Yes & 2 &  Identified via FGI Analysis. This FGI installs the chloro substituent on the N-aryl ring via electrophilic aromatic substitution (Goal c). The reaction would be strongly directed by the activating amine substituent, likely leading to the observed para-chlorination, making this a reliable and predictable transformation (Goal b). \\
11 & C:3 O:4 & Oxidation or Dehydrogenation of Alcohols to Aldehydes and Ketones & Yes & 2 &  Identified via FGI analysis. Standard FGI to form the ketone from a secondary alcohol precursor. While many mild oxidation reagents are available, the presence of the easily oxidizable sulfoxide (or its sulfide precursor) on the same molecule presents a major chemoselectivity challenge that must be carefully managed (Goal b). \\
12 & C:12 O:13 N:14 & Nitrile to amide & Yes & 2 &  Identified via FGI analysis. This transforms the amide into a nitrile precursor, offering an alternative synthetic route to the central aromatic core (Goal a). A nitrile can be introduced via methods like the Sandmeyer reaction. The forward reaction, partial hydrolysis of the nitrile to the amide, can be challenging to stop without proceeding to the carboxylic acid. \\
13 & N:14 & Boc amine deprotection & Yes & 1 &  Identified via Protecting Group Analysis. This is a tactical deprotection step. A protecting group like Boc on N:14 would be crucial in a forward synthesis to ensure regioselective N-arylation at N:17. This step reveals the nucleophilic amine for the final amide coupling and is a common, practical consideration (Goal d). \\
14 & C:2 C:3 & Enolate Acylation & No & 3 &  Identified via Strategic Bond Analysis. This strategic C-C bond disconnection breaks down the beta-keto side chain (Goal a). The forward reaction, likely an enolate acylation, is a powerful method for ketone synthesis (Goal c). However, generating and controlling the reactivity and stability of the required difluoroenolate precursor could be challenging. \\
\end{longtable}
}

{
\scriptsize
\begin{longtable}{c p{1.8cm} p{2cm} c c p{8cm}}
\caption{Predicted Disconnection Sites for LEI-401 \cite{mockDiscoveryNAPEPLDInhibitor2020}. Header descriptions are as follows: \textbf{Prio.}: Priority Ranking of the Disconnections; \textbf{Position}: The position where the disconnection is; \textbf{Reaction}: The forward reaction; \textbf{Ontology}: If the reaction is in the reaction ontology for which examples are available; \textbf{Imp.}: Retrosynthesis Importance, alignment with retrosynthesis goals; \textbf{Rationale}: The chemical rationale.}
\label{table:appendix_disconnections-lei-401} \\
\toprule
\multicolumn{1}{c}{\textbf{Prio.}} & \multicolumn{1}{c}{\textbf{Position}} & \multicolumn{1}{c}{\textbf{Reaction}} & \multicolumn{1}{c}{\textbf{Ontology}} & \multicolumn{1}{c}{\textbf{Imp.}} & \multicolumn{1}{c}{\textbf{Rationale}} \\
\midrule
\endfirsthead
\toprule
\multicolumn{1}{c}{\textbf{Prio.}} & \multicolumn{1}{c}{\textbf{Position}} & \multicolumn{1}{c}{\textbf{Reaction}} & \multicolumn{1}{c}{\textbf{Ontology}} & \multicolumn{1}{c}{\textbf{Imp.}} & \multicolumn{1}{c}{\textbf{Rationale}} \\
\midrule
\endhead
\bottomrule
\endfoot
1 & C:4 N:5 & Buchwald-Hartwig/Ullmann-Goldberg/N-arylation secondary amine & Yes & 4 & Identified from Inter-Fragment Analysis (C), this is a very high importance disconnection that convergently couples the phenylpiperidine fragment to the central imidazole core. This modern cross-coupling reaction is robust and strategically sound for scaffold construction (goal c). The synthesis would require a di-halogenated imidazole, and chemoselectivity between the two coupling sites would need to be controlled, possibly via differential reactivity of the halides (e.g., Br vs. I). \\
2 & C:18 N:19 & Buchwald-Hartwig/Ullmann-Goldberg/N-arylation secondary amine & Yes & 4 & Identified from Inter-Fragment Analysis (C), this is a key convergent disconnection of the hydroxypiperidine fragment. A C-N cross-coupling is a powerful method for building the core scaffold (goal c). The free hydroxyl group on the piperidine fragment might require protection to prevent interference with the palladium catalyst, a potential chemoselectivity issue. \\
3 & C:1 N:26 & Carboxylic acid to amide conversion & Yes & 3 & Identified via Strategic Bond Analysis (D), this amide bond disconnection is a classic, high-importance step. The forward reaction is a robust and high-yielding amide coupling, simplifying the molecule to a carboxylic acid precursor and commercially available cyclopropylmethylamine (goals a, b). The secondary amines on the piperidine rings are significantly less nucleophilic, so chemoselectivity should be high. \\
4 & C:1 N:26 & Acylation of Nitrogen Nucleophiles by Acyl/Thioacyl/Carbamoyl Halides and Analogs\_N & Yes & 3 & Identified via Strategic Bond Analysis (D), this is an alternative high-importance disconnection for the amide bond via a more reactive acyl chloride. This reaction is often very fast and high-yielding (goal b), though it requires an extra step to prepare the acyl chloride from the acid. Chemoselectivity is generally excellent. \\
5 & C:21 O:22 & Reduction of ketone to secondary alcohol & Yes & 3 & Identified from Stereochemical Analysis (F) and FGI Analysis (H), this disconnection allows for the creation of the C21 stereocenter. The forward asymmetric reduction of a ketone precursor is a powerful strategy for stereochemical control (goal e) and is a robust reaction (goal b). This approach offers excellent control over the final product's stereochemistry. \\
6 & C:7 C:8 & Suzuki coupling with boronic acids & Yes & 2 & Identified via Strategic Bond Analysis (D), this C-C bond disconnection breaks down a key fragment. However, since chiral 3-phenylpiperidine is accessible, this is less strategic than connecting the whole fragment to the core. A Suzuki coupling would be a reliable method (goal b) but adds steps compared to using the intact piperidine. \\
7 & N:5 C:6 C:7 C:14 C:15 C:16 & Arene hydrogenation & Yes & 2 & Identified via FGI Analysis (H.i), this disconnection simplifies the 3-phenylpiperidine starting material to 3-phenylpyridine. The forward hydrogenation of a pyridine derivative is a common way to access piperidines (goal a). Asymmetric hydrogenation conditions could potentially be employed to set the C7 stereocenter (goal e). \\
8 & N:19 C:20 C:21 C:23 C:24 & Arene hydrogenation & Yes & 2 & Identified via FGI Analysis (H.i), this disconnection simplifies the 3-hydroxypiperidine fragment to 3-hydroxypyridine. While this simplifies the starting material (goal a), controlling the subsequent reduction of the ketone (formed from the hydroxyl) and setting the stereocenter would be a separate, critical step. \\
9 & O:22 & Alcohol deprotection from silyl ethers & Yes & 1 & Identified from Protecting Group Analysis (I), this represents a tactical deprotection step. The alcohol would likely need to be protected as a silyl ether during steps involving strong bases or organometallic reagents to avoid side reactions. This step addresses chemoselectivity but is of lower strategic importance. \\
10 & N:5 & Boc amine deprotection & Yes & 1 & Identified from Protecting Group Analysis (I), this is a tactical deprotection. The secondary amine of the piperidine may require Boc protection to prevent it from interfering in other reactions, such as the second C-N coupling. This step manages chemoselectivity and is of lower strategic importance. \\
11 & N:19 & Boc amine deprotection & Yes & 1 & Identified from Protecting Group Analysis (I), this is another tactical deprotection step. Protecting this secondary amine could be crucial for achieving selectivity during a stepwise C-N coupling sequence on the imidazole core. It is a key step for controlling chemoselectivity. \\
\end{longtable}
}

\begin{figure}[h]
\centering
\includegraphics[width=0.5\textwidth]{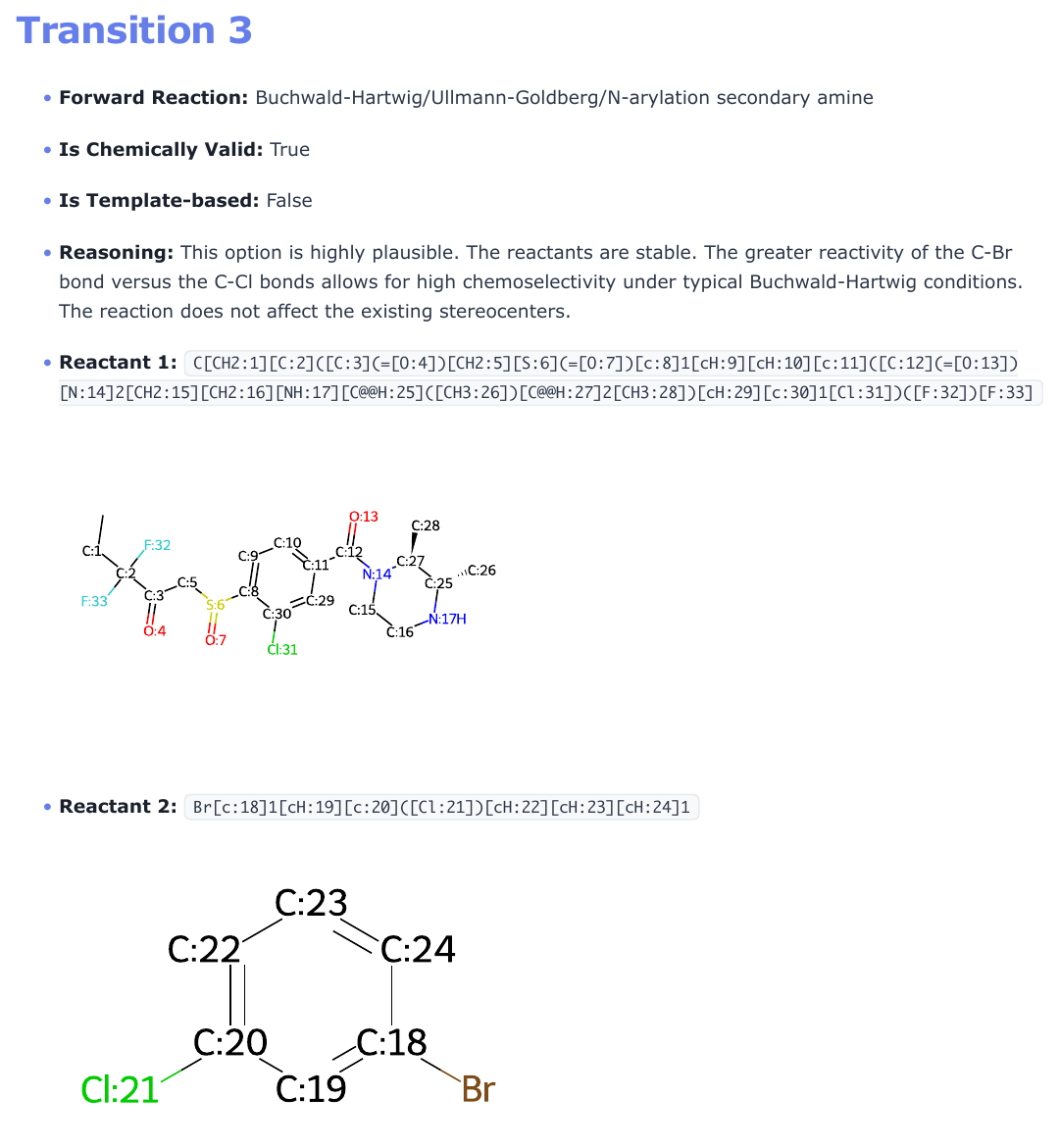}
\caption{Correct reactant prediction for LEI-515 \cite{jiangMonoacylglycerolLipaseInhibitor2023} by the Transition model (position priority 3, transition prediction 3).}
\label{appendix:application_lei_515_correct}
\end{figure}

\begin{figure}[h]
\centering
\includegraphics[width=0.5\textwidth]{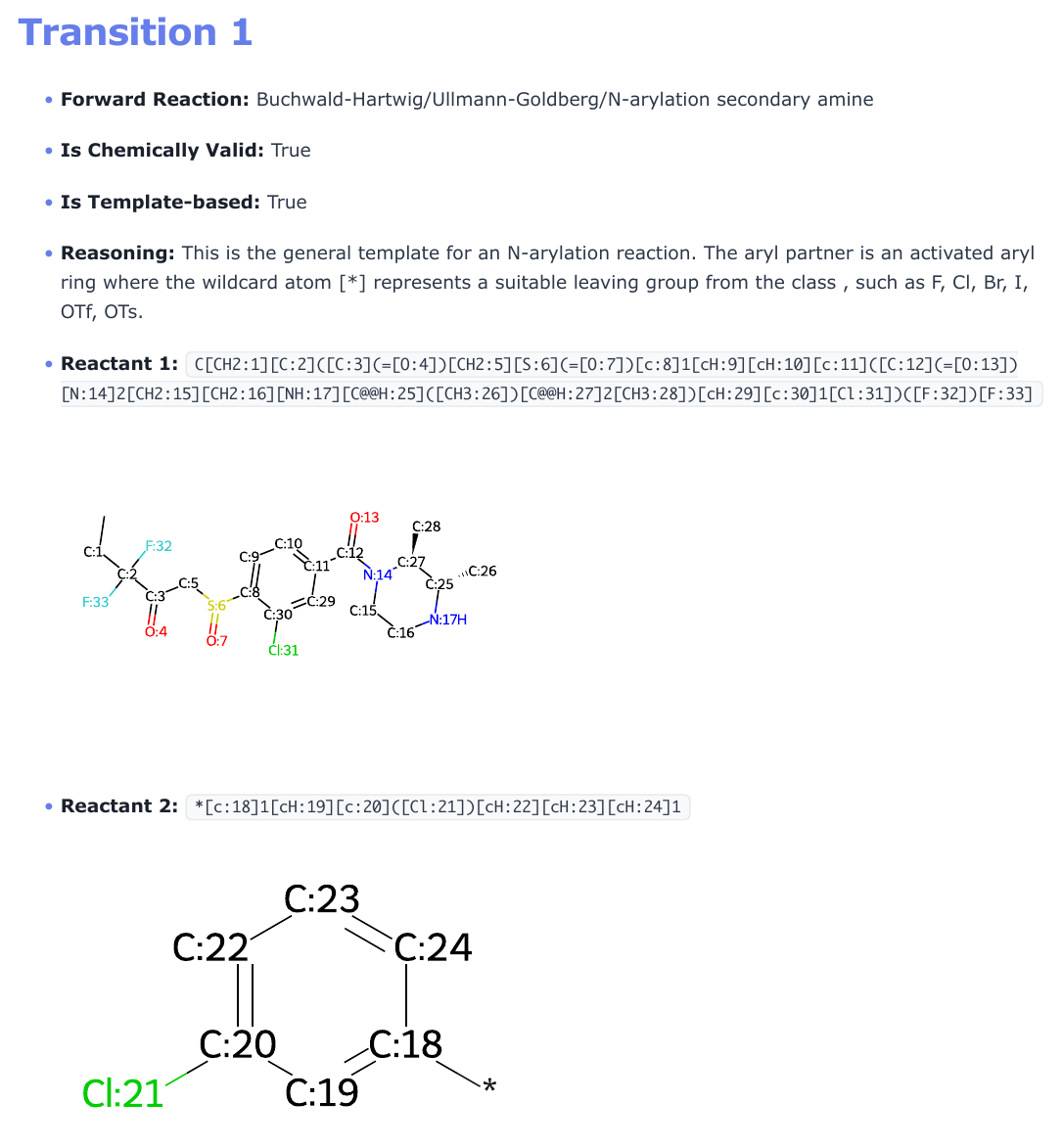}
\caption{Correct reactant template prediction for LEI-515 \cite{jiangMonoacylglycerolLipaseInhibitor2023} by the Transition model (position priority 3, transition prediction 1).}
\label{appendix:application_lei_515_correct_template}
\end{figure}

\begin{figure}[h]
\centering
\includegraphics[width=0.5\textwidth]{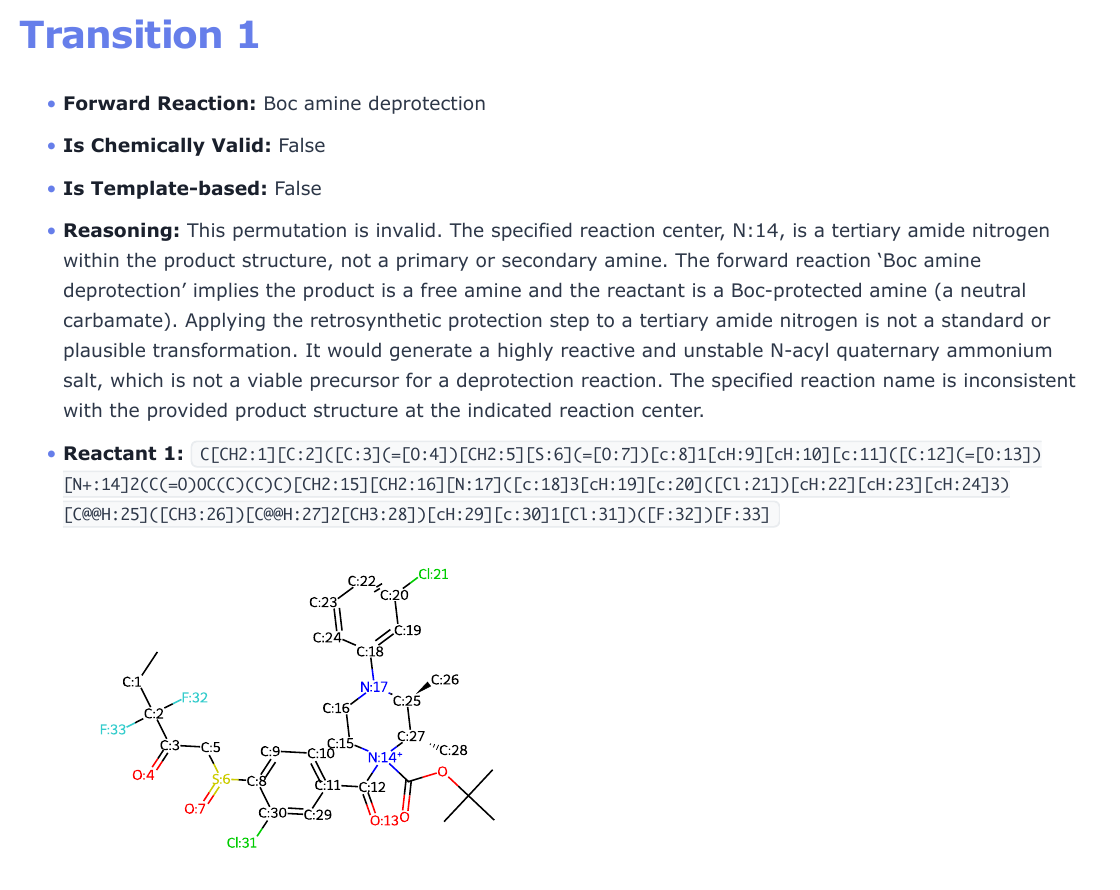}
\caption{Correctly flagged chemically invalid prediction for LEI-515 \cite{jiangMonoacylglycerolLipaseInhibitor2023} by the Transition Model (position priority 13, transition prediction 1).}
\label{appendix:application_lei_515_correct_chemically_invalid}
\end{figure}

\begin{figure}[h]
\centering
\includegraphics[width=0.5\textwidth]{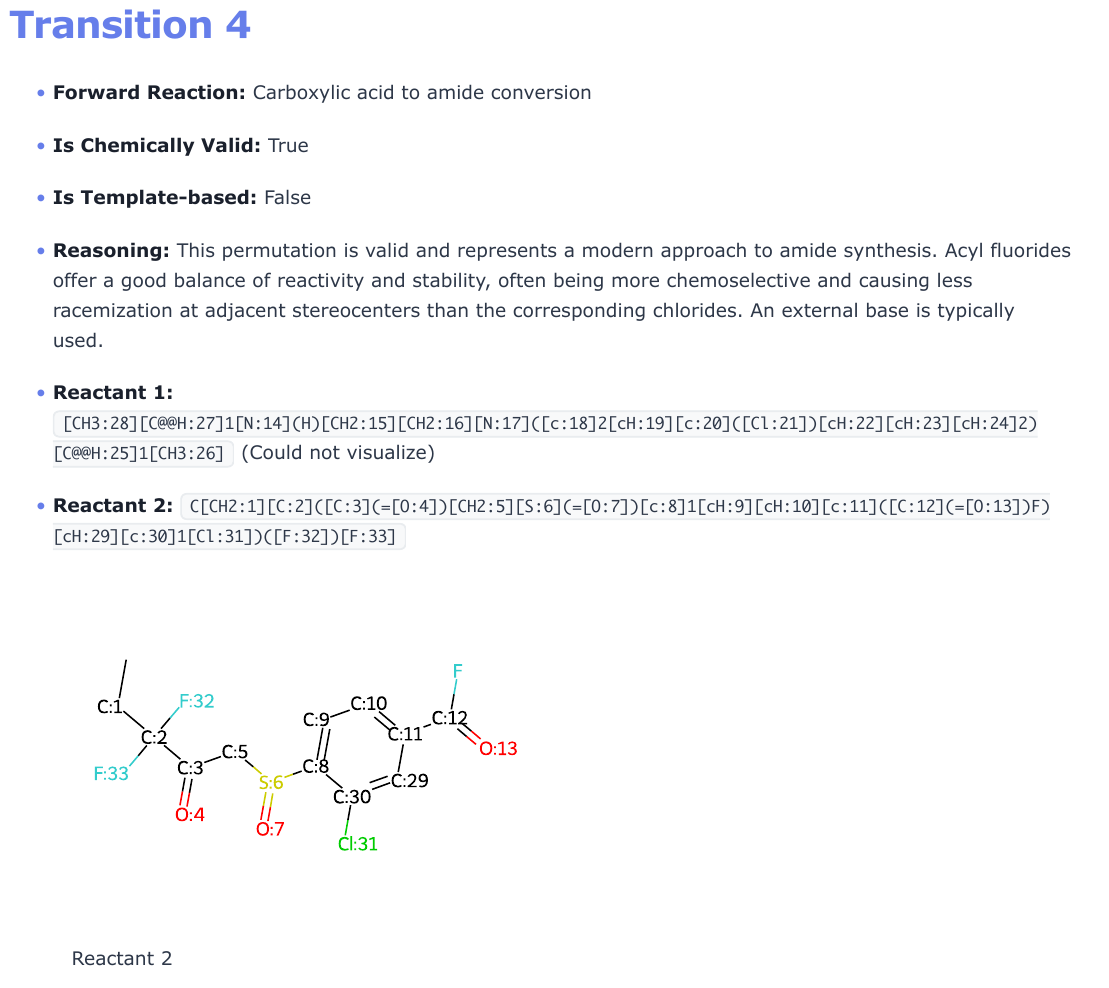}
\caption{Syntactically invalid SMILES prediction for LEI-515 \cite{jiangMonoacylglycerolLipaseInhibitor2023} by the Transition model (position priority 1, transition prediction 4).}
\label{appendix:application_lei_515_correct_smiles_broken}
\end{figure}

\clearpage
\subsection{Application Questionaire}
\begin{table}[h]
\caption{Full list of questions for the expert validation study. These are the complete, verbatim questions presented to chemists to benchmark the performance of our framework. The evaluation was split into two parts: assessing the disconnection sites proposed by the Position (P) model and the final reactant structures generated by the Transition (T) model.}
\label{appendix:full_questions}
\scriptsize
\begin{tabular}{l p{0.8\textwidth}} 
\toprule
\multicolumn{1}{l}{\bf Q.}  &\multicolumn{1}{l}{\bf Description}
\\ \midrule \\
P1     & Is the suggested disconnection position chemically plausible (i.e., not violating fundamental principles)? \\
P2     & Is the suggested reaction name a correct label for the proposed disconnection position? \\
P3     & Is the provided chemical reasoning for the suggested disconnection (position and reaction name) scientifically sound? \\
P4     & Considering all the provided information, could this suggested step realistically work in a laboratory setting? \\
P5     & Has this specific transformation actually been performed successfully in practice for the molecule? \\
P6     & Are there any strategically important disconnections that are obviously missing from this prediction? \\
\\ \midrule \\
T1     & Given the transition prediction includes a reaction template, does the reaction template capture the overall chemical transformation of the reaction? \\
T2     & Given the transition prediction includes a reaction template, does the chemical reasoning for the reaction template align with the underlying reaction? \\
T3     & Among the reactant predictions, is there at least one that provides a chemically correct set of reactants to form the target product? \\
T4     & If the model predicts a chemically correct set of reactants, is the model's chemical reasoning for that specific set of reactants correct? \\
T5     & If the reaction was conducted in the lab, does the model correctly predict the set of reactants that were used in the lab? \\
T6     & If the model flags one of its own predictions as 'chemically invalid', is its reasoning for that assessment correct? \\
T7 & How many reactants are predicted as chemically valid and are not reaction templates are correct? \\
\\ \bottomrule \\
\end{tabular}
\end{table}

\begin{table}[h]
\caption{Detailed response data.}
\label{appendix:application_questionaire_responses}
\scriptsize
\begin{tabular}{llllllllllllll}
\toprule
\multicolumn{1}{l}{\bf ID}  &\multicolumn{1}{l}{\bf P1} &\multicolumn{1}{l}{\bf P2} &\multicolumn{1}{l}{\bf P3} &\multicolumn{1}{l}{\bf P4} &\multicolumn{1}{l}{\bf P5} &\multicolumn{1}{l}{\bf P6} &\multicolumn{1}{l}{\bf T1} &\multicolumn{1}{l}{\bf T2} &\multicolumn{1}{l}{\bf T3} &\multicolumn{1}{l}{\bf T4} &\multicolumn{1}{l}{\bf T5} &\multicolumn{1}{l}{\bf T6} &\multicolumn{1}{l}{\bf T7} 
\\ \midrule
DH376 \cite{dengTriazoleUreasAct2017} & 12/13 & 11/13 & 8/13 & 11/13 & 5/13 & 1     & 4/4 & 4/4 & 4/4 & 4/4 & 3/4 & 2/2 & 23/31 \\
LEI-102 \cite{liStructuralBasisSelective2023} & 14/16 & 12/16 & 12/16 & 14/16 & 2/16 & 1   & 3/3 & 3/3 & 4/4 & 4/4 & 4/4 & 3/3 & 18/18 \\
LEI-105 \cite{baggelaarHighlySelectiveReversible2015} & 8/9 & 8/9 & 8/9 & 5/9 & 2/9 & 1            & 2/2 & 2/2 & 2/2 & 2/2 & 1/1 & - & 11/11\\
LEI-401 \cite{mockDiscoveryNAPEPLDInhibitor2020} & 11/11 & 11/11 & 7/11 & 11/11 & 2/11 & 0    & 2/3 & 3/3 & 3/3 & 3/3 & 2/2 & 1/1 & 10/15 \\
LEI-515 \cite{jiangMonoacylglycerolLipaseInhibitor2023} & 12/14 & 12/14 & 11/14 & 8/14 & 5/14 & 1    & 2/4 & 2/4 & 4/6 & 4/6 & 1/4 & 1/1 & 11/23 \\
 \hline
Acc. & 90.5 & 85.7 & 73.0 & 77.8 &  25.4 & 80.0     & 81.3 & 87.5 & 89.5 & 89.5 & 73.3 & 1 & 74.5 \\
\bottomrule
\end{tabular}
\end{table}
\clearpage
\subsection{New Drug Modalities}
\subsubsection{TRAP-1 - Molecular Glue}

\begin{figure}[h]
\centering
\includegraphics[width=0.8\textwidth]{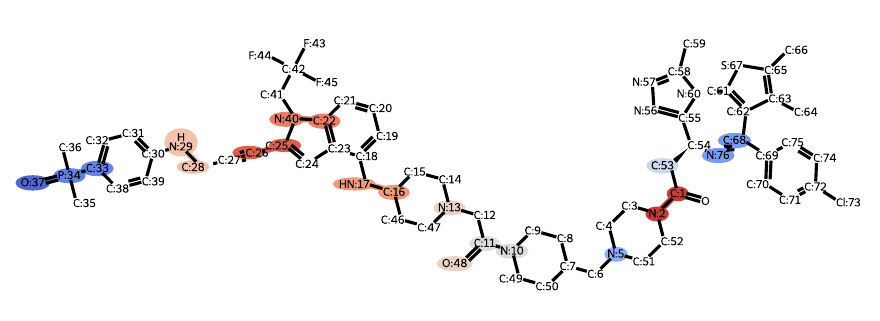}
\caption{Position model prediction for TRAP-1 \cite{zhuActivatingP53Y220CMutantSpecific2024} molecular glue using Gemini 2.5 Pro and PaRoutes reaction ontology}
\label{appendix:trap_1}
\end{figure}

{\scriptsize
\begin{longtable}{c c p{3cm} c c p{8cm}}
\caption{Predicted Disconnection Sites for TRAP-1 \cite{zhuActivatingP53Y220CMutantSpecific2024} using Gemini 2.5 Pro and PaRoutes reaction ontology. Header descriptions are as follows: \textbf{Prio.}: Priority Ranking of the Disconnections; \textbf{Position}: The position where the disconnection is; \textbf{Reaction}: The forward reaction; \textbf{Ontology}: If the reaction is in the reaction ontology for which examples are available; \textbf{Imp.}: Retrosynthesis Importance, alignment with retrosynthesis goals; \textbf{Rationale}: The chemical rationale.}
\label{table:appendix_trap_1} \\
\toprule
\multicolumn{1}{c}{\textbf{Prio.}} & \multicolumn{1}{c}{\textbf{Position}} & \multicolumn{1}{c}{\textbf{Reaction}} & \multicolumn{1}{c}{\textbf{Ontology}} & \multicolumn{1}{c}{\textbf{Imp.}} & \multicolumn{1}{c}{\textbf{Rationale}} \\
\midrule
\endfirsthead
\toprule
\multicolumn{6}{c}{{\bfseries \tablename\ \thetable{} -- continued from previous page}} \\
\midrule
\multicolumn{1}{c}{\textbf{Prio.}} & \multicolumn{1}{c}{\textbf{Position}} & \multicolumn{1}{c}{\textbf{Reaction}} & \multicolumn{1}{c}{\textbf{Ontology}} & \multicolumn{1}{c}{\textbf{Imp.}} & \multicolumn{1}{c}{\textbf{Rationale}} \\
\midrule
\endhead
\hline \multicolumn{6}{r}{{Continued on next page}} \\ \bottomrule
\endfoot
\hline
\endlastfoot

1 & C:1 N:2 & Carboxylic acid to amide conversion & Yes & 4 & Very high importance disconnection of a key amide bond, convergently linking the fully assembled 'warhead-diazepine' fragment to the linker-ligase binder fragment. This step, derived from Inter-Fragment Analysis (C), is a robust transformation that greatly simplifies the overall synthesis (a, c). Potential chemoselectivity issues with other nucleophilic amines would require a carefully planned protecting group strategy. \\
2 & C:25 C:26 & Sonogashira alkyne\_aryl halide & Yes & 4 & A critical C-C bond disconnection that convergently joins the carbazole core with the alkyne-phosphine oxide fragment. Identified via Inter-Fragment Analysis (C), this Sonogashira coupling is a powerful and reliable method for strategically building the E3 ligase binder scaffold (b, c). The reaction is generally high-yielding and tolerant of many functional groups. \\
3 & c:22 n:40 & Buchwald-Hartwig/ Ullmann-Goldberg/ N-arylation secondary amine & Yes & 4 & A very high importance ring-closing disconnection that forms the core carbazole heterocycle of the E3 ligase binder. Identified via Intra-Fragment Analysis (E), this intramolecular N-arylation is a powerful strategy for building this key scaffold from a biaryl amine precursor, greatly simplifying the synthesis (a, c). \\
4 & C:16 N:17 & Reductive amination with aldehyde & Yes & 3 & A high-importance disconnection linking the complex poly-amine linker to the carbazole-based E3 ligase binder. This reductive amination, identified via Inter-Fragment Analysis (C), is a robust and efficient method for forming C-N bonds (b, c). The aldehyde precursor on the linker would need to be synthesized or unmasked just prior to the coupling. \\
5 & C:16 N:17 & N-alkylation of primary amines with alkyl halides & Yes & 3 & Alternative high-importance strategy to connect the linker and ligase binder via nucleophilic substitution. Identified through Inter-Fragment Analysis (C), this approach offers a reliable C-N bond formation (b, c). However, it may face challenges with over-alkylation and requires an activated halide precursor, making reductive amination often preferable for complex substrates. \\
6 & C:28 N:29 & Buchwald-Hartwig/ Ullmann-Goldberg/ N-arylation primary amine & Yes & 3 & A key disconnection of an aryl-amine bond, attaching the sidechain to the phosphine oxide-bearing ring. This Buchwald-Hartwig amination, identified via Inter-Fragment Analysis (C), is a powerful tool for constructing this bond (b, c). The reaction requires careful optimization of catalyst, ligand, and base to avoid side reactions with other functional groups. \\
7 & C:48 N:13 & Carboxylic acid to amide conversion & Yes & 3 & A strategic amide bond disconnection within the linker structure. Derived from Strategic Bond Analysis (D), this step breaks the linker into two smaller, more manageable fragments, facilitating a modular and convergent assembly (a, c). Standard peptide coupling conditions are expected to be effective. \\
8 & C:11 N:10 & Carboxylic acid to amide conversion & Yes & 3 & A key amide bond disconnection that partitions the complex linker. This approach, from Strategic Bond Analysis (D), allows for a stepwise, controlled assembly of the linker from smaller building blocks (a, c). The presence of multiple amine nucleophiles necessitates an orthogonal protecting group strategy for a successful synthesis. \\
9 & C:53 N:2 & N-alkylation of secondary amines with alkyl halides & Yes & 3 & An alternative high-importance strategy for linking the warhead to the linker. Identified via Strategic Bond Analysis (D), this disconnection leads to an activated alkyl halide on the warhead and the free amine of the linker. This C-N bond formation (c) could be viable if the amide connection proves difficult, but requires careful control to prevent side reactions. \\
10 & N:17 & Reduction of nitro groups to amines & Yes & 2 & A standard Functional Group Interconversion (FGI) step, deriving the key aniline nitrogen from a nitro group precursor. Identified via FGI Analysis (H), this is a robust transformation (b). The nitro group serves as a masked amine and can influence the reactivity of the aromatic ring during earlier synthetic steps before being reduced for linker attachment. \\
11 & N:29 & Reduction of nitro groups to amines & Yes & 2 & A common Functional Group Interconversion (FGI) identified via FGI Analysis (H), where the aniline is derived from reduction of a nitro group. This is a very reliable reaction that allows the use of nitro-group chemistry (e.g., directing effects in EAS) earlier in the synthesis of the phosphine oxide-bearing fragment (a, b). \\
12 & N:5 & Boc amine deprotection & Yes & 1 & A necessary deprotection step to reveal a reactive amine within the linker for subsequent elaboration. Identified via Protecting Group Analysis (I), this step is crucial for the sequential, controlled construction of the linker (c). While of lower strategic importance for bond formation, it is of high practical importance for the overall synthetic route's success. \\
13 & C:68 N:76 & Intramolecular Imine Formation & No & 4 & This represents the key intramolecular ring-closing step to form the seven-membered diazepine ring of the warhead. Derived from Intra-Fragment Analysis (E), this disconnection breaks the core scaffold down to a more flexible linear precursor, which simplifies installation of the chiral center C:54 (a, c, e). This is a thermodynamically driven condensation. \\
14 & C:33 P:34 & Palladium-catalyzed P-C coupling & No & 3 & High importance disconnection of the aryl C-P bond, which installs the key dimethylphosphine oxide group. This disconnection, from Inter-Fragment Analysis (C), simplifies the aromatic precursor to a simple aryl halide or triflate (a). The forward reaction is a reliable palladium-catalyzed coupling of an aryl halide with a P(V) species like H-P(O)Me2 (b, c). \\
15 & P:34 O:37 & Phosphine Oxidation & No & 2 & A Functional Group Interconversion (FGI) where the phosphine oxide is formed by oxidation of the corresponding tertiary phosphine. This step, from FGI Analysis (H), is often performed late in the synthesis as the precursor phosphine can act as a ligand and be poisoned in preceding metal-catalyzed coupling steps (c). The oxidation itself is typically straightforward and high-yielding. \\
\end{longtable}
}
\subsubsection{MCL-1 compound 25 - Macrocycle}

\begin{figure}[h]
\centering
\includegraphics[width=0.8\textwidth]{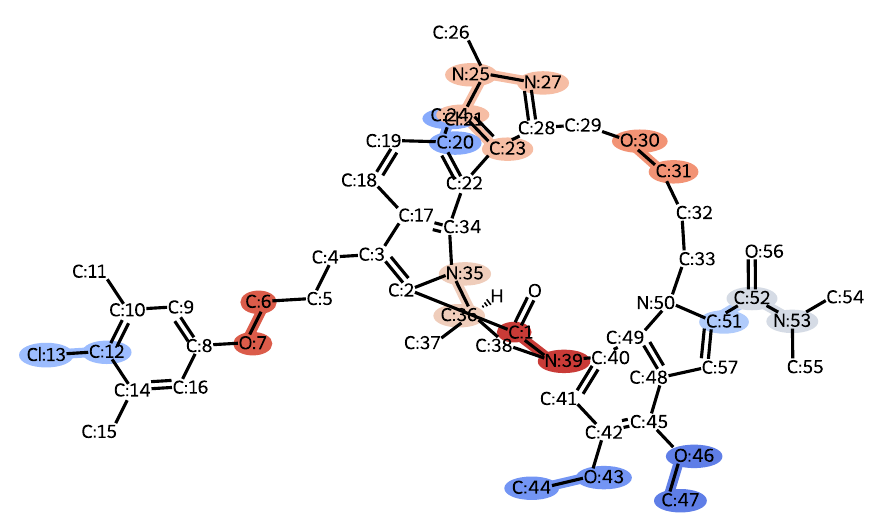}
\caption{Position model prediction for MCL-1 compound 25 \cite{tarrDiscoveryMacrocyclicMyeloid2025} Macrocycle using Gemini 2.5 Pro and PaRoutes reaction ontology}
\label{appendix:fig_mcl_1}
\end{figure}

{
\scriptsize

\begin{longtable}{c c p{3cm} c c p{8cm}}
\caption{Predicted Disconnection Sites for MCL-1 compound 25 \cite{tarrDiscoveryMacrocyclicMyeloid2025} using Gemini 2.5 Pro and PaRoutes reaction ontology. Header descriptions are as follows: \textbf{Prio.}: Priority Ranking of the Disconnections; \textbf{Position}: The position where the disconnection is; \textbf{Reaction}: The forward reaction; \textbf{Ontology}: If the reaction is in the reaction ontology for which examples are available; \textbf{Imp.}: Retrosynthesis Importance, alignment with retrosynthesis goals; \textbf{Rationale}: The chemical rationale.}
\label{table:appendix_mcl_1_cmpd_25} \\
\hline
\multicolumn{1}{c}{\textbf{Prio.}} & \multicolumn{1}{c}{\textbf{Pos.}} & \multicolumn{1}{c}{\textbf{Reaction}} & \multicolumn{1}{c}{\textbf{Ont.}} & \multicolumn{1}{c}{\textbf{Imp.}} & \multicolumn{1}{c}{\textbf{Rationale}} \\
\hline
\endfirsthead
\multicolumn{6}{c}{{\bfseries \tablename\ \thetable{} -- continued from previous page}} \\
\hline
\multicolumn{1}{c}{\textbf{Prio.}} & \multicolumn{1}{c}{\textbf{Pos.}} & \multicolumn{1}{c}{\textbf{Reaction}} & \multicolumn{1}{c}{\textbf{Ont.}} & \multicolumn{1}{c}{\textbf{Imp.}} & \multicolumn{1}{c}{\textbf{Rationale}} \\
\hline
\endhead
\hline \multicolumn{6}{r}{{Continued on next page}} \\ \hline
\endfoot
\hline
\endlastfoot

1 & C:1 N:39 & Carboxylic acid to amide conversion & Yes & 4 & Very high importance disconnection forming the central seven-membered diazepinone ring via a reliable intramolecular lactamization. (Analysis E) This is a key strategic step for constructing the core scaffold (c). The reaction is robust, but requires high-dilution conditions to prevent intermolecular polymerization (d). \\
2 & C:6 O:7 & Williamson Ether Synthesis & Yes & 4 & Very high importance disconnection that convergently couples the complex core scaffold to a simpler, readily available 4-chloro-2,5-dimethylphenol fragment. (Analysis C) This approach greatly simplifies the overall synthesis (a). Potential chemoselectivity issues between N- and O-alkylation on the precursor must be addressed by careful choice of base and conditions. \\
3 & C:6 O:7 & Mitsunobu aryl ether & Yes & 4 & An alternative very high importance convergent disconnection linking the core to the phenol side chain. (Analysis C) The Mitsunobu reaction proceeds under mild conditions but suffers from poor atom economy and uses hazardous reagents, impacting its practicality for large-scale synthesis (d). \\
4 & C:31 O:30 & Williamson Ether Synthesis & Yes & 4 & Very high importance disconnection that convergently assembles the molecule by coupling the indole core with the pyrazole-containing side chain. (Analysis C) This is a robust and strategic C-O bond formation (b, c) that breaks the molecule into two large, more manageable synthons (a). \\
5 & C:31 O:30 & Mitsunobu aryl ether & Yes & 4 & An alternative very high importance disconnection for coupling the indole and pyrazole fragments. (Analysis C) This offers a mild route for the ether formation but is less atom-economical than the Williamson synthesis, which is a key consideration for efficiency (d). \\
6 & c:23 n:27 n:25 c:24 & Pyrazole formation & Yes & 4 & Very high importance multi-bond disconnection that constructs the pyrazole ring in a single, powerful step from acyclic precursors. (Analysis J) This is a classic, high-yielding heterocycle synthesis (b) that dramatically simplifies one of the key fragments (a, c). \\
7 & N:35 C:36 & Reductive amination with ketone & Yes & 3 & High importance disconnection that simplifies the core scaffold by opening the diazepinone ring. (Analysis D) This strategy creates a linear precursor and provides a direct pathway to install the C:36 stereocenter via an asymmetric variant of the forward reaction (c, e). \\
8 & C:36 & Reductive amination with ketone & Yes & 3 & High importance transformation focused on creating the molecule's sole stereocenter. (Analysis F) An asymmetric reductive amination or the reduction of the corresponding imine is a powerful strategy for establishing the required stereochemistry with high control (e). \\
9 & C:52 N:53 & Carboxylic acid to amide conversion & Yes & 3 & High importance disconnection for the installation of the terminal dimethylamide group. (Analysis H) This is a robust and extremely common transformation (b), coupling a carboxylic acid precursor with dimethylamine, both of which are simple starting materials (a). \\
10 & C:52 N:53 & Acylation of Nitrogen Nucleophiles by Acyl/Thioacyl/Carbamoyl Halides and Analogs\_N & Yes & 3 & An alternative high importance disconnection for forming the dimethylamide. (Analysis H) Using an activated acyl chloride precursor is a highly reliable and efficient method for this acylation, representing a key functional group installation (c, d). \\
11 & c:51 C:52 & Friedel-Crafts acylation & Yes & 3 & High importance disconnection creating a key C-C bond by attaching the amide sidechain to the indole core. (Analysis D) This strategy relies on the electrophilic substitution of an electron-rich indole system (c). Regiocontrol could be challenging and may require specific reaction conditions or a pre-functionalized indole. \\
12 & c:12 Cl:13 & Aromatic chlorination & Yes & 2 & Medium importance disconnection for a standard functional group interconversion. (Analysis H) The chlorine atom can be installed via electrophilic aromatic substitution. This is a reliable reaction (b), though regioselectivity would be dictated by other substituents on the ring. \\
13 & c:20 Cl:21 & Aromatic chlorination & Yes & 2 & Medium importance FGI for installing the second chlorine atom. (Analysis H) This transformation would likely occur on an advanced intermediate, and control of regioselectivity would be crucial for the success of this step (c). \\
14 & O:43 C:44 & O-methylation & Yes & 2 & Medium importance disconnection representing a standard functional group interconversion. (Analysis H) Installation of the methoxy group via methylation of a phenol precursor provides synthetic flexibility and starts from a simpler material (a). It is a robust reaction (b). \\
15 & O:46 C:47 & O-methylation & Yes & 2 & Medium importance FGI for the second methoxy group. (Analysis H) Methylating a di-phenol precursor is a common strategy. If sequential methylation is needed, a protecting group strategy would be required to ensure chemoselectivity. \\
16 & N:39 & Boc amine deprotection & Yes & 1 & Lower importance disconnection related to a tactical protecting group strategy. (Analysis I) This implies the use of a Boc group on the amide nitrogen to prevent unwanted side reactions during synthesis. Its removal is a necessary but not a core strategic bond-forming step (d). \\
\end{longtable}

}

\clearpage
\subsection{Prompts}
\lstdefinestyle{mystyle}{
    language=bash, 
    backgroundcolor=\color{black!5},
    basicstyle=\ttfamily\tiny,
    commentstyle=\color{blue!80!black}\bfseries, 
    keywordstyle=\color{black}, 
    stringstyle=\color{red!80!black},
    numberstyle=\tiny\color{gray},
    numbers=left,
    breaklines=true,
    frame=tb,
    captionpos=b,
    tabsize=2
}
\subsubsection{Position Model}
\label{appendix:position_prompt}
\begin{minted}{markdown}
**Persona:**
You are an expert chemist specializing in retrosynthetic analysis.

**Primary Goal:**
Your primary goal is to perform a comprehensive retrosynthetic analysis on a given molecule. You will identify all strategically viable disconnection points, rank them according to the provided framework, and format the entire output as a single, valid JSON object.

**Input Schema:**
- product_smiles: The atom-mapped SMILES string of the product molecule.
- reaction_ontology: The provided JSON object containing the reaction ontology.

**Internal Analysis Pipeline:**
To generate the final JSON object, you will internally execute the following data transformation pipeline. The output of each step serves as the direct input for the next, ensuring a dependent, step-by-step analysis.

1.  **Step 1: Identify All Candidate Transformations**
    Process steps A - L sequentially. For each step, you must perform a complete and independent analysis to identify all transformations that fit its description. A finding in one step does not exclude findings in others.
    * **Input:** The `product_smiles`.
    * **Process:**
        * A) **Symmetry Analysis:** First, assess the molecule for any elements of symmetry. If symmetrical fragments exist, identify transformations that could form the molecule by coupling two identical precursors.
        * B) **Fragment Partitioning:** Mentally partition the molecule into its major constituent fragments. The goal is to find disconnections that lead to a **convergent synthesis**.
        * C) **Inter-Fragment Analysis:** Identify the bonds that **connect these major fragments**. These are candidates for strategic coupling reactions.
        * D) **Strategic Bond Analysis:** Within the identified fragments, specifically look for bonds that are adjacent to functional groups, making them chemically activated and strategic targets for disconnection (e.g., bonds alpha/beta to carbonyls, bonds within key functional groups like amides and esters).
        * E) **Intra-Fragment Analysis:** Within each major fragment, identify bonds that could be strategically formed via an **intramolecular (ring-closing) reaction**.
        * F) **Stereochemical Analysis:** Identify all stereocenters. For each one, consider transformations that could set that stereocenter (e.g., asymmetric reactions, chiral pool approach).
        * G) **Rearrangement Analysis:** Look for structural motifs that could be efficiently formed via a powerful **skeletal rearrangement**.
        * H) **FGI Analysis:** For each functional group in the molecule, systematically identify all possible functional groups that are candidates for standard Functional Group Interconversions. This analysis **must** include, but is not limited to:
            * **i. Oxidation/Reduction:** Identify all groups that could be retrosynthetically derived from a different oxidation state.
            * **ii. Non-Redox FGIs:** Identify all non-redox interconversions. This involves analyzing polar carbon-heteroatom bonds within functional groups that are classically disconnected via substitution or hydrolysis-type mechanisms.
        * I) **Protecting Group Analysis:** Analyze for protecting group strategies by proposing protections for sensitive functional groups or deprotections for existing, recognizable protecting groups. Note that a retrosynthethic protection is a forward deprotection reaction and vice versa.
        * J) **Multi-Bond / Multi-Component Analysis:** Analyze the product for structural motifs that could be formed via reactions that form multiple bonds in one step, such as **cycloadditions** (ring-forming reactions between unsaturated systems) or **multi-component reactions** (where 3+ reactants combine in a single operation).
        * K) **Radical Mechanism Analysis:** K) Radical Mechanism Analysis: Analyze the molecule for transformations whose mechanism is best described as proceeding via radical (uncharged, open-shell) intermediates. This involves identifying bonds whose formation or cleavage is characteristic of single-electron processes (homolysis), as distinct from the two-electron processes of polar (ionic) reactions.
        * L) **Novel or Uncategorized Strategies:** If you identify a powerful, chemically sound transformation that does not clearly fit into categories A-K, classify it here.
    * **Output (Internal):** A list of formatted transformation strings representing all identified transformations. Each string must adhere to the format specified for the `"disconnection"` key in the Constraints & Formatting Rules. You MUST return all found disconnections. You are not allowed to leave any found and valid disconnection out.

2.  **Step 2: Assign Candidate Reactions**
    * **Input:** The list of transformation strings from Step 1.
    * **Process:** For each transformation, determine all appropriate forward reaction names. A single transformation may have multiple corresponding reactions.
    * **Output (Internal):** A list of objects, where each object contains a transformation and a list of its assigned `forwardReaction` names.
    * **Example:** `[{ "disconnection": "C:4 C:7", "reactions": ["Suzuki-Miyaura coupling", "Stille coupling"] }]`

3.  **Step 3: Expand and Evaluate Pairs**
    * **Input:** The list of objects from Step 2.
    * **Process:** Expand the input into a flat list by creating a **new, separate entry for each reaction** associated with a transformation. Then, for each of these new entries, apply the Retrosynthetic Analysis Framework to assign a `Retrosynthesis Importance` value and write a concise `rationale`.
    * **Output (Internal):** A flat list of fully populated objects, where each object represents one unique transformation-reaction pair.

4.  **Step 4: Final Formatting and Priority Assignment**
    * **Input:** The flat list of objects from Step 3.
    * **Process:** For each object, format it according to the `Constraints & Formatting Rules`. Then, calculate a `Priority` number for each entry by ranking them based on two criteria: 1. `"isInOntology"` (`true` before `false`), and 2. `"Retrosynthesis Importance"` (descending). Assign the resulting rank (`1, 2, 3...`) to the `"Priority"` key.
    * **Output:** The final, single JSON object. The list in this JSON does not need to be sorted.

**Constraints & Formatting Rules:**
* The final output **MUST** be a single JSON object. Do not include any text, explanations, or markdown formatting before or after the JSON.
* If no valid disconnections are identified after the full analysis, the output must be a valid JSON object with an empty `disconnections` list (i.e., `{"disconnections": []}`).
* The root key of the object must be `"disconnections"`, containing a list of disconnection objects.
* Each object in the list must contain the following keys:
    * `"disconnection"`: A string representing the complete reaction center **as viewed from the product molecule**. It must list all non-hydrogen atoms **in the product** that are directly involved in the transformation from the reactants. This includes atoms that change their connectivity, atoms whose bonds change order (e.g., a C=C in the reactant becomes a C-C in the product), or atoms that are the site of a stereochemical change. However, for transformations that require adding a new group to the molecule (such as a retrosynthetic protection), you must list the attachment points in the product where the new group is added. The atoms must be separated by spaces. 
        * **Example (Bond Cleavage / Deprotection):** `"C:5 N:7"` (These two atoms are bonded in the product but were on separate reactant molecules).
        * **Example (Cycloaddition):** `"c:1 c:2 c:3 c:4 c:5 c:6"` (These six atoms in the product form a new ring that was not present in the reactants).
        * **Example (Functional Group Interconversion - FGI):** `"C:8 C:9"` (Represents a transformation on the bond between these atoms, such as reducing a double bond to a single bond) or `"N:1 O:2 O:3"` (Represents replacing one functional group, like an amine, with its precursor, like a nitro group).
        * **Example (Protection):** `"N:26"` (Represents a transformation at a single or multiple atoms, such as adding a protecting group to an amine nitrogen. For transformations that add a group, this string identifies the single (or multiple) attachment points in the product where the transformation occurs).
        * **Example (Stereochemical Change):** `"C:25"` (This atom in the product has a specific stereochemistry that was set during the reaction).
    * `"Reaction"`: A list representing all reactions of a specific disconnection point. Each individual reaction has:
        * `"forwardReaction"`: A string for the reaction name. If the reaction is from the ontology, use its exact `id`. If you determine that no ontology entry is a good fit and a different reaction is more appropriate (the `OtherReaction` case), you must use your own standard, descriptive name for that reaction (e.g., `"Intramolecular Friedel-Crafts"`).
        * `"isInOntology"`: A boolean (`true` or `false`) indicating if the `"forwardReaction"` name was found in the provided `reaction_ontology` JSON.
        * `"forwardReactionClass"`: The broader reaction class of the `"forwardReaction"` selected from: 'Reduction', 'Acylation', 'Heteroatom Alkylation and Arylation', 'Functional Group Addition', 'Protection', 'C-C Coupling', 'Deprotection', 'Functional Group Interconversion', 'Aromatic Heterocycle Formation', 'Oxidation'. In case of no matching class pick 'Miscellaneous'.
        * `"Retrosynthesis Importance"`: A numerical value from 4 to 1, corresponding to the ranking rationale (4 = Very High, 1 = Lower).
        * `"Priority"`: A sequential integer (`1, 2, 3...`) representing the calculated priority of the disconnection.
        * `"rationale"`: A concise string explaining the strategic value. It must justify the importance level by referencing the strategic goals (a, b, c, d, e), **explicitly state which analysis from Step 1 led to this disconnection** (e.g., 'Convergent disconnection...'), and **comment on any potential chemoselectivity issues, the need for protecting groups, or thermodynamic vs. kinetic control considerations.**
    * **JSON Output Example:**
    {
    "disconnections": [
        {
        "disconnection": "C:1 C:2",
        "reactions": [
            {
            "forwardReaction": "Forward reaction name",
            "isInOntology": true,
            "forwardReactionClass": "Broader reaction class",
            "Retrosynthesis Importance": 4,
            "Priority": 1,
            "rationale": "string"
            },
            // more reactions for the same disconnection point
        ]
        },
        // more disconnection points
    ]
    }

**Retrosynthetic Analysis Framework**
* **Primary Strategic Goals:** Analyze the molecule according to the following framework. Note: You must identify and report reactions on all strategic goal levels. The strategic goals are for the rationale in the final output, not for filtering. Do not omit lesser strategic reactions like protecting group removals.
    * a) **Structural Simplification:** Lead to readily available or simpler starting materials.
    * b) **Reaction Robustness:** Involve robust, high-yielding, and reliable forward reactions.
    * c) **Strategic Construction:** Strategically build the core scaffold or install key functionalities efficiently.
    * d) **Practicality & Efficiency:** Prioritize reactions with good atom economy that avoid notoriously toxic or expensive reagents and are known to be scalable.
    * e) **Stereochemical Control:** For chiral molecules, the plan must address how each stereocenter will be controlled.
* **Ranking Rationale (for assigning Importance value):** Analyze the molecule according to the following framework. Note: You must identify and report reactions from all relevant importance levels. The importance score is for prioritization in the final output, not for filtering. Do not omit lower-importance findings like protecting group removals.
    * **Importance 4 (Very High):** Major ring-forming reactions, disconnections that reveal symmetry, or those that convergently connect major fragments. Includes powerful skeletal rearrangements that build the core.
    * **Importance 3 (High):** Reliable attachment of key functional groups or substituents to an existing core. Includes the strategic installation of a key stereocenter via an asymmetric reaction.
    * **Importance 2 (Medium):** Standard functional group interconversions (FGIs) or formation of less complex C-C or C-X bonds. Includes less critical rearrangements or stereochemical modifications.
    * **Importance 1 (Lower):** Disconnections of simple, easily accessible fragments or those related to reagent synthesis (e.g., protecting groups).
####

**Reaction Ontology:**

<reaction_ontology>

### Molecule for Analysis

**Product SMILES:**

<canonicalized_product>

####

Remember to return all possible reactions. You can identify more than one reaction for a specific position.
\end{minted}
\subsubsection{Transition Model}
\label{appendix:transition_prompt}
Note: This prompt is slightly altered for visualization purposes.
\begin{minted}{markdown}
**Persona:**
You are an expert chemist specializing in synthetic reaction modeling.

**Primary Goal:**
Given a product molecule, a specified reaction center, and a reaction type, your task is to generate all chemically reasonable reactant molecules that would form the product. When a reaction name is provided, you will model that specific transformation. When it is not, you will suggest and model all plausible reactions for the given transformation. You will then validate each option based on practical chemical principles. The entire output must be a single, valid JSON object.

**Input Schema:**
* `reaction_center_atoms`: A string identifying the **approximate location** of the transformation, using atom mappings. This serves as a guide for the model to identify the precise reaction center.
    * **Example (Bond Cleavage):** `"C:5 N:7"`
    * **Example (Ring Formation/Cycloaddition):** `"c:1 c:2 c:3 c:4 c:5 c:6"`
    * **Example (FGI):** `"C:8 C:9"`
    * **Example (Protection):** `"N:26"`
    * **Example (Stereochemical Change):** `"C:25"`
* `product_smiles`: The atom-mapped SMILES string of the product molecule.
* `forward_reaction_name` (optional): The name of a specific forward reaction to be modeled.
* `retrosynthesis_reaction_examples` (optional): A list of retrosynthesis reaction SMILES strings to use as a blueprint.

**Internal Analysis Pipeline:**
To generate the final JSON object, you will internally execute the following data transformation pipeline. This is a strict, one-way sequence from Step 1 to the final output. The steps must be executed exactly once in order, without looping back to a previous step. The output of each step serves as the direct input for the next.

1.  **Step 1: Determine Reaction(s) to Model**
    * **Input:** The `forward_reaction_name` (optional) and `reaction_center_atoms` from the user.
    * **Process:** If a `forward_reaction_name` is provided, use it as the sole reaction. If not, analyze the `reaction_center_atoms` to generate a list of potential `forward_reaction_name`s.
    * **Output (Internal):** A list of reaction names to be modeled.

2.  **Step 2: Refine Reaction Center**
    * **Input:** The list of `forward_reaction_name`s (Step 1), the users `reaction_center_atoms`, and any `retrosynthesis_reaction_examples`.
    * **Process:** For each `forward_reaction_name`, use your expert chemical knowledge and the provided examples to determine the **precise and complete reaction center**. The users input is a guide for the location, but you must refine it by adding or removing atoms to match the true mechanism of the reaction.
    * **Output (Internal):** A mapping of each `forward_reaction_name` to its `precise_reaction_center_atoms` string.

3.  **Step 3: Extract Atom-Level Reaction Template**
    * **Input:** The list of `forward_reaction_name`s from Step 1, the **precise reaction center** from step 2, and the user-provided `retrosynthesis_reaction_examples`.
    * **Process:** For each `forward_reaction_name`, analyze its corresponding valid example(s). Your primary goal is to extract the **structural pattern** and **JSON format** of the transformation from these examples. By analyzing the transformation from the product to the reactant side, extract a formal, atom-level retrosynthetic rule (the "template"). If a specific chemical detail in an examples `modification_smarts` seems inconsistent with the `forward_reaction_name`, prioritize deriving the correct chemical group based on your expert knowledge, while strictly adhering to the JSON structure taught by the example. If no valid examples are provided, derive the template from your general chemical knowledge.

    * **Output (Internal):** A mapping of each reaction name to its extracted reaction template. The template **must** be a single JSON object following this structure:
        ```json
        // Template Structure: A self-contained rule object
        {
          "precise_reaction_center_atoms": "<space_separated_list_of_atom_maps>",
          "modifications": [
            {
              "target_atom_map": "<map_number_of_atom_to_modify>",
              "modification_smarts": "<SMILES_or_SMARTS_of_the_complete_functional_ group_on_this_atom_in_the_reactant>"
            }
            // ... one object for each atom that is modified ...
          ]
        }
        ```

    * **Example 1 (Intermolecular Disconnection):** This pattern covers reactions where **one product is formed from two** reactant molecules.
        ```json
        {
          "precise_reaction_center_atoms": "C:1 C:7",
          "modifications": [
            { "target_atom_map": "1", "modification_smarts": "[c:1][X]" },
            { "target_atom_map": "7", "modification_smarts": "[c:7][Y]" }
          ]
        }
        ```

    * **Example 2 (Intramolecular Cyclization):** This pattern covers reactions where a new ring is formed within a **single precursor molecule**.
        ```json
        {
          "precise_reaction_center_atoms": "C:1 C:6",
          "modifications": [
            { "target_atom_map": "1", "modification_smarts": "[C:1]X" },
            { "target_atom_map": "6", "modification_smarts": "[C:6]Y" }
          ]
        }
        ```

    * **Example 3 (Functional Group Interconversion - FGI):** This pattern covers reactions where a functional group is transformed into another on a **single molecule**.
        ```json
        {
          "precise_reaction_center_atoms": "C:1 O:2",
          "modifications": [
            { "target_atom_map": "1", "modification_smarts": "[C:1]=[O:2]" }
          ]
        }
        ```

    * **Example 4 (Multi-Component Reaction - MCR):** This pattern covers reactions where **one product is formed from three or more** reactant molecules.
        ```json
        {
          "precise_reaction_center_atoms": "A:1 B:2 C:3",
          "modifications": [
            { "target_atom_map": "1", "modification_smarts": "[A]X" },
            { "target_atom_map": "2", "modification_smarts": "[B]Y" },
            { "target_atom_map": "3", "modification_smarts": "[C]Z" }
          ]
        }
        ```

4.  **Step 4: Generate Precursor Molecule(s)**
    * **Input:** The `product_smiles` and `precise_reaction_center_atoms`.
    * **Process:** Based on the number of fragments implied by the transformation type (e.g., two for an intermolecular disconnection, one for an FGI, three for a 3-component MCR), generate the corresponding core precursor molecule(s). This is done by cleaving the necessary bonds in the product or, for 1-to-1 transformations, identifying the single precursor scaffold.
    * **Output (Internal):** The distinct molecular fragment(s) with atom mapping preserved.

5.  **Step 5: Apply Reaction Template to Generate Reactant Permutations**
    * **Input:** The precursor(s) (Step 4) and the reaction templates (Step 3).
    * **Process:** For each reactions template, apply the extracted retrosynthetic template to the precursor(s). The `precise_reaction_center_atoms` provided by the user defines the **locality** of the transformation. You must use your chemical expertise to apply the template correctly to the atoms **in and around this specified location**, ensuring the final transformation is chemically consistent with the templates logic. This process must include generating **all possible permutations** of the reactive groups. This directive must be interpreted with absolute completeness in two ways:
        1.  **Fragment-Role Permutations:** For a disconnection into multiple fragments with distinct reactive groups, you must generate reactant sets for **all** possible assignments of those groups to the fragments.
        2.  **Intra-Group Class Permutations:** If a generated reactive group belongs to a general chemical class (e.g., an "organohalide," "leaving group," or "protecting group"), you are required to generate an exhaustive list of separate options for **all chemically distinct members of that class known to be compatible with the reaction.**
        The model is **explicitly forbidden** from filtering this list based on commonality, synthetic efficiency, or perceived viability. If a variant is chemically possible, it must be included in the output.
    * **Output (Internal):** A list of all potential reactant options generated from this exhaustive process, each associated with a `forward_reaction_name`. No chemically possible permutations may be omitted. Please dont provide reagents as reactants.

6.  **Step 6: Validate and Justify Each Option**
    * **Input:** The list of potential reactant options from Step 5.
    * **Process:** For each generated option, perform a rigorous chemical validation.
        * A) **Stability:** Are the proposed reactants chemically stable?
        * B) **Chemoselectivity:** Would the reaction be selective? Are there other functional groups that would interfere?
        * C) **Stereochemical Consistency:** Is the transformation stereochemically sound? Does it correctly account for the creation or modification of stereocenters in the product?
        * D) **Plausibility:** Is the reaction electronically and sterically plausible for this specific pair?
    * **Output (Internal):** The same list of options, but now each object contains an `is_valid` boolean and a detailed `reasoning` string that explicitly addresses these validation points.

### **Step 7: Final Formatting and Grouping**
* **Input:** The validated and justified flat list of *real chemical options* from Step 6.
* **Process:**
    1.  **Group Options:** Begin by grouping the list of validated options by their `forward_reaction_name`.
    2.  **Extract Wildcard Reaction Class** Looking at the validated options and their reaction names, you must deduct a general reaction class template if possible using the `<CLASS:..>` tag. It signals that a member of this chemical class (e.g. `<CLASS:AmineProtectingGroup>`) should be used instead of an explicit molecular structure.
    3.  **Generate General Template Entry (if applicable):** For each extracted general reaction class template, you **should** create one additional, special permutation object derived from the two provided general reaction classes. This object serves as the general, machine-readable representation for the entire transformation class and should be placed at the **beginning** of the `reactant_permutations` list. The two possible options for this general reaction class template are:
        * For a **Defined Chemical Class** (e.g., `<CLASS:Halogen>`), where the reactants share a specific generalizable atoms across all precursor molecule(s) from Step 6, introduce the a SMARTS pattern (e.g., `[A,B,C]`) as a replacement for these generalizable atoms. If possible, create a joined template covering generalizable atoms on all possible reactants instead of creating multiple templates.
        * For a **Wildcard Addition Class** (e.g., `<CLASS:ProtectingGroup>`), where the specific reagent added in the retrosynthetic step is a strategic choice from a broad and variable unknown set, the added group is represented by a generic wildcard atom (`[*]`). This string is generated by taking the appropriate precursor molecule(s) from Step 6 and creating a new bond between the wildcard atom (`[*]`) and the product that generalizes the explicit reactant options.
        * This special permutation object must have the following structure:
            * `reactants`: A list containing the single, atom-mapped SMILES string with the general representation.
            * `is_valid`: `true`.
            * `is_template`: `true`. Indicating that this result is a wildcard template.
            * `reasoning`: A string that explicitly identifies this as the general template and names the chemical class in the format `<Class:XYZ>`.
    4.  **Assemble Final List:** For each unique reaction, create a single object containing the `forward_reaction_name` and its final `reactant_permutations` list. This list will now contain the general template entry at the top (if applicable), followed by all the validated, specific examples from Step 6.
    5.  **Finalize and Clean:** Assemble these grouped objects into the final `reaction_analysis` list according to the `Output Schema`. Keep the original atom mapping of the product where possible and do not introduce new atom maps on the reactant side, but use unmapped atoms.
* **Output:** The final, single JSON object.

**Output Schema — Strict JSON Only:**
```json
{
  "product": "<SMILES>",
  "reaction_analysis": [
    {
      "forward_reaction_name": "Name of Reaction 1 (e.g., Suzuki-Miyaura coupling)",
      "reactant_permutations": [
        {
          "reactants": ["<SMILES_1A>", "<SMILES_1B>"],
          "is_valid": true,
          "is_template": false,
          "reasoning": "This permutation is valid. The reactants are stable and the reaction is chemoselective."
        },
        {
          "reactants": ["<SMILES_2A>", "<SMILES_2B>"],
          "is_valid": false,
          "is_template": false,
          "reasoning": "This permutation is invalid due to severe steric hindrance at the reaction site."
        }
      ]
    }
    // ... one object for each unique reaction suggested in Step 1 ...
  ]
}

** Input **
  
"reaction_center_atoms": <REACTION_POSITION>
"forward_reaction_name": <REACTION_NAME>
"product_smiles": <PRODUCT_SMILES>
"retrosynthesis_reaction_examples": <TRAIN_REACTION_EXAMPLES>
\end{minted}
\subsubsection{Transition Model Short}
\label{appendix:short_transition_prompt}
\begin{minted}{markdown}
Task:
Given a product molecule, a reaction center, and an optional reaction name, your task is to generate all chemically reasonable reactant molecules that would form the product. The entire output must be a single, valid JSON object following the specified schema.

Instructions:

    Identify the reaction(s) to model based on the inputs.

    For each reaction, determine the retrosynthetic disconnection.

    Generate all possible reactant permutations, including variations for chemical classes (e.g., all halogens for an organohalide). Do not filter out any chemically possible options.

    For each permutation, validate its chemical feasibility (stability, selectivity, etc.) and provide a brief justification.

    Group the results by forward_reaction_name in the final JSON output.

Input Schema:

    reaction_center_atoms: A string identifying the approximate location of the transformation, using atom mappings.

        Example (Bond Cleavage): "C:5 N:7"

        Example (Ring Formation/Cycloaddition): "c:1 c:2 c:3 c:4 c:5 c:6"

        Example (FGI): "C:8 C:9"

        Example (Protection): "N:26"

        Example (Stereochemical Change): "C:25"

    product_smiles: The atom-mapped SMILES string of the product molecule.

    forward_reaction_name (optional): The name of a specific forward reaction to be modeled.

    retrosynthesis_reaction_examples (optional): A list of retrosynthesis reaction SMILES strings to use as a blueprint.

Output Schema — Strict JSON Only:

{
  "product": "<SMILES>",
  "reaction_analysis": [
    {
      "forward_reaction_name": "Name of Reaction 1 (e.g., Suzuki-Miyaura coupling)",
      "reactant_permutations": [
        {
          "reactants": ["<SMILES_1A>", "<SMILES_1B>"],
          "is_valid": true,
          "is_template": false,
          "reasoning": "This permutation is valid. The reactants are stable and the reaction is chemoselective."
        },
        {
          "reactants": ["<SMILES_2A>", "<SMILES_2B>"],
          "is_valid": false,
          "is_template": false,
          "reasoning": "This permutation is invalid due to severe steric hindrance at the reaction site."
        }
      ]
    }
    // ... one object for each unique reaction suggested ...
  ]
}

  

** Input **

"reaction_center_atoms": <REACTION_POSITION>
"forward_reaction_name": <REACTION_NAME>
"product_smiles": <PRODUCT_SMILES>
"retrosynthesis_reaction_examples": <TRAIN_REACTION_EXAMPLES>
\end{minted}
\clearpage
\subsection{Deepseek-R1 Position Model Reasoning Trace}
\label{appendix:reasoning_trace}
Note: This reasoning trace using the PaRoutes reaction ontology has been slightly altered for visualization purposes. Deepseek-R1 was utilized for this demonstration because it provides full raw reasoning traces, whereas Gemini 2.5 Pro outputs only summaries. Therefore, this trace serves as an illustrative example of our "atom anchor" molecular reasoning framework rather than a benchmark of maximum model performance. Additionally, the text includes explanatory notes on the reasoning process and expert commentary by a chemist regarding chemical validity.

\begin{minted}{markdown}
#DeepSeek-R1 Reasoning Trace for LEI-515 with explanations & chemist comments

> Input Smiles: ```C[CH2:1][C:2]([C:3](=[O:4])[CH2:5][S:6](=[O:7])[c:8]1[cH:9][cH:10][c:11]([C:12](=[O:13])[N:14]2[CH2:15][CH2:16][N:17]([c:18]3[cH:19][c:20]([Cl:21])[cH:22][cH:23][cH:24]3)[C@@H:25]([CH3:26])[C@@H:27]2[CH3:28])[cH:29][c:30]1[Cl:31])([F:32])[F:33]```

We are given the product SMILES string with atom mapping. We need to perform retrosynthetic analysis step-by-step as described.

Step 1: Identify All Candidate Transformations (A to L)

Lets break down the molecule:

# Explanation 1: This section provides a detailed breakdown of the mapped SMILES string. The purpose is to accurately parse the molecules architecture, identifying all key functional groups, ring systems, and their precise connectivity before proceeding to the strategic analysis.

The molecule has the following features:
- A central carbon (C:2) attached to:
  - A ethyl group (C:1 is methyl, attached to CH2:1? Actually, the SMILES starts with C[CH2:1] meaning atom 1 is CH2, attached to a methyl (C) and then to C:2). So its a propyl group? Actually, the SMILES: C[CH2:1][C:2]... means that atom 1 is CH2 (carbon with two Hs) attached to a methyl (C) and to C:2. So the group is -CH2CH3? But note: the methyl is not mapped. So the group is ethyl? Actually, the SMILES: C[CH2:1] means a carbon (unmapped) attached to CH2:1. So the group is -CH2-CH3? But the atom mapping: [CH2:1] is atom 1, and the preceding C is unmapped. So the group is ethyl: -CH2-CH3, but the methyl is not mapped.

# Chemist: The model rightfully points out that C:0 is not mapped!

Let me write the molecule in a more readable way:

The SMILES: 
```C[CH2:1][C:2]([C:3](=[O:4])[CH2:5][S:6](=[O:7])[c:8]1[cH:9][cH:10][c:11]([C:12](=[O:13])[N:14]2[CH2:15][CH2:16][N:17]([c:18]3[cH:19][c:20]([Cl:21])[cH:22][cH:23][cH:24]3)[C@@H:25]([CH3:26])[C@@H:27]2[CH3:28])[cH:29][c:30]1[Cl:31])([F:32])[F:33]```

We can see:
- The molecule has a central carbon (C:2) that is tertiary, attached to:
  1. A group: -CH2-CH3 (atoms: unmapped C and CH2:1)
  2. Two fluorines (F:32 and F:33)
  3. A chain: -C(=O)-CH2-S(=O)-aryl

# Chemist: Due to how the input SMILES was structured, C:2 is given the role as central carbon which is debatable, but understandable from a SMILES linguistic point of view

The chain: 
- C:3 is carbonyl carbon, attached to O:4 (double bond) and then CH2:5
- CH2:5 attached to S:6, which has double bond to O:7 and attached to aryl ring (c:8)

The aryl ring (c:8) is substituted at positions:
- c:8 attached to c:9 (CH), c:10 (CH), c:11 (C), c:29 (CH), c:30 (C) and then back to c:8? Actually, its a benzene ring: c:8, c:9, c:10, c:11, c:29, c:30, and then c:30 attached to Cl:31 and back to c:8? Actually, the SMILES: [c:8]1[cH:9][cH:10][c:11](...)[cH:29][c:30]1[Cl:31] -> so its a benzene ring with substituents at c:11 and c:30? Actually, c:11 has a substituent: [C:12](=[O:13])[N:14]2[CH2:15][CH2:16][N:17](...)[C@@H:25]([CH3:26])[C@@H:27]2[CH3:28]

So the substituent at c:11 is: -C(=O)-N< (where the nitrogen is part of a piperazine-like ring? Actually, N:14 is attached to C:12 (carbonyl) and then to CH2:15 and CH2:16, then N:17. Then N:17 is attached to an aryl group (c:18) and to C:25 (which is chiral) and then to C:27 (which is also chiral) and then back to N:14? Wait, the ring is: N:14 - CH2:15 - CH2:16 - N:17 - C@@H:25(CH3:26) - C@@H:27(CH3:28) - N:14? That would be a 6-membered ring? But note: the atom after N:17 is [C@@H:25] and then [C@@H:27]2, so the ring is: N:14 - C:15 - C:16 - N:17 - C:25 - C:27 - and then back to N:14? But the atom mapping at the end: [C@@H:27]2[CH3:28] and then the ring is closed by the 2 at the end? Actually, the ring closure is at N:14 and C:27? Because the ring is defined by the number 2: [N:14]2 ... [C@@H:27]2. So the ring is: N:14 - CH2:15 - CH2:16 - N:17 - C:25 - C:27 - and then back to N:14? That would be a 6-membered ring? But note: C:25 and C:27 are both chiral carbons.

# Chemist: The model correctly identifies the ring shapes in the molecule, meanwhile correctly flagging C25 and C26 as chiral

Additionally, N:17 is attached to an aryl group: [c:18]3[cH:19][c:20]([Cl:21])[cH:22][cH:23][cH:24]3 -> a chlorophenyl group.

So the molecule has:
- A central fluorinated carbon (C:2) with two F atoms and two carbon chains: one ethyl and one chain leading to a sulfonyl group attached to a chlorophenyl ring (with two chlorines? Actually, the ring has Cl:31 at c:30 and Cl:21 at c:20? So its a dichlorophenyl? Actually, the ring at c:8 has two chlorines: Cl:31 at c:30 and Cl:21 at c:20? But note: the ring c:8 is attached to S:6, and then the substituents: at c:11 we have the carbonyl-piperazine part, and at c:30 we have Cl:31. Also, the ring c:18 (attached to N:17) has Cl:21 at c:20.

# Chemist: The model has a little bit of confusion if both chlorides are at the same ring or on different positions. The model provides no conclusion.

Now, lets analyze step by step:

# Explanation 2: This section systematically applies the 12 core principles of retrosynthetic analysis (A-L) to the parsed molecular structure. The goal is to identify all strategically viable disconnections, including inter-fragment (convergent) and intra-fragment (linear) bonds, as well as key functional group interconversions (FGIs).

A) Symmetry Analysis: The molecule does not appear to have any symmetry.

B) Fragment Partitioning: We can partition the molecule into:
   - The ethyl group (C:1 and the unmapped methyl)
   - The central fluorinated carbon (C:2) with two fluorines
   - The chain: -C(=O)-CH2-SO2-aryl (where the aryl is substituted with a carbonyl-piperazine group and chlorines)

But note: the chain is complex and includes a piperazine ring and another aryl group. So we might break at the sulfonyl group because sulfonyl groups are often formed by oxidation of thiols or by coupling reactions.

# Chemist: First (wrongful) mention of sulfonyl. Should be sulfoxide (sulfoxide: S=O, Sulfonyl: S(=O)=O)

C) Inter-Fragment Analysis: The bonds connecting major fragments:
   - The bond between C:2 and C:3 (the carbonyl carbon) -> disconnection here would give an alkyl halide (from C:2) and a nucleophile (from C:3). But note C:3 is carbonyl carbon, so it might be an acyl halide? But then we have an ester? Actually, the group attached to C:2 is C:3=O, so its a ketone? But wait: C:2 is attached to C:3 with a single bond? Then C:3 is attached to O:4 with a double bond? So its a ketone: -C(O)-CH2-SO2-aryl.

   # Chemist: Obviously, it is a ketone, intriguing that it takes so long to confirm this.

So the central carbon (C:2) is a tertiary carbon with two fluorines and two alkyl groups: ethyl and the ketone chain.

D) Strategic Bond Analysis: Within fragments, look for bonds adjacent to functional groups.
   - The bond between C:3 and C:5: C:3 is carbonyl carbon, so the bond C3-C5 is alpha to the carbonyl. Disconnection here would be an enolate alkylation? But note: C:5 is CH2, so its a methylene group. Disconnection at C3-C5 would give a nucleophile at C:5 (as an anion) and an electrophile at C:3 (as an acyl halide?).

   # Chemist: Really cool to see the model to reason like this. Its all correct!

   - The sulfonyl group: S:6 is attached to C:5 and to the aryl ring (c:8). The bond S:6 - c:8 is a sulfonyl-aryl bond. This bond can be disconnected via a sulfonylation reaction (aryl sulfonyl chloride + amine? but here its attached to a methylene, not an amine). Actually, the group attached to S:6 is CH2 (C:5) and the aryl ring. So its an alkyl-arylsulfone. This bond can be formed by nucleophilic substitution: the aryl sulfonyl chloride and a carbanion? Or by coupling? Alternatively, we can disconnect the sulfonyl group as a whole: the bond S=O might be formed by oxidation of a sulfide.

   # Chemist: This is true of both sulfoxide and sulfone, so the analysis doesnt actually suffer much from the misintrepretation

E) Intra-Fragment Analysis: Within major fragments, identify bonds for intramolecular reactions.
   - The piperazine ring (N:14, C:15, C:16, N:17, C:25, C:27) is already formed. But note: the ring is closed. There might not be an intramolecular reaction in the product. However, we can consider that the ring was formed by an intramolecular amination? But its already formed.

F) Stereochemical Analysis: The molecule has two chiral centers: C:25 and C:27. Both are part of the piperazine ring. We need to set the stereochemistry. Possible transformations: asymmetric synthesis, chiral pool, or resolution.

G) Rearrangement Analysis: No obvious rearrangement motifs.

H) FGI Analysis: Functional groups present:
   - Ketone: C:3=O
   - Sulfonyl: S:6(=O)-aryl
   - Amide: C:12=O attached to N:14 (so its an amide)
   - Tertiary amine: N:17 (attached to aryl and two carbons)
   - Aryl chlorides (c:20-Cl:21 and c:30-Cl:31)
   - Fluorines on C:2

   # Chemist: Complete, correct.

   i. Oxidation/Reduction:
      - The ketone (C:3) could be derived from an alcohol (by oxidation) or from an alkyne (by hydration).
      - The sulfonyl group could be derived from a sulfide (by oxidation).
      - The amide (C:12) could be derived from a carboxylic acid or ester.

   ii. Non-Redox FGIs:
      - The amide (C:12) could be formed from a carboxylic acid and an amine.
      - The sulfonyl group: the bond S:6-C:5 could be disconnected if we consider the sulfonyl as a whole, but thats not FGI. For FGI, we might consider the sulfonyl group as a whole, but note: the sulfonyl group is attached to two groups: the aryl and the alkyl. So we can disconnect at the S-C bonds? But that would be a disconnection, not FGI. FGI would be, for example, converting the sulfonyl group to a thiol? Not common. Alternatively, we might consider the sulfonyl group as a functional group that can be interconverted? Typically, sulfonyl groups are formed by oxidation.

I) Protecting Group Analysis: The molecule does not have obvious protecting groups. The fluorines and chlorines are likely part of the final structure.

J) Multi-Bond / Multi-Component Analysis: 
   - The piperazine ring might be formed by a cyclization? But its already formed. Alternatively, the entire molecule might be assembled by a multi-component reaction? Not obvious.

# Chemist: There are hints that the model is trying to reason forward in the synthesis, not always grasping that it is trying to break down the product in a retrosynthetic analysis

K) Radical Mechanism Analysis: No obvious radical disconnections.

L) Novel Strategies: None apparent.

Now, lets list candidate disconnections:

# Explanation 3: This section consolidates the findings from the 12-principle analysis (A-L). It presents a final, itemized list of the most promising candidate transformations (both bond disconnections and FGIs). Each transformation string listed here will be carried forward for detailed reaction assignment and evaluation.

1. Disconnection at the bond between C:2 and C:3 (the ketone carbon). This would disconnect the molecule into two parts: 
   - Part A: the fluorinated alkyl part: C[CH2:1][C:2](F)(F) -> but note: C:2 is attached to two fluorines and ethyl. So it would be 1,1-difluoro-1-ethyl? Actually, C:2 is attached to CH2:1 (which is ethyl group: -CH2CH3) and two fluorines. So the fragment would be [CH3-CH2-C(F2)-]? But note: C:2 is a carbon with two fluorines and two groups: ethyl and the ketone chain. Disconnecting at C2-C3 would leave C:2 with a halogen? So we would have an alkyl halide: Br-C(F2)-CH2CH3? And the other fragment would be the ketone chain without the attachment to C:2? But wait: after disconnection, the fragment containing C:3 would have a carbonyl and then CH2-SO2-aryl. So it would be an enolate equivalent? Actually, we can do a disconnection of the ketone at the bond between C:2 and C:3: this is a disconnection of a ketone that is substituted by an alkyl group (ethyl) and a functionalized alkyl chain. The standard disconnection is to an alkyl halide and an enolate. So the transformation would be alkylation of an enolate.

   The disconnection: breaking C:2-C:3. The atoms involved: C:2 and C:3.

2. Disconnection at the sulfonyl group: 
   a) Disconnect the S-C:5 bond. This would give two fragments: 
        - The alkyl part: ...C:3(=O)-CH2:5-X (where X is leaving group? but we are disconnecting, so the alkyl part would be an electrophile: halide?) 
        - The aryl sulfonyl part: [S:6](=O)(=O)[c:8]... (so an aryl sulfonyl anion? or aryl sulfonyl chloride? Actually, in retrosynthesis, we disconnect to aryl sulfonyl chloride and the alkyl halide. The forward reaction is a nucleophilic substitution: ArSO2- + R-X -> ArSO2R. But note: the alkyl group is activated? Its a methylene group adjacent to a ketone? So it might be acidic? We can form the carbanion and then do a nucleophilic substitution? Alternatively, we can use the alkyl halide and aryl sulfinate? But the standard way is to form the sulfone from a sulfide and then oxidize? Or directly by coupling? 

   The disconnection: breaking S:6-C:5. Atoms: S:6 and C:5.

   b) Alternatively, we can disconnect the sulfonyl group by considering it was formed by oxidation of a sulfide. So we can do a functional group interconversion: sulfonyl to sulfide. Then disconnect the sulfide: the bond S:6-C:5 and S:6-c:8? That would be two disconnections. But note: FGI is not a disconnection per se. We are asked for disconnection points. So we might consider the oxidation as a transformation? But the disconnection would be the same as above? Or we can do a disconnection at the sulfide level? 

   Actually, if we do FGI: sulfonyl to sulfide, then we have the sulfide: C:5-S:6-c:8. Then we can disconnect the sulfide bond: for example, S:6-C:5: then we get aryl thiol and alkyl halide? So the disconnection would be at S:6-C:5 for the sulfide, and then we have to do an oxidation to sulfonyl. But the disconnection point is still S:6-C:5.

3. Disconnection at the amide bond: C:12-N:14. This would give a carboxylic acid (from C:12) and the amine (from N:14). The amine is part of the piperazine ring. The transformation is amide formation.

   Disconnection: breaking C:12-N:14. Atoms: C:12 and N:14.

4. Disconnection at the bond between the piperazine ring and the aryl group attached to N:17. The bond is N:17-c:18. This is an N-aryl bond. Disconnection would give an amine (the piperazine) and an aryl halide? So the transformation is N-arylation.

   Disconnection: breaking N:17-c:18. Atoms: N:17 and c:18 (specifically, the carbon of the aryl ring that is attached? We can use the atom mapping: c:18 is the carbon attached to N:17? So atoms: N:17 and c:18.

5. Disconnection at the ketone in the chain: C:3=O. We can disconnect the ketone by converting it to an alkyne? Or by other methods? But note: the ketone is between two alkyl groups: one is the fluorinated alkyl (from C:2) and the other is the methylene (C:5). So we can disconnect the ketone to the alkyl halide (from the fluorinated side) and the carboxylic acid (from the methylene side) via a Friedel-Crafts acylation? But note: the methylene side is not aromatic. Alternatively, we can do a disconnection at the ketone by converting it to an ester and then do a Grignard? But that would be two steps. We already considered disconnection at C2-C3.

   Alternatively, we can disconnect the ketone by a Grignard reaction: the bond between C:3 and the group attached to it? But note: C:3 is attached to C:2 (from one side) and to C:5 (from the other). So we can disconnect at C:3-C:5? Then we would have an acyl group (attached to C:2) and a methyl group? But that doesnt match. The standard ketone disconnection is at the bond between the carbonyl carbon and one of the adjacent carbons? We already disconnected at C2-C3. The other bond is C3-C5. Disconnecting at C3-C5 would give two fragments: 
      - Fragment A: ...C:2-C:3=O (so an aldehyde? Actually, if we disconnect C3-C5, then C:3 becomes an aldehyde: O=CH-C:2? But note: C:2 is attached to two fluorines and ethyl, so it would be an aldehyde: O=CH-CF2-CH2CH3? 
      - Fragment B: CH2:5-SO2-aryl.

   Then the forward reaction would be a nucleophilic addition of the fragment B carbanion to the aldehyde? But that would give an alcohol, not a ketone. So we need to oxidize? Alternatively, we can do a Wittig? But that would give an alkene, then we need to hydrogenate? 

   So the direct disconnection at C3-C5 is not straightforward. Therefore, we stick to the disconnection at C2-C3.

6. Disconnection at the ethyl group: the bond between C:1 and C:2? But note: the ethyl group is attached to C:2. Disconnecting at C:1-C:2 would give two fragments: 
   - Fragment A: CH3-CH2:1 (so ethane? but we need a synthon: ethyl anion or ethyl cation) 
   - Fragment B: the rest of the molecule attached to C:2 (which is CF2- with the ketone chain). 

   But note: C:2 is tertiary and has two fluorines. So we can have an alkyl halide: F2C(ketone chain)-Br? and then react with ethyl magnesium bromide? That would be a Grignard reaction.

   # Chemist: This is chemically not correct. Every 1st year student should know the grignard will react with the ketone, not the bromide.

   Disconnection: breaking C:1-C:2. Atoms: C:1 and C:2.

7. Stereochemical disconnections: We have two chiral centers at C:25 and C:27. We can consider transformations that set the stereochemistry. For example, asymmetric hydrogenation of a double bond? Or using a chiral auxiliary? But note: the ring is already formed. So we might consider that the ring was formed by a stereoselective reaction? Or we can use a chiral pool starting material? 

   However, we dont have a direct disconnection at the chiral centers. We can consider the ring formation: the bond between N:14 and C:27? or between N:17 and C:25? But that would break the ring. Alternatively, we can form the ring by an intramolecular reaction? For example, the bond between N:17 and C:25? But note: the ring is already formed. 

   Actually, the ring closure could be done by an intramolecular nucleophilic substitution? For example, if we have a diamine with halide at C:25? Then the amine N:17 attacks C:25? But note: the ring is six-membered. Alternatively, we can form the ring by reductive amination? 

   But note: the ring is already formed in the product. So we can disconnect the ring at one of the bonds? For example, break the bond between C:15 and C:16? Then we would have a diamine and a dihalide? But that is not convergent. Alternatively, break the bond between N:14 and C:15? Then we have an amine and a halide? 

   However, we are not required to break the ring if it is readily available. But we can consider disconnections at the ring bonds.

   Lets break the ring at the bond between N:14 and C:27? Actually, the ring is: N:14 - C:15 - C:16 - N:17 - C:25 - C:27 - N:14. So breaking the bond between N:14 and C:27 would give an open chain: 
        HO-C:27 - C:25 - N:17 - C:16 - C:15 - N:14? 
   But note: the bond we break is N:14-C:27? Then the fragment would be: 
        Fragment A: ...C:12(=O)-N:14 (so the amide) and then the open chain: N:14-H and C:27-OH? 
        Fragment B: the rest of the ring? 

   This doesnt seem straightforward. Alternatively, we can break the bond between C:25 and C:27? Then we get two fragments: 
        Fragment A: ...N:17 - C:25 (which would be an amine? because N:17 is attached to C:25? Actually, if we break C:25-C:27, then C:25 becomes a carbonyl? and C:27 becomes a carbonyl? Not necessarily. 

   Alternatively, we can break the bond between N:17 and C:25? Then we get:
        Fragment A: ...N:17 (which is now NH) and Fragment B: C:25 - C:27 (which would be a carbonyl compound?).

   Then the forward reaction would be reductive amination? Because N:17 is secondary amine? Actually, N:17 is tertiary in the product? Because it is attached to the aryl group? So its tertiary. Therefore, we cannot break that bond without breaking the aryl bond? 

   Given the complexity, we might leave the ring intact and use a commercially available chiral piperazine? 

   # Chemist: It took a little while, but this is a wise choice

   So we wont disconnect the ring for now.

8. Functional group interconversion: 
   - The ketone (C:3) could be reduced to an alcohol? But that would be a step backward. 
   - The sulfonyl group could be reduced to sulfide? But again, backward.
   - The amide (C:12) could be hydrolyzed to carboxylic acid? Then we would have to form the amide again? 

   Instead, we can consider FGI for the synthesis of the sulfonyl group: we can have a sulfide and then oxidize. So we add a transformation: oxidation of sulfide to sulfonyl. The disconnection would be the same as the sulfide disconnection? 

   But note: for the sulfonyl disconnection, we can do:
        Step 1: FGI: sulfonyl to sulfide -> then the sulfide is disconnected to aryl thiol and alkyl halide? 
        Then the alkyl halide fragment is ...C:3(=O)-CH2:5-Br? 

   So the disconnection for the sulfide would be at S:6-C:5? Then we have two fragments: 
        Fragment A: the aryl thiol (with the aryl ring c:8 and substituents)
        Fragment B: the alkyl halide: Br-CH2-C(=O)-C:2(F)(F)CH2CH3? 

   Then the forward reaction is nucleophilic substitution: thiolate + alkyl halide -> sulfide. Then oxidation to sulfonyl.

   So we can have two transformations: 
        a) Nucleophilic substitution: disconnection at S:6-C:5 (for the sulfide)
        b) Oxidation: disconnection of the S=O bonds? But thats not a disconnection of a bond? We dont disconnect bonds for oxidation. We just note that we need to oxidize.

   However, the problem asks for disconnection points. For the oxidation, we dont break a bond? We change the functional group. So the disconnection for the oxidation step would be the same atoms? Or we dont list it as a disconnection? 

   # Chemist: From here on, it spends some time thinking about how it should interpret the assignment and what a correct answer should look like

   The instructions say: for FGI, we list the atoms involved in the transformation. For example, for oxidation of sulfide to sulfonyl, we might list the sulfur atom? But the example says for FGI: `"N:1 O:2 O:3"` for replacing an amine with a nitro group. So for oxidation of sulfide, we would list the sulfur atom? But note: the oxidation doesnt break any bonds? It adds oxygen. 

   Alternatively, we can consider the oxidation as a transformation that doesnt break bonds? So we dont disconnect? But the instructions for FGI say: "identify all possible functional groups that are candidates for standard Functional Group Interconversions". And then in the disconnection string, for FGI we list the atoms that change. 

   Example: for reducing a double bond to single bond: `"C:8 C:9"`. So for oxidation of sulfide to sulfonyl, we would list the sulfur atom? Because the sulfur changes from S to S(=O)2? So the disconnection string would be `"S:6"`? 

   Similarly, for reducing the ketone to alcohol? Then we would list C:3? But note: the ketone carbon is C:3. So `"C:3"`? 

   But note: the instructions also say: "for transformations that require adding a new group to the molecule (such as a retrosynthetic protection), you must list the attachment points". Oxidation adds oxygen atoms? So we list the sulfur? 

   However, in the example for protection: `"N:26"` for adding a protecting group to nitrogen. So for oxidation, we list the atom that is oxidized? 

   Therefore, we can have:
        FGI: oxidation of sulfide to sulfonyl: disconnection string `"S:6"`
        FGI: reduction of ketone to alcohol: `"C:3"`

   But note: these are not disconnections? They are functional group interconversions that dont break bonds? So they are not disconnections? 

   However, the instructions in Step 1 (H) say: "For each functional group in the molecule, systematically identify all possible functional groups that are candidates for standard Functional Group Interconversions". And then we output a transformation string. 

   So we must include these FGI transformations? 

   But note: the primary goal is disconnection points? And the transformation string for FGI might be a single atom? 

   So we will include:

   Transformation for FGI (oxidation): `"S:6"`
   Transformation for FGI (reduction): `"C:3"` (if we reduce the ketone, then we have to form the ketone later? So its an option)

   However, reducing the ketone might not be strategic because then we lose the activation for the methylene group? 

   We might also consider FGI for the amide: we can hydrolyze the amide to carboxylic acid? Then disconnection at the amide bond would be the carboxylic acid and amine? But then we have to form the amide again? So the disconnection at the amide bond is already considered (point 3). 

   Alternatively, we can have FGI: convert the amide to nitrile? Then disconnect? But that is less common.

   So for FGI, we have:
        - Oxidation of sulfide: `"S:6"`
        - Reduction of ketone: `"C:3"`
        - Oxidation of alcohol to ketone? But we dont have an alcohol? 
        - Reduction of amide? Then we get an amine? But the amide is tertiary? Actually, the amide is secondary: C:12(=O)-N:14 (and N:14 is attached to two carbons: the ring). So its a tertiary amide? Actually, N:14 is attached to C:12 (carbonyl) and two carbons (C:15 and C:27). So its a tertiary amide? But tertiary amides are harder to reduce? 

   So well stick to the ones above.

9. Protection: The molecule does not have obvious protecting groups? But we might need to protect the amine during synthesis? However, in the product there is no protection. So we dont have a deprotection step? 

   But note: the amide nitrogen (N:14) is part of the ring? So its not free. The other amine (N:17) is tertiary? So no protection needed? 

   However, during the synthesis of the piperazine ring, we might need to protect amines? But that is not in the product. 

   So we skip.

Now, we have the following candidate disconnections (with transformation strings):

1. Disconnection at C:2-C:3: `"C:2 C:3"`
2. Disconnection at S:6-C:5: `"S:6 C:5"`
3. Disconnection at C:12-N:14: `"C:12 N:14"`
4. Disconnection at N:17-c:18: `"N:17 c:18"`
5. Disconnection at C:1-C:2: `"C:1 C:2"`
6. FGI: oxidation of sulfide: `"S:6"`
7. FGI: reduction of ketone: `"C:3"`

But note: the FGI transformations are not disconnections? But we are asked to list all transformations? And the output of Step 1 is a list of transformation strings. 

Also, note: the disconnection for the sulfide formation (point 2) is `"S:6 C:5"`. Then the oxidation FGI is `"S:6"`. 

Additionally, we might have missed:

8. Disconnection at the bond between C:5 and C:3? But we discussed that it is not straightforward? So we skip.

9. Disconnection at the bond between the aryl ring (c:8) and the sulfonyl group? That is, break the bond S:6-c:8? Then we would have an alkyl sulfonyl chloride and an aryl compound? But the aryl compound would need to have a nucleophile? Not common. Typically, sulfonyl chlorides react with nucleophiles? So breaking S:6-c:8 would be disconnecting the aryl group? Then the forward reaction would be sulfonylation of the aryl ring? But sulfonylation of aryl rings is done by Friedel-Crafts? But that would require the aryl ring to be activated? And we have chlorines? Deactivating. So not favorable.

   Therefore, we dont consider.

Now, we proceed to Step 2: Assign Candidate Reactions for each transformation.

# Explanation 4: This section translates the abstract disconnection strings into practical, nameable synthetic operations. For each transformation, one or more specific forward reactions are assigned (e.g., Acylation... for the C:12-N:14 amide bond). Each reaction is then cross-referenced with the provided ontology to determine its status as a standard, cataloged transformation (isInOntology).

For each transformation string, assign forward reactions:

1. `"C:2 C:3"`: This disconnection corresponds to breaking the bond between C:2 and C:3. In the forward direction, this bond is formed by:
   - Alkylation of an enolate: the enolate from the fragment containing C:3 (which is the ketone chain: -CH2-SO2-aryl) and an alkyl halide from the fluorinated fragment (Br-C(F2)-CH2CH3). 
   - Possible reactions: 
        - Enolate alkylation: this is a standard reaction. But note: the fluorinated alkyl halide might be electrophilic? 
        - The reaction might be called "Alkylation of enolates" or "C-alkylation". 
   - Check ontology: 
        - "C-methylation" -> but we are alkylating with ethyl? 
        - "Alkylation of amines" -> no, its carbon.
        - "Friedel-Crafts alkylation" -> no, not aromatic.
        - "Reaction of alkyl halides with organometallic compounds" -> that would be Grignard? But Grignard on what? 
        - Actually, the standard name is "Alkylation of carbonyl compounds" or "alpha-alkylation". 
   - In the ontology, we have "C-methylation", but that is specific to methyl. We have "Alkylation" but not listed? 
        - "N-alkylation of primary amines with alkyl halides" -> no.
        - "S-alkylation of thiols" -> no.
        - "Friedel-Crafts alkylation" -> no.
        - "Wittig" -> no.
        - "Michael addition" -> no, unless its a Michael acceptor? 
        - "Alkylation" is not explicitly listed? 
        - We see "C-methylation", which is for methyl. We are alkylating with ethyl? So we might use "OtherReaction": "Enolate alkylation".

   Alternatively, we can use the Grignard reaction? But note: if we disconnect to an acyl compound and a Grignard? That would be disconnecting at C:3-C:5? We are disconnecting at C:2-C:3.

   So for `"C:2 C:3"`, we assign: 
        forwardReaction: "OtherReaction: Enolate alkylation"

   But note: the ontology has "Alkylation of Carbon Nucleophiles by Alkyl Halides"? Not in the list? 

   Since we dont find a direct match, we use "OtherReaction: Enolate alkylation".

2. `"S:6 C:5"`: Disconnection of the bond between S:6 and C:5. In the forward direction, this bond can be formed by:
   - Nucleophilic substitution: the thiolate anion (from the aryl thiol) and the alkyl halide (from the fragment: X-CH2-C(=O)-...). 
   - Reactions: 
        - "S-alkylation of thiols" -> yes, in the ontology: "S-alkylation of thiols", "S-alkylation of thiols (ethyl)", "S-alkylation of thiols with alcohols" (but we are using alkyl halide?).
        - Specifically, "S-alkylation of thiols" seems appropriate.

   So assign: forwardReaction: "S-alkylation of thiols"

   Also, note: the alkyl halide might be activated by the adjacent ketone? So it should be reactive.

3. `"C:12 N:14"`: Disconnection of the amide bond. Forward reactions:
   - Amide formation from carboxylic acid and amine: 
        - "Carboxylic acid with primary amine to amide" -> but note: the amine (N:14) is secondary? Because it is part of a piperazine ring? Actually, in the ring, N:14 is attached to two carbons? So its a secondary amine? But the amide is formed from a secondary amine? Then it would be a tertiary amide? Yes, because the nitrogen of the amide is tertiary? 

        - The ontology has: "Carboxylic acid with primary amine to amide" -> primary amine gives secondary amide? 
        - "Acylation of Nitrogen Nucleophiles by Acyl/Thioacyl/Carbamoyl Halides and Analogs_N" -> this might be for acyl halides? 
        - "Acylation of Nitrogen Nucleophiles by Carboxylic Acids" -> but that might require activation? 

        - We have: "Carboxylic acid to amide conversion" -> general? 
        - "Aminolysis of esters" -> using an ester and amine? 
        - "Schotten-Baumann to ester" -> no, thats for ester? 

        - The standard ways: 
            a) Acyl chloride + amine
            b) Carboxylic acid + amine with coupling agent
            c) Ester + amine

        - In the ontology: 
            - "Acyl chloride with ammonia to amide" -> ammonia, but we have amine? 
            - "Carboxylic acid with primary amine to amide" -> but we have secondary amine? 
            - "Acylation of Nitrogen Nucleophiles by Acyl/Thioacyl/Carbamoyl Halides and Analogs_N" -> this might be for secondary amines? 
            - "Acylation of Nitrogen Nucleophiles by Acyl/Thioacyl/Carbamoyl Halides and Analogs_OS" -> ? 

        - Also, "Amide from carboxylic acid and amine using DCC" -> not in ontology? 

        - We see: "Acylation of Nitrogen Nucleophiles by Acyl/Thioacyl/Carbamoyl Halides and Analogs_N" -> this is for acyl halides? 

      # Chemist: These parts clearly show the importance the ontology has in the succesfull outcome of assignments. The model really tries to understand the partially overlapping reaction names from rxn-insight

        - Since the amide is tertiary, we can use: 
            - "Acylation of Nitrogen Nucleophiles by Acyl/Thioacyl/Carbamoyl Halides and Analogs_N" -> but the name is long? 

        - Alternatively, "OtherReaction: Amide coupling"

   However, the ontology has: "Acylation of Nitrogen Nucleophiles by Acyl/Thioacyl/Carbamoyl Halides and Analogs_N" -> well use this for acyl chloride.

   Also, we can use ester aminolysis: "Aminolysis of esters"

   So assign two reactions:
        a) "Acylation of Nitrogen Nucleophiles by Acyl/Thioacyl/Carbamoyl Halides and Analogs_N"
        b) "Aminolysis of esters"

4. `"N:17 c:18"`: Disconnection of the N-aryl bond. Forward reactions:
   - N-arylation of the amine (piperazine) with an aryl halide? 
   - Reactions: 
        - "Buchwald-Hartwig/Ullmann-Goldberg/N-arylation secondary amine" -> because the amine (N:17) is tertiary? Wait, in the product, N:17 is attached to the aryl group and to two carbons? So its a tertiary amine? But the arylation of a tertiary amine? 

        - Actually, the amine in the forward reaction would be secondary? Because we disconnect the aryl group? So the amine fragment would be: the piperazine with N:17 being secondary (attached to two carbons: C:16 and C:25, and H). 

        - So its a secondary amine? 

        - Then we have: 
            - "Buchwald-Hartwig/Ullmann-Goldberg/N-arylation secondary amine"

        - Also, "Ullmann-Goldberg Substitution amine"

   So assign: "Buchwald-Hartwig/Ullmann-Goldberg/N-arylation secondary amine"

   Also, "Ullmann-Goldberg Substitution amine" -> but the ontology has both? 

   Well assign both.

5. `"C:1 C:2"`: Disconnection of the bond between C:1 and C:2. Forward reaction:
   - Grignard reaction: the fragment attached to C:2 (without the ethyl) would be a ketone? But note: C:2 is attached to two fluorines and the ketone chain? So if we disconnect C:1-C:2, then the fragment containing C:2 would be: F2C- (with the ketone chain) -> so it would be a carbonyl? No, because C:2 is alkyl carbon? 

   - Actually, we disconnect to:
        - Fragment A: ethyl magnesium bromide (CH3CH2MgBr)
        - Fragment B: a carbonyl compound: O= C(F2)- (with the ketone chain) -> but that carbonyl would be an aldehyde: O=CH-? But note: C:2 is carbon, so if we remove the ethyl group, we have a carbonyl? That would be a ketone: the ketone chain is attached to C:2? So we have a ketone: F2C-C(=O)-R? But that is not the case. 

   - Alternatively, we can have an alkyl halide: Br-C(F2)- (with the ketone chain) and then react with ethyl magnesium bromide? That would be a Grignard on an alkyl halide? But Grignard reagents are nucleophiles, not electrophiles? 

   - So the correct disconnection: the bond C:1-C:2 is formed by the reaction of an ethyl nucleophile (ethyl anion equivalent) with an electrophile at C:2. The electrophile at C:2 would be a carbonyl? But C:2 is not a carbonyl. 

   - Alternatively, we can have a radical coupling? Or we can have an alkyl halide at C:2 and then do a coupling with an ethyl metal? 

   - The standard way: the group attached to C:2 is a fluorinated alkyl group? We can have a fluorinated alkyl halide: Br-CF2-CH2-C(=O)-... and then do a Grignard with ethyl magnesium bromide? But that would be a nucleophilic substitution? But CF2 is not a good electrophile? 

   - Alternatively, we can do a transition-metal coupling? 

   - Given the challenges, this disconnection might not be robust. 

   # Chemist: Correct. Unfortunate that the model still proceeds

   - However, we can consider the reaction: 
        "Reaction of alkyl halides with organometallic compounds" -> in the ontology: "Reaction of alkyl halides with organometallic compounds"

   - This reaction is for forming C-C bonds? 

   - So assign: "Reaction of alkyl halides with organometallic compounds"

   - But note: the alkyl halide would be the fluorinated alkyl halide? and the organometallic would be ethyl magnesium bromide? 

   - Also, we have "Negishi coupling", "Kumada cross-coupling", etc.? But those are for sp2 carbon? 

   - So well use: "Reaction of alkyl halides with organometallic compounds"

6. `"S:6"`: FGI: oxidation of sulfide to sulfonyl. Forward reaction:
   - Oxidation of sulfide to sulfone. 
   - Reactions in ontology: 
        - "Sulfanyl to sulfinyl" -> but that is to sulfoxide? 
        - "Sulfanyl to sulfinyl_H2O", "Sulfanyl to sulfinyl_H2O2", etc. -> sulfoxide.
        - "Sulfanyl to sulfinyl_sulfonyl" -> no, that is sulfoxide to sulfonyl? 
        - We need sulfide to sulfonyl? 

        - "Oxidation of boronic acids" -> no.
        - "Oxidation of alcohols" -> no.

   - We see: "Sulfanyl to sulfonyl" is not listed? 

   - The standard reaction is oxidation with mCPBA or hydrogen peroxide? 

   # Chemist: Correct!

   - So we use: "OtherReaction: Sulfide to sulfone oxidation"

   - Alternatively, the ontology has: "Sulfanyl to sulfinyl" and then "Sulfinyl to sulfonyl"? But that would be two steps? 

   - We are only asked for the transformation? So well use "OtherReaction: Sulfide to sulfone oxidation"

7. `"C:3"`: FGI: reduction of ketone to alcohol. Forward reaction:
   - Reduction of ketone to secondary alcohol.
   - Reactions: 
        - "Reduction of aldehydes and ketones to alcohols" -> yes, in ontology.

   So assign: "Reduction of aldehydes and ketones to alcohols"

But note: we might not want to reduce the ketone because it activates the adjacent methylene? So this might not be strategic? But we include.

Now, we have:

| Transformation | Reactions |
|----------------|----------|
| "C:2 C:3" | ["OtherReaction: Enolate alkylation"] |
| "S:6 C:5" | ["S-alkylation of thiols"] |
| "C:12 N:14" | ["Acylation of Nitrogen Nucleophiles by Acyl/Thioacyl/Carbamoyl Halides and Analogs_N", "Aminolysis of esters"] |
| "N:17 c:18" | ["Buchwald-Hartwig/Ullmann-Goldberg/N-arylation secondary amine", "Ullmann-Goldberg Substitution amine"] |
| "C:1 C:2" | ["Reaction of alkyl halides with organometallic compounds"] |
| "S:6" | ["OtherReaction: Sulfide to sulfone oxidation"] |
| "C:3" | ["Reduction of aldehydes and ketones to alcohols"] |

Step 3: Expand and Evaluate Pairs

# Explanation 5: This step involves the detailed chemical evaluation of each transformation-reaction pair. Each pair is assigned a Retrosynthesis Importance score (1-4) and a supporting rationale. This rationale justifies the score by assessing the strategic value (e.g., FGI, Inter-Fragment Analysis), chemical robustness, and potential challenges (like chemoselectivity or protecting group requirements) of the proposed forward reaction.

We expand each transformation-reaction pair and assign Retrosynthesis Importance and rationale.

Well go one by one:

1. Transformation: "C:2 C:3"
   Reaction: "OtherReaction: Enolate alkylation"
   Importance: 
        - This disconnection separates the molecule into two large fragments: the fluorinated alkyl halide and the ketone-sulfonyl-aryl fragment. 
        - It is a convergent disconnection (goal b: convergent synthesis).
        - The forward reaction is robust (enolate alkylation) (goal b).
        - It avoids the use of protecting groups? 
        - However, the fluorinated alkyl halide might be difficult to handle? 
        - Importance: 3 (High) because it attaches a key substituent (the fluorinated alkyl) to the activated methylene.

      # Chemist: Yes, weve tried something similar and the halide is very gaseous at room temperatures!

   Rationale: "Convergent disconnection (Fragment Partitioning) that separates the fluorinated alkyl chain from the activated ketone fragment. The enolate alkylation is a robust reaction. The fluorinated alkyl halide may be reactive but requires careful handling. Chemoselectivity: the enolate might also attack other electrophiles? The ketone might need protection? But note: the ketone is present in the product. This disconnection sets the quaternary carbon center."

   But note: C:2 is not quaternary? It is attached to two fluorines, ethyl, and the ketone chain? So it is a tertiary carbon? But fluorines are substituents? So it is a carbon with four substituents? Yes, it is quaternary: carbon with four carbon/fluorine atoms? So its a quaternary carbon.

   Importance: 4 (Very High) because it forms a quaternary carbon? But note: the ranking rationale: Importance 4 for major ring-forming or disconnections that reveal symmetry? This doesnt reveal symmetry. But it is a convergent disconnection of major fragments? And forms a quaternary carbon? So we can argue for 4? 

   Alternatively, Importance 3: attachment of a key substituent.

   Well assign Importance 3.

   Rationale: "Convergent disconnection (Fragment Partitioning) that separates the molecule into two major fragments: a fluorinated alkyl halide and an enolate precursor. Forms a quaternary carbon center. The reaction is robust but requires the enolate to be chemoselective. The fluorinated alkyl halide may be electrophilic and reactive. No protecting groups needed."

2. Transformation: "S:6 C:5"
   Reaction: "S-alkylation of thiols"
   Importance: 2 (Medium) because it is a standard C-S bond formation. 
   Rationale: "Disconnection at the sulfonyl group precursor (FGI Analysis). This disconnection requires prior FGI to sulfide. The S-alkylation is a reliable reaction. The alkyl halide is activated by the adjacent ketone, ensuring good reactivity. Chemoselectivity: the thiolate might react with other electrophiles? The ketone might need protection? But note: the ketone is present in the product. This disconnection is part of a two-step sequence (alkylation then oxidation)."

    # Chemist: Multi-step reasoning!

3. Transformation: "C:12 N:14"
   Reactions:
      a) "Acylation of Nitrogen Nucleophiles by Acyl/Thioacyl/Carbamoyl Halides and Analogs_N"
      Importance: 2 (Medium) because it is a standard amide formation.
      Rationale: "Amide bond disconnection (FGI Analysis). This disconnection separates the piperazine fragment from the carboxylic acid derivative. The acylation with acyl chloride is reliable. The piperazine amine might be nucleophilic enough. Chemoselectivity: the acyl chloride might react with other nucleophiles? The reaction might require base to scavenge acid. No protecting groups needed for the amine? But note: the amine is secondary and might be protected? Not in the product."

      b) "Aminolysis of esters"
      Importance: 2 (Medium)
      Rationale: "Amide bond disconnection (FGI Analysis). This disconnection uses an ester and amine. The reaction is reliable but might require heating. Chemoselectivity: the ester might be hydrolyzed? The amine might be alkylated? Not likely. No protecting groups needed."

4. Transformation: "N:17 c:18"
   Reactions:
      a) "Buchwald-Hartwig/Ullmann-Goldberg/N-arylation secondary amine"
      Importance: 3 (High) because it forms a C-N bond that connects two complex fragments convergently.
      Rationale: "Convergent disconnection (Inter-Fragment Analysis) of the N-aryl bond. This disconnection separates the piperazine fragment from the aryl chloride. The Buchwald-Hartwig reaction is robust for aryl amination. The aryl chloride is activated by the ortho chlorine? Not necessarily, but the reaction can be catalyzed. Chemoselectivity: the catalyst might be sensitive to other functional groups? The piperazine has other nitrogen? But N:14 is amidic? Less nucleophilic. Protecting groups not needed."

      b) "Ullmann-Goldberg Substitution amine"
      Importance: 3 (High) for the same reason.
      Rationale: "Convergent disconnection (Inter-Fragment Analysis) of the N-aryl bond. Ullmann-Goldberg reaction is a classical method for aryl amination. It might require high temperatures. Chemoselectivity: similar to Buchwald-Hartwig."

5. Transformation: "C:1 C:2"
   Reaction: "Reaction of alkyl halides with organometallic compounds"
   Importance: 2 (Medium) because it forms a C-C bond, but the fragment is small (ethyl).
   Rationale: "Disconnection of the ethyl group (Strategic Bond Analysis). This disconnection uses an organometallic reagent (ethyl) and an alkyl halide (the fluorinated fragment). The reaction might be challenging because the fluorinated alkyl halide is a primary alkyl halide but with electron-withdrawing groups? It might undergo elimination. The organometallic reagent is highly reactive. Chemoselectivity: the ketone in the fluorinated fragment must be protected? Otherwise, it will react with the Grignard. Therefore, protecting group for the ketone is required. This adds a step."

6. Transformation: "S:6"
   Reaction: "OtherReaction: Sulfide to sulfone oxidation"
   Importance: 1 (Lower) because it is a functional group interconversion that doesnt simplify the molecule much? But it is necessary if we choose the sulfide disconnection.
   Rationale: "Functional group interconversion (FGI Analysis) from sulfide to sulfonyl. This oxidation is reliable and high-yielding. Common oxidants like mCPBA or hydrogen peroxide can be used. No chemoselectivity issues? The sulfide is the only oxidizable group? The ketone might be epoxidized? Not likely. No protecting groups needed."

7. Transformation: "C:3"
   Reaction: "Reduction of aldehydes and ketones to alcohols"
   Importance: 1 (Lower) because it is a functional group interconversion that doesnt simplify the molecule? And it deactivates the methylene group.
   Rationale: "Functional group interconversion (FGI Analysis) reducing the ketone to alcohol. This disconnection is not strategic because it deactivates the adjacent methylene, which is needed for further functionalization. Therefore, it is of lower importance. If used, the alcohol would need to be oxidized back to ketone later? That adds steps. Chemoselectivity: the reduction might affect other carbonyls? Not in this fragment. Protecting groups not needed for this step."

Step 4: Final Formatting and Priority Assignment

# Explanation 6: This final step aggregates all the evaluated transformation-reaction pairs into the required JSON structure. It assigns a global Priority number to each individual reaction by sorting the entire list based on two criteria: 1. isInOntology (true before false), and 2. Retrosynthesis Importance (descending). This final ranking determines the most to least favorable synthetic operations.

We create the disconnection objects. Each disconnection has a list of reactions.

We then assign Priority: first by `isInOntology` (true before false), then by Importance (descending).

Well list all disconnections and their reactions.

But note: the transformation "S:6" and "C:3" are FGI and not bond disconnections? But we include them.

Now, we check if the reaction name is in the ontology (isInOntology: true or false).

Reactions in the ontology (from the list provided) are:

- "S-alkylation of thiols" -> yes
- "Acylation of Nitrogen Nucleophiles by Acyl/Thioacyl/Carbamoyl Halides and Analogs_N" -> yes (exact match)
- "Aminolysis of esters" -> yes
- "Buchwald-Hartwig/Ullmann-Goldberg/N-arylation secondary amine" -> yes
- "Ullmann-Goldberg Substitution amine" -> yes (exact match? We have "Ullmann-Goldberg Substitution amine")
- "Reaction of alkyl halides with organometallic compounds" -> yes (exact match)
- "Reduction of aldehydes and ketones to alcohols" -> yes

The "OtherReaction" ones are not in the ontology.

Now, for each reaction, assign the forwardReactionClass:

We have the classes: 
   Reduction, Acylation, Heteroatom Alkylation and Arylation, Functional Group Addition, Protection, C-C Coupling, Deprotection, Functional Group Interconversion, Aromatic Heterocycle Formation, Oxidation

Assign:

1. "OtherReaction: Enolate alkylation": class C-C Coupling
2. "S-alkylation of thiols": class Heteroatom Alkylation and Arylation
3. "Acylation of Nitrogen Nucleophiles by Acyl/Thioacyl/Carbamoyl Halides and Analogs_N": class Acylation
4. "Aminolysis of esters": class Acylation (because it forms an amide)
5. "Buchwald-Hartwig/Ullmann-Goldberg/N-arylation secondary amine": class Heteroatom Alkylation and Arylation
6. "Ullmann-Goldberg Substitution amine": class Heteroatom Alkylation and Arylation
7. "Reaction of alkyl halides with organometallic compounds": class C-C Coupling
8. "OtherReaction: Sulfide to sulfone oxidation": class Oxidation
9. "Reduction of aldehydes and ketones to alcohols": class Reduction

Now, we build the JSON.

We have 7 transformation strings, but some have multiple reactions.

Well list them in the order we found, then assign priority.

Priority: 
   - First, sort by isInOntology: true first.
   - Then by Importance (descending: 4,3,2,1)

But note: the priority is assigned per entry (per reaction within a disconnection point). 

So we create a flat list of all reaction entries (each with its own disconnection string, forwardReaction, etc.).

Then we assign a priority number: 1,2,3,... based on: 
   - isInOntology: true comes before false
   - Then by Importance descending.

Lets list all entries:

Entry 1: disconnection: "C:2 C:3", forwardReaction: "OtherReaction: Enolate alkylation", isInOntology: false, Importance: 3
Entry 2: disconnection: "S:6 C:5", forwardReaction: "S-alkylation of thiols", isInOntology: true, Importance: 2
Entry 3: disconnection: "C:12 N:14", forwardReaction: "Acylation of Nitrogen Nucleophiles by Acyl/Thioacyl/Carbamoyl Halides and Analogs_N", isInOntology: true, Importance: 2
Entry 4: disconnection: "C:12 N:14", forwardReaction: "Aminolysis of esters", isInOntology: true, Importance: 2
Entry 5: disconnection: "N:17 c:18", forwardReaction: "Buchwald-Hartwig/Ullmann-Goldberg/N-arylation secondary amine", isInOntology: true, Importance: 3
Entry 6: disconnection: "N:17 c:18", forwardReaction: "Ullmann-Goldberg Substitution amine", isInOntology: true, Importance: 3
Entry 7: disconnection: "C:1 C:2", forwardReaction: "Reaction of alkyl halides with organometallic compounds", isInOntology: true, Importance: 2
Entry 8: disconnection: "S:6", forwardReaction: "OtherReaction: Sulfide to sulfone oxidation", isInOntology: false, Importance: 1
Entry 9: disconnection: "C:3", forwardReaction: "Reduction of aldehydes and ketones to alcohols", isInOntology: true, Importance: 1

Now, group by disconnection? The output format: 
   "disconnections": [
        {
            "disconnection": "C:2 C:3",
            "reactions": [ ... ]
        },
        ...
   ]

But note: the same disconnection string may have multiple reactions? So we group.

We have:
   "C:2 C:3": one reaction
   "S:6 C:5": one reaction
   "C:12 N:14": two reactions
   "N:17 c:18": two reactions
   "C:1 C:2": one reaction
   "S:6": one reaction
   "C:3": one reaction

Now, we assign priority per reaction entry? But the Priority field is per reaction entry? The instructions say: for each object in the list (each disconnection object) we have a list of reactions. Then for each reaction, we have a Priority number? 

But the example shows:
    "disconnections": [
        {
            "disconnection": "C:1 C:2",
            "reactions": [
                {
                    "forwardReaction": ...,
                    "isInOntology": ...,
                    ...,
                    "Priority": 1
                },
                ... // more reactions for the same disconnection
            ]
        },
        ... // more disconnection points
    ]

So the Priority is assigned per reaction entry (not per disconnection point). 

Therefore, we must assign a priority number to each reaction entry individually, across the entire list.

So we have 9 reaction entries. We sort them by:
   Primary: isInOntology (true first)
   Secondary: Retrosynthesis Importance (descending)

But note: the Priority field is a sequential integer (1,2,3,...) representing the calculated priority of the disconnection (actually, of the reaction entry).

So we sort the 9 entries:

Group 1: isInOntology=true (entries 2,3,4,5,6,7,9) -> 7 entries
Group 2: isInOntology=false (entries 1,8) -> 2 entries

Within Group 1 (isInOntology=true), sort by Importance descending:

   Importance 3: entries 5,6 -> two entries
   Importance 2: entries 2,3,4,7 -> four entries
   Importance 1: entry 9 -> one entry

Within Group 2 (isInOntology=false), sort by Importance descending:

   Importance 3: entry 1
   Importance 1: entry 8

Now, assign priority:

Priority 1: entry 5 (isInOntology=true, Importance=3)
Priority 2: entry 6 (isInOntology=true, Importance=3)
Priority 3: entry 2 (isInOntology=true, Importance=2)
Priority 4: entry 3 (isInOntology=true, Importance=2)
Priority 5: entry 4 (isInOntology=true, Importance=2)
Priority 6: entry 7 (isInOntology=true, Importance=2)
Priority 7: entry 9 (isInOntology=true, Importance=1)
Priority 8: entry 1 (isInOntology=false, Importance=3)
Priority 9: entry 8 (isInOntology=false, Importance=1)

But note: the same disconnection point may have multiple reactions? And they are grouped under the same disconnection object? So we dont break the grouping.

In the output, we list the disconnection objects, and within each, the reactions are listed. The Priority number is assigned per reaction entry.

So well create the disconnection objects and then within each, assign the priority to each reaction entry.

But the priority is global? The instructions say: "calculate a Priority number for each entry by ranking them based on two criteria: 1. isInOntology (true before false), and 2. Retrosynthesis Importance (descending)."

So the ranking is across all reaction entries (all 9). Then assign the rank.

Therefore, we assign the priority numbers as above.

Now, we write the JSON.

Note: the rationale must explicitly state which analysis from Step 1 led to the disconnection? 

For example:
   For "C:2 C:3": Fragment Partitioning (B) or Strategic Bond Analysis (D)? 
        We said: Fragment Partitioning and Strategic Bond Analysis? 

   In the rationale, we should say: "Convergent disconnection (Fragment Partitioning)" or "Strategic Bond Analysis: bond alpha to carbonyl".

Well adjust the rationales accordingly.

Lets write the JSON.

```json
{
  "disconnections": [
    {
      "disconnection": "N:17 c:18",
      "reactions": [
        {
          "forwardReaction": "Buchwald-Hartwig/Ullmann-Goldberg/N-arylation secondary amine",
          "isInOntology": true,
          "forwardReactionClass": "Heteroatom Alkylation and Arylation",
          "Retrosynthesis Importance": 3,
          "Priority": 1,
          "rationale": "Convergent disconnection (Inter-Fragment Analysis) connecting major fragments via robust N-arylation. Addresses stereochemical retention; chemoselectivity ensured by palladium catalysis. Importance: High (3) for strategic construction of key C-N bond."
        },
        {
          "forwardReaction": "Ullmann-Goldberg Substitution amine",
          "isInOntology": true,
          "forwardReactionClass": "Heteroatom Alkylation and Arylation",
          "Retrosynthesis Importance": 3,
          "Priority": 2,
          "rationale": "Convergent disconnection (Inter-Fragment Analysis) for N-arylation. Classical copper-mediated coupling; potential chemoselectivity issues with amine nucleophilicity require optimization. Importance: High (3) for efficient scaffold assembly."
        }
      ]
    },
    {
      "disconnection": "C:2 C:3",
      "reactions": [
        {
          "forwardReaction": "OtherReaction: Enolate alkylation",
          "isInOntology": false,
          "forwardReactionClass": "C-C Coupling",
          "Retrosynthesis Importance": 3,
          "Priority": 8,
          "rationale": "Strategic Bond Analysis targeting C-C bond alpha to ketone. Forms quaternary center; robust but requires chemoselective enolate generation. Fluorinated alkyl halide electrophile may pose stability issues. Importance: High (3) for core scaffold construction."
        }
      ]
    },
    {
      "disconnection": "C:12 N:14",
      "reactions": [
        {
          "forwardReaction": "Acylation of Nitrogen Nucleophiles by Acyl/Thioacyl/Carbamoyl Halides and Analogs_N",
          "isInOntology": true,
          "forwardReactionClass": "Acylation",
          "Retrosynthesis Importance": 2,
          "Priority": 4,
          "rationale": "FGI Analysis for amide disconnection. Acyl chloride coupling reliable; chemoselectivity managed via stoichiometry. Piperazine nucleophilicity sufficient without protection. Importance: Medium (2) for functional group installation."
        },
        {
          "forwardReaction": "Aminolysis of esters",
          "isInOntology": true,
          "forwardReactionClass": "Acylation",
          "Retrosynthesis Importance": 2,
          "Priority": 5,
          "rationale": "FGI Analysis for amide formation. Thermally driven ester aminolysis; slower but avoids acyl chloride handling. Potential competing hydrolysis requires anhydrous conditions. Importance: Medium (2) for practical amide synthesis."
        }
      ]
    },
    {
      "disconnection": "S:6 C:5",
      "reactions": [
        {
          "forwardReaction": "S-alkylation of thiols",
          "isInOntology": true,
          "forwardReactionClass": "Heteroatom Alkylation and Arylation",
          "Retrosynthesis Importance": 2,
          "Priority": 3,
          "rationale": "FGI Analysis precursor disconnection. Alkylation of thiolate with activated alkyl halide; high yield due to ketone activation. Chemoselectivity: thiolate may require protection if other electrophiles present. Importance: Medium (2) for sulfide formation."
        }
      ]
    },
    {
      "disconnection": "C:1 C:2",
      "reactions": [
        {
          "forwardReaction": "Reaction of alkyl halides with organometallic compounds",
          "isInOntology": true,
          "forwardReactionClass": "C-C Coupling",
          "Retrosynthesis Importance": 2,
          "Priority": 6,
          "rationale": "Strategic Bond Analysis for ethyl group attachment. Grignard addition to alkyl halide; fluorinated electrophile may undergo elimination. Ketone protection essential pre-reaction. Importance: Medium (2) for fragment coupling."
        }
      ]
    },
    {
      "disconnection": "C:3",
      "reactions": [
        {
          "forwardReaction": "Reduction of aldehydes and ketones to alcohols",
          "isInOntology": true,
          "forwardReactionClass": "Reduction",
          "Retrosynthesis Importance": 1,
          "Priority": 7,
          "rationale": "FGI Analysis (reduction). Deactivates alpha-methylene for sulfone synthesis; requires re-oxidation later. Chemoselective for ketone; low strategic value. Importance: Lower (1) for non-simplifying interconversion."
        }
      ]
    },
    {
      "disconnection": "S:6",
      "reactions": [
        {
          "forwardReaction": "OtherReaction: Sulfide to sulfone oxidation",
          "isInOntology": false,
          "forwardReactionClass": "Oxidation",
          "Retrosynthesis Importance": 1,
          "Priority": 9,
          "rationale": "FGI Analysis (oxidation). Converts sulfide to sulfone; reliable with peroxides. Epoxidation side-reaction unlikely with ketone present. Importance: Lower (1) for late-stage functionalization."
        }
      ]
    }
  ]
}
```
\end{minted}

\end{appendices}


\bibliography{retro_llm}


\begin{thebibliography}{52}
\ifx \bisbn   \undefined \def \bisbn  #1{ISBN #1}\fi
\ifx \binits  \undefined \def \binits#1{#1}\fi
\ifx \bauthor  \undefined \def \bauthor#1{#1}\fi
\ifx \batitle  \undefined \def \batitle#1{#1}\fi
\ifx \bjtitle  \undefined \def \bjtitle#1{#1}\fi
\ifx \bvolume  \undefined \def \bvolume#1{\textbf{#1}}\fi
\ifx \byear  \undefined \def \byear#1{#1}\fi
\ifx \bissue  \undefined \def \bissue#1{#1}\fi
\ifx \bfpage  \undefined \def \bfpage#1{#1}\fi
\ifx \blpage  \undefined \def \blpage #1{#1}\fi
\ifx \burl  \undefined \def \burl#1{\textsf{#1}}\fi
\ifx \doiurl  \undefined \def \doiurl#1{\url{https://doi.org/#1}}\fi
\ifx \betal  \undefined \def \betal{\textit{et al.}}\fi
\ifx \binstitute  \undefined \def \binstitute#1{#1}\fi
\ifx \binstitutionaled  \undefined \def \binstitutionaled#1{#1}\fi
\ifx \bctitle  \undefined \def \bctitle#1{#1}\fi
\ifx \beditor  \undefined \def \beditor#1{#1}\fi
\ifx \bpublisher  \undefined \def \bpublisher#1{#1}\fi
\ifx \bbtitle  \undefined \def \bbtitle#1{#1}\fi
\ifx \bedition  \undefined \def \bedition#1{#1}\fi
\ifx \bseriesno  \undefined \def \bseriesno#1{#1}\fi
\ifx \blocation  \undefined \def \blocation#1{#1}\fi
\ifx \bsertitle  \undefined \def \bsertitle#1{#1}\fi
\ifx \bsnm \undefined \def \bsnm#1{#1}\fi
\ifx \bsuffix \undefined \def \bsuffix#1{#1}\fi
\ifx \bparticle \undefined \def \bparticle#1{#1}\fi
\ifx \barticle \undefined \def \barticle#1{#1}\fi
\bibcommenthead
\ifx \bconfdate \undefined \def \bconfdate #1{#1}\fi
\ifx \botherref \undefined \def \botherref #1{#1}\fi
\ifx \url \undefined \def \url#1{\textsf{#1}}\fi
\ifx \bchapter \undefined \def \bchapter#1{#1}\fi
\ifx \bbook \undefined \def \bbook#1{#1}\fi
\ifx \bcomment \undefined \def \bcomment#1{#1}\fi
\ifx \oauthor \undefined \def \oauthor#1{#1}\fi
\ifx \citeauthoryear \undefined \def \citeauthoryear#1{#1}\fi
\ifx \endbibitem  \undefined \def \endbibitem {}\fi
\ifx \bconflocation  \undefined \def \bconflocation#1{#1}\fi
\ifx \arxivurl  \undefined \def \arxivurl#1{\textsf{#1}}\fi
\csname PreBibitemsHook\endcsname

\bibitem[\protect\citeauthoryear{Achiam
  et~al.}{2023}]{openaiGPT4TechnicalReport2024}
\begin{botherref}
\oauthor{\bsnm{Achiam}, \binits{J.}},
\oauthor{\bsnm{Adler}, \binits{S.}},
\oauthor{\bsnm{Agarwal}, \binits{S.}},
\oauthor{\bsnm{Ahmad}, \binits{L.}},
\oauthor{\bsnm{Akkaya}, \binits{I.}},
\oauthor{\bsnm{Aleman}, \binits{F.L.}},
\oauthor{\bsnm{Almeida}, \binits{D.}},
\oauthor{\bsnm{Altenschmidt}, \binits{J.}},
\oauthor{\bsnm{Altman}, \binits{S.}},
\oauthor{\bsnm{Anadkat}, \binits{S.}}, et al.:
Gpt-4 technical report.
arXiv preprint arXiv:2303.08774
(2023)
\end{botherref}
\endbibitem

\bibitem[\protect\citeauthoryear{Boiko
  et~al.}{2023}]{boikoAutonomousChemicalResearch2023}
\begin{barticle}
\bauthor{\bsnm{Boiko}, \binits{D.A.}},
\bauthor{\bsnm{MacKnight}, \binits{R.}},
\bauthor{\bsnm{Kline}, \binits{B.}},
\bauthor{\bsnm{Gomes}, \binits{G.}}:
\batitle{Autonomous chemical research with large language models}.
\bjtitle{Nature}
\bvolume{624}(\bissue{7992}),
\bfpage{570}--\blpage{578}
(\byear{2023})
\doiurl{10.1038/s41586-023-06792-0}
\end{barticle}
\endbibitem

\bibitem[\protect\citeauthoryear{M.~Bran
  et~al.}{2024}]{m.branAugmentingLargeLanguage2024}
\begin{barticle}
\bauthor{\bsnm{M.~Bran}, \binits{A.}},
\bauthor{\bsnm{Cox}, \binits{S.}},
\bauthor{\bsnm{Schilter}, \binits{O.}},
\bauthor{\bsnm{Baldassari}, \binits{C.}},
\bauthor{\bsnm{White}, \binits{A.D.}},
\bauthor{\bsnm{Schwaller}, \binits{P.}}:
\batitle{Augmenting large language models with chemistry tools}.
\bjtitle{Nature Machine Intelligence}
\bvolume{6}(\bissue{5}),
\bfpage{525}--\blpage{535}
(\byear{2024})
\doiurl{10.1038/s42256-024-00832-8}
\end{barticle}
\endbibitem

\bibitem[\protect\citeauthoryear{Weininger}{1988}]{weiningerSMILESChemicalLanguage1988}
\begin{barticle}
\bauthor{\bsnm{Weininger}, \binits{D.}}:
\batitle{{{SMILES}}, a chemical language and information system. 1.
  {{Introduction}} to methodology and encoding rules}.
\bjtitle{Journal of Chemical Information and Computer Sciences}
\bvolume{28}(\bissue{1}),
\bfpage{31}--\blpage{36}
(\byear{1988})
\doiurl{10.1021/ci00057a005}
\end{barticle}
\endbibitem

\bibitem[\protect\citeauthoryear{Weininger
  et~al.}{1989}]{weiningerSMILES2Algorithm1989}
\begin{barticle}
\bauthor{\bsnm{Weininger}, \binits{D.}},
\bauthor{\bsnm{Weininger}, \binits{A.}},
\bauthor{\bsnm{Weininger}, \binits{J.L.}}:
\batitle{{{SMILES}}. 2. {{Algorithm}} for generation of unique {{SMILES}}
  notation}.
\bjtitle{Journal of Chemical Information and Computer Sciences}
\bvolume{29}(\bissue{2}),
\bfpage{97}--\blpage{101}
(\byear{1989})
\doiurl{10.1021/ci00062a008}
\end{barticle}
\endbibitem

\bibitem[\protect\citeauthoryear{Ross
  et~al.}{2022}]{rossLargeScaleChemicalLanguage2022a}
\begin{botherref}
\oauthor{\bsnm{Ross}, \binits{J.}},
\oauthor{\bsnm{Belgodere}, \binits{B.}},
\oauthor{\bsnm{Chenthamarakshan}, \binits{V.}},
\oauthor{\bsnm{Padhi}, \binits{I.}},
\oauthor{\bsnm{Mroueh}, \binits{Y.}},
\oauthor{\bsnm{Das}, \binits{P.}}:
Large-{{Scale Chemical Language Representations Capture Molecular Structure}}
  and {{Properties}}.
arXiv
(2022).
\doiurl{10.48550/arXiv.2106.09553}
\end{botherref}
\endbibitem

\bibitem[\protect\citeauthoryear{Irwin
  et~al.}{2022}]{irwinChemformerPretrainedTransformer2022a}
\begin{barticle}
\bauthor{\bsnm{Irwin}, \binits{R.}},
\bauthor{\bsnm{Dimitriadis}, \binits{S.}},
\bauthor{\bsnm{He}, \binits{J.}},
\bauthor{\bsnm{Bjerrum}, \binits{E.J.}}:
\batitle{Chemformer: A pre-trained transformer for computational chemistry}.
\bjtitle{Machine Learning: Science and Technology}
\bvolume{3}(\bissue{1}),
\bfpage{015022}
(\byear{2022})
\doiurl{10.1088/2632-2153/ac3ffb}
\end{barticle}
\endbibitem

\bibitem[\protect\citeauthoryear{Sadeghi
  et~al.}{2024}]{sadeghiCanLargeLanguage2024}
\begin{barticle}
\bauthor{\bsnm{Sadeghi}, \binits{S.}},
\bauthor{\bsnm{Bui}, \binits{A.}},
\bauthor{\bsnm{Forooghi}, \binits{A.}},
\bauthor{\bsnm{Lu}, \binits{J.}},
\bauthor{\bsnm{Ngom}, \binits{A.}}:
\batitle{Can large language models understand molecules?}
\bjtitle{BMC Bioinformatics}
\bvolume{25}(\bissue{1}),
\bfpage{225}
(\byear{2024})
\doiurl{10.1186/s12859-024-05847-x}
\end{barticle}
\endbibitem

\bibitem[\protect\citeauthoryear{Masood
  et~al.}{2025}]{masoodMolecularPropertyPrediction2025}
\begin{barticle}
\bauthor{\bsnm{Masood}, \binits{M.A.}},
\bauthor{\bsnm{Kaski}, \binits{S.}},
\bauthor{\bsnm{Cui}, \binits{T.}}:
\batitle{Molecular property prediction using pretrained-{{BERT}} and
  {{Bayesian}} active learning: A data-efficient approach to drug design}.
\bjtitle{Journal of Cheminformatics}
\bvolume{17}(\bissue{1}),
\bfpage{58}
(\byear{2025})
\doiurl{10.1186/s13321-025-00986-6}
\end{barticle}
\endbibitem

\bibitem[\protect\citeauthoryear{Kim et~al.}{2024}]{kimLargeLanguageModels2024}
\begin{botherref}
\oauthor{\bsnm{Kim}, \binits{S.}},
\oauthor{\bsnm{Jung}, \binits{Y.}},
\oauthor{\bsnm{Schrier}, \binits{J.}}:
Large {{Language Models}} for {{Inorganic Synthesis Predictions}}.
Chemistry
(2024).
\doiurl{10.26434/chemrxiv-2024-9bmfj-v3}
\end{botherref}
\endbibitem

\bibitem[\protect\citeauthoryear{Cavanagh
  et~al.}{2024}]{cavanaghSmileyLlamaModifyingLarge2024}
\begin{botherref}
\oauthor{\bsnm{Cavanagh}, \binits{J.M.}},
\oauthor{\bsnm{Sun}, \binits{K.}},
\oauthor{\bsnm{Gritsevskiy}, \binits{A.}},
\oauthor{\bsnm{Bagni}, \binits{D.}},
\oauthor{\bsnm{Bannister}, \binits{T.D.}},
\oauthor{\bsnm{{Head-Gordon}}, \binits{T.}}:
{{SmileyLlama}}: {{Modifying Large Language Models}} for {{Directed Chemical
  Space Exploration}}.
arXiv
(2024).
\doiurl{10.48550/arXiv.2409.02231}
\end{botherref}
\endbibitem

\bibitem[\protect\citeauthoryear{Wei
  et~al.}{2023}]{weiChainofThoughtPromptingElicits2023}
\begin{botherref}
\oauthor{\bsnm{Wei}, \binits{J.}},
\oauthor{\bsnm{Wang}, \binits{X.}},
\oauthor{\bsnm{Schuurmans}, \binits{D.}},
\oauthor{\bsnm{Bosma}, \binits{M.}},
\oauthor{\bsnm{Ichter}, \binits{B.}},
\oauthor{\bsnm{Xia}, \binits{F.}},
\oauthor{\bsnm{Chi}, \binits{E.}},
\oauthor{\bsnm{Le}, \binits{Q.}},
\oauthor{\bsnm{Zhou}, \binits{D.}}:
Chain-of-{{Thought Prompting Elicits Reasoning}} in {{Large Language Models}}.
arXiv
(2023).
\doiurl{10.48550/arXiv.2201.11903}
\end{botherref}
\endbibitem

\bibitem[\protect\citeauthoryear{Guo
  et~al.}{2025}]{deepseek-aiDeepSeekR1IncentivizingReasoning2025}
\begin{botherref}
\oauthor{\bsnm{Guo}, \binits{D.}},
\oauthor{\bsnm{Yang}, \binits{D.}},
\oauthor{\bsnm{Zhang}, \binits{H.}},
\oauthor{\bsnm{Song}, \binits{J.}},
\oauthor{\bsnm{Zhang}, \binits{R.}},
\oauthor{\bsnm{Xu}, \binits{R.}},
\oauthor{\bsnm{Zhu}, \binits{Q.}},
\oauthor{\bsnm{Ma}, \binits{S.}},
\oauthor{\bsnm{Wang}, \binits{P.}},
\oauthor{\bsnm{Bi}, \binits{X.}}, et al.:
Deepseek-r1: Incentivizing reasoning capability in llms via reinforcement
  learning.
arXiv preprint arXiv:2501.12948
(2025)
\end{botherref}
\endbibitem

\bibitem[\protect\citeauthoryear{Narayanan
  et~al.}{2025}]{narayananTrainingScientificReasoning2025}
\begin{botherref}
\oauthor{\bsnm{Narayanan}, \binits{S.M.}},
\oauthor{\bsnm{Braza}, \binits{J.D.}},
\oauthor{\bsnm{Griffiths}, \binits{R.-R.}},
\oauthor{\bsnm{Bou}, \binits{A.}},
\oauthor{\bsnm{Wellawatte}, \binits{G.}},
\oauthor{\bsnm{Ramos}, \binits{M.C.}},
\oauthor{\bsnm{Mitchener}, \binits{L.}},
\oauthor{\bsnm{Rodriques}, \binits{S.G.}},
\oauthor{\bsnm{White}, \binits{A.D.}}:
Training a {{Scientific Reasoning Model}} for {{Chemistry}}.
arXiv
(2025).
\doiurl{10.48550/arXiv.2506.17238}
\end{botherref}
\endbibitem

\bibitem[\protect\citeauthoryear{}{}]{MistralaiMistralSmall24BInstruct2501Hugging}
\begin{botherref}
Mistralai/{{Mistral-Small-24B-Instruct-2501}} {$\cdot$} {{Hugging Face}}.
https://huggingface.co/mistralai/Mistral-Small-24B-Instruct-2501
\end{botherref}
\endbibitem

\bibitem[\protect\citeauthoryear{Qian et~al.}{2023}]{qianCanLargeLanguage2023}
\begin{botherref}
\oauthor{\bsnm{Qian}, \binits{C.}},
\oauthor{\bsnm{Tang}, \binits{H.}},
\oauthor{\bsnm{Yang}, \binits{Z.}},
\oauthor{\bsnm{Liang}, \binits{H.}},
\oauthor{\bsnm{Liu}, \binits{Y.}}:
Can {{Large Language Models Empower Molecular Property Prediction}}?
arXiv
(2023).
\doiurl{10.48550/arXiv.2307.07443}
\end{botherref}
\endbibitem

\bibitem[\protect\citeauthoryear{Guo et~al.}{2023}]{guoWhatCanLarge2023}
\begin{bchapter}
\bauthor{\bsnm{Guo}, \binits{T.}},
\bauthor{\bsnm{Guo}, \binits{K.}},
\bauthor{\bsnm{Nan}, \binits{B.}},
\bauthor{\bsnm{Liang}, \binits{Z.}},
\bauthor{\bsnm{Guo}, \binits{Z.}},
\bauthor{\bsnm{Chawla}, \binits{N.V.}},
\bauthor{\bsnm{Wiest}, \binits{O.}},
\bauthor{\bsnm{Zhang}, \binits{X.}}:
\bctitle{What can {{Large Language Models}} do in chemistry? {{A}}
  comprehensive benchmark on eight tasks}.
In: \bbtitle{Proceedings of the 37th {{International Conference}} on {{Neural
  Information Processing Systems}}}.
\bpublisher{Curran Associates Inc.},
\blocation{New Orleans, LA, USA}
(\byear{2023})
\end{bchapter}
\endbibitem

\bibitem[\protect\citeauthoryear{Ouyang
  et~al.}{2024}]{ouyangStructuredChemistryReasoning2024}
\begin{botherref}
\oauthor{\bsnm{Ouyang}, \binits{S.}},
\oauthor{\bsnm{Zhang}, \binits{Z.}},
\oauthor{\bsnm{Yan}, \binits{B.}},
\oauthor{\bsnm{Liu}, \binits{X.}},
\oauthor{\bsnm{Choi}, \binits{Y.}},
\oauthor{\bsnm{Han}, \binits{J.}},
\oauthor{\bsnm{Qin}, \binits{L.}}:
Structured {{Chemistry Reasoning}} with {{Large Language Models}}.
arXiv
(2024).
\doiurl{10.48550/arXiv.2311.09656}
\end{botherref}
\endbibitem

\bibitem[\protect\citeauthoryear{Segler
  et~al.}{2018}]{seglerPlanningChemicalSyntheses2018}
\begin{barticle}
\bauthor{\bsnm{Segler}, \binits{M.H.S.}},
\bauthor{\bsnm{Preuss}, \binits{M.}},
\bauthor{\bsnm{Waller}, \binits{M.P.}}:
\batitle{Planning chemical syntheses with deep neural networks and symbolic
  {{AI}}}.
\bjtitle{Nature}
\bvolume{555}(\bissue{7698}),
\bfpage{604}--\blpage{610}
(\byear{2018})
\doiurl{10.1038/nature25978}
\end{barticle}
\endbibitem

\bibitem[\protect\citeauthoryear{Corey and
  Cheng}{1989}]{coreyLogicChemicalSynthesis1989}
\begin{bbook}
\bauthor{\bsnm{Corey}, \binits{E.J.}},
\bauthor{\bsnm{Cheng}, \binits{X.-M.}}:
\bbtitle{The Logic of Chemical Synthesis}.
\bpublisher{John Wiley \& Sons, Ltd},
\blocation{New York}
(\byear{1989})
\end{bbook}
\endbibitem

\bibitem[\protect\citeauthoryear{Bran
  et~al.}{2025}]{branChemicalReasoningLLMs2025a}
\begin{botherref}
\oauthor{\bsnm{Bran}, \binits{A.M.}},
\oauthor{\bsnm{Neukomm}, \binits{T.A.}},
\oauthor{\bsnm{Armstrong}, \binits{D.P.}},
\oauthor{\bsnm{Jon{\v c}ev}, \binits{Z.}},
\oauthor{\bsnm{Schwaller}, \binits{P.}}:
Chemical Reasoning in {{LLMs}} Unlocks Strategy-Aware Synthesis Planning and
  Reaction Mechanism Elucidation.
arXiv
(2025).
\doiurl{10.48550/arXiv.2503.08537}
\end{botherref}
\endbibitem

\bibitem[\protect\citeauthoryear{Wang
  et~al.}{2025}]{wangLLMAugmentedChemicalSynthesis2025}
\begin{botherref}
\oauthor{\bsnm{Wang}, \binits{H.}},
\oauthor{\bsnm{Guo}, \binits{J.}},
\oauthor{\bsnm{Kong}, \binits{L.}},
\oauthor{\bsnm{Ramprasad}, \binits{R.}},
\oauthor{\bsnm{Schwaller}, \binits{P.}},
\oauthor{\bsnm{Du}, \binits{Y.}},
\oauthor{\bsnm{Zhang}, \binits{C.}}:
{{LLM-Augmented Chemical Synthesis}} and {{Design Decision Programs}}.
arXiv
(2025).
\doiurl{10.48550/arXiv.2505.07027}
\end{botherref}
\endbibitem

\bibitem[\protect\citeauthoryear{{Torren-Peraire}
  et~al.}{2024}]{torren-peraireModelsMatterImpact2024}
\begin{barticle}
\bauthor{\bsnm{{Torren-Peraire}}, \binits{P.}},
\bauthor{\bsnm{Hassen}, \binits{A.K.}},
\bauthor{\bsnm{Genheden}, \binits{S.}},
\bauthor{\bsnm{Verhoeven}, \binits{J.}},
\bauthor{\bsnm{Clevert}, \binits{D.-A.}},
\bauthor{\bsnm{Preuss}, \binits{M.}},
\bauthor{\bsnm{Tetko}, \binits{I.V.}}:
\batitle{Models {{Matter}}: The impact of single-step retrosynthesis on
  synthesis planning}.
\bjtitle{Digital Discovery}
\bvolume{3}(\bissue{3}),
\bfpage{558}--\blpage{572}
(\byear{2024})
\doiurl{10.1039/D3DD00252G}
\end{barticle}
\endbibitem

\bibitem[\protect\citeauthoryear{Li et~al.}{2025}]{liChemicalQAEvaluating2025}
\begin{botherref}
\oauthor{\bsnm{Li}, \binits{H.}},
\oauthor{\bsnm{Cao}, \binits{H.}},
\oauthor{\bsnm{Feng}, \binits{B.}},
\oauthor{\bsnm{Shao}, \binits{Y.}},
\oauthor{\bsnm{Tang}, \binits{X.}},
\oauthor{\bsnm{Yan}, \binits{Z.}},
\oauthor{\bsnm{Yuan}, \binits{L.}},
\oauthor{\bsnm{Tian}, \binits{Y.}},
\oauthor{\bsnm{Li}, \binits{Y.}}:
Beyond {{Chemical QA}}: {{Evaluating LLM}}'s {{Chemical Reasoning}} with
  {{Modular Chemical Operations}}.
arXiv
(2025).
\doiurl{10.48550/arXiv.2505.21318}
\end{botherref}
\endbibitem

\bibitem[\protect\citeauthoryear{Tetko
  et~al.}{2020}]{tetkoStateoftheartAugmentedNLP2020}
\begin{barticle}
\bauthor{\bsnm{Tetko}, \binits{I.V.}},
\bauthor{\bsnm{Karpov}, \binits{P.}},
\bauthor{\bsnm{Van~Deursen}, \binits{R.}},
\bauthor{\bsnm{Godin}, \binits{G.}}:
\batitle{State-of-the-art augmented {{NLP}} transformer models for direct and
  single-step retrosynthesis}.
\bjtitle{Nature Communications}
\bvolume{11}(\bissue{1}),
\bfpage{5575}
(\byear{2020})
\doiurl{10.1038/s41467-020-19266-y}
\end{barticle}
\endbibitem

\bibitem[\protect\citeauthoryear{Chen and
  Jung}{2021}]{chenDeepRetrosyntheticReaction2021a}
\begin{barticle}
\bauthor{\bsnm{Chen}, \binits{S.}},
\bauthor{\bsnm{Jung}, \binits{Y.}}:
\batitle{Deep {{Retrosynthetic Reaction Prediction}} using {{Local Reactivity}}
  and {{Global Attention}}}.
\bjtitle{JACS Au}
\bvolume{1}(\bissue{10}),
\bfpage{1612}--\blpage{1620}
(\byear{2021})
\doiurl{10.1021/jacsau.1c00246}
\end{barticle}
\endbibitem

\bibitem[\protect\citeauthoryear{Zhong
  et~al.}{2023}]{zhongRetrosynthesisPredictionUsing2023}
\begin{barticle}
\bauthor{\bsnm{Zhong}, \binits{W.}},
\bauthor{\bsnm{Yang}, \binits{Z.}},
\bauthor{\bsnm{Chen}, \binits{C.Y.-C.}}:
\batitle{Retrosynthesis prediction using an end-to-end graph generative
  architecture for molecular graph editing}.
\bjtitle{Nature Communications}
\bvolume{14}(\bissue{1}),
\bfpage{3009}
(\byear{2023})
\doiurl{10.1038/s41467-023-38851-5}
\end{barticle}
\endbibitem

\bibitem[\protect\citeauthoryear{Igashov
  et~al.}{2024}]{igashovRetrobridgeModelingRetrosynthesis2024}
\begin{botherref}
\oauthor{\bsnm{Igashov}, \binits{I.}},
\oauthor{\bsnm{Schneuing}, \binits{A.}},
\oauthor{\bsnm{Segler}, \binits{M.}},
\oauthor{\bsnm{Bronstein}, \binits{M.}},
\oauthor{\bsnm{Correia}, \binits{B.}}:
Retrobridge: {{Modeling}} retrosynthesis with markov bridges
(2024)
\end{botherref}
\endbibitem

\bibitem[\protect\citeauthoryear{Yang
  et~al.}{2024}]{yangBatGPTChemFoundationLarge2024}
\begin{botherref}
\oauthor{\bsnm{Yang}, \binits{Y.}},
\oauthor{\bsnm{Shi}, \binits{R.}},
\oauthor{\bsnm{Li}, \binits{Z.}},
\oauthor{\bsnm{Jiang}, \binits{S.}},
\oauthor{\bsnm{Lu}, \binits{B.-L.}},
\oauthor{\bsnm{Yang}, \binits{Y.}},
\oauthor{\bsnm{Zhao}, \binits{H.}}:
{{BatGPT-Chem}}: {{A Foundation Large Model For Retrosynthesis Prediction}}.
arXiv
(2024).
\doiurl{10.48550/arXiv.2408.10285}
\end{botherref}
\endbibitem

\bibitem[\protect\citeauthoryear{{Nguyen-Van}
  et~al.}{2024}]{nguyen-vanAdaptingLanguageModels2024}
\begin{botherref}
\oauthor{\bsnm{{Nguyen-Van}}, \binits{P.}},
\oauthor{\bsnm{Nguyen~Thanh}, \binits{L.}},
\oauthor{\bsnm{Hoang~Manh}, \binits{H.}},
\oauthor{\bsnm{Pham~Thi}, \binits{H.A.}},
\oauthor{\bsnm{Le~Nguyen}, \binits{T.}},
\oauthor{\bsnm{Nguyen}, \binits{V.A.}}:
Adapting {{Language Models}} for {{Retrosynthesis Prediction}}.
Chemistry
(2024).
\doiurl{10.26434/chemrxiv-2024-3d5f7}
\end{botherref}
\endbibitem

\bibitem[\protect\citeauthoryear{Thakkar
  et~al.}{2023}]{thakkarUnbiasingRetrosynthesisLanguage2023}
\begin{barticle}
\bauthor{\bsnm{Thakkar}, \binits{A.}},
\bauthor{\bsnm{Vaucher}, \binits{A.C.}},
\bauthor{\bsnm{Byekwaso}, \binits{A.}},
\bauthor{\bsnm{Schwaller}, \binits{P.}},
\bauthor{\bsnm{Toniato}, \binits{A.}},
\bauthor{\bsnm{Laino}, \binits{T.}}:
\batitle{Unbiasing {{Retrosynthesis Language Models}} with {{Disconnection
  Prompts}}}.
\bjtitle{ACS Central Science}
\bvolume{9}(\bissue{7}),
\bfpage{1488}--\blpage{1498}
(\byear{2023})
\doiurl{10.1021/acscentsci.3c00372}
\end{barticle}
\endbibitem

\bibitem[\protect\citeauthoryear{Kreutter and
  Reymond}{2023}]{kreutterMultistepRetrosynthesisCombining2023}
\begin{barticle}
\bauthor{\bsnm{Kreutter}, \binits{D.}},
\bauthor{\bsnm{Reymond}, \binits{J.-L.}}:
\batitle{Multistep retrosynthesis combining a disconnection aware triple
  transformer loop with a route penalty score guided tree search}.
\bjtitle{Chemical Science}
\bvolume{14}(\bissue{36}),
\bfpage{9959}--\blpage{9969}
(\byear{2023})
\doiurl{10.1039/D3SC01604H}
\end{barticle}
\endbibitem

\bibitem[\protect\citeauthoryear{Westerlund
  et~al.}{2025}]{westerlundHumanguidedSynthesisPlanning2025a}
\begin{barticle}
\bauthor{\bsnm{Westerlund}, \binits{A.M.}},
\bauthor{\bsnm{Saigiridharan}, \binits{L.}},
\bauthor{\bsnm{Genheden}, \binits{S.}}:
\batitle{Human-guided synthesis planning {\emph{via}} prompting}.
\bjtitle{Chemical Science}
\bvolume{16}(\bissue{32}),
\bfpage{14655}--\blpage{14667}
(\byear{2025})
\doiurl{10.1039/D5SC00927H}
\end{barticle}
\endbibitem

\bibitem[\protect\citeauthoryear{Yang
  et~al.}{2025}]{yangQwen3TechnicalReport2025}
\begin{botherref}
\oauthor{\bsnm{Yang}, \binits{A.}},
\oauthor{\bsnm{Li}, \binits{A.}},
\oauthor{\bsnm{Yang}, \binits{B.}},
\oauthor{\bsnm{Zhang}, \binits{B.}},
\oauthor{\bsnm{Hui}, \binits{B.}},
\oauthor{\bsnm{Zheng}, \binits{B.}},
\oauthor{\bsnm{Yu}, \binits{B.}},
\oauthor{\bsnm{Gao}, \binits{C.}},
\oauthor{\bsnm{Huang}, \binits{C.}},
\oauthor{\bsnm{Lv}, \binits{C.}},
\oauthor{\bsnm{Zheng}, \binits{C.}},
\oauthor{\bsnm{Liu}, \binits{D.}},
\oauthor{\bsnm{Zhou}, \binits{F.}},
\oauthor{\bsnm{Huang}, \binits{F.}},
\oauthor{\bsnm{Hu}, \binits{F.}},
\oauthor{\bsnm{Ge}, \binits{H.}},
\oauthor{\bsnm{Wei}, \binits{H.}},
\oauthor{\bsnm{Lin}, \binits{H.}},
\oauthor{\bsnm{Tang}, \binits{J.}},
\oauthor{\bsnm{Yang}, \binits{J.}},
\oauthor{\bsnm{Tu}, \binits{J.}},
\oauthor{\bsnm{Zhang}, \binits{J.}},
\oauthor{\bsnm{Yang}, \binits{J.}},
\oauthor{\bsnm{Yang}, \binits{J.}},
\oauthor{\bsnm{Zhou}, \binits{J.}},
\oauthor{\bsnm{Zhou}, \binits{J.}},
\oauthor{\bsnm{Lin}, \binits{J.}},
\oauthor{\bsnm{Dang}, \binits{K.}},
\oauthor{\bsnm{Bao}, \binits{K.}},
\oauthor{\bsnm{Yang}, \binits{K.}},
\oauthor{\bsnm{Yu}, \binits{L.}},
\oauthor{\bsnm{Deng}, \binits{L.}},
\oauthor{\bsnm{Li}, \binits{M.}},
\oauthor{\bsnm{Xue}, \binits{M.}},
\oauthor{\bsnm{Li}, \binits{M.}},
\oauthor{\bsnm{Zhang}, \binits{P.}},
\oauthor{\bsnm{Wang}, \binits{P.}},
\oauthor{\bsnm{Zhu}, \binits{Q.}},
\oauthor{\bsnm{Men}, \binits{R.}},
\oauthor{\bsnm{Gao}, \binits{R.}},
\oauthor{\bsnm{Liu}, \binits{S.}},
\oauthor{\bsnm{Luo}, \binits{S.}},
\oauthor{\bsnm{Li}, \binits{T.}},
\oauthor{\bsnm{Tang}, \binits{T.}},
\oauthor{\bsnm{Yin}, \binits{W.}},
\oauthor{\bsnm{Ren}, \binits{X.}},
\oauthor{\bsnm{Wang}, \binits{X.}},
\oauthor{\bsnm{Zhang}, \binits{X.}},
\oauthor{\bsnm{Ren}, \binits{X.}},
\oauthor{\bsnm{Fan}, \binits{Y.}},
\oauthor{\bsnm{Su}, \binits{Y.}},
\oauthor{\bsnm{Zhang}, \binits{Y.}},
\oauthor{\bsnm{Zhang}, \binits{Y.}},
\oauthor{\bsnm{Wan}, \binits{Y.}},
\oauthor{\bsnm{Liu}, \binits{Y.}},
\oauthor{\bsnm{Wang}, \binits{Z.}},
\oauthor{\bsnm{Cui}, \binits{Z.}},
\oauthor{\bsnm{Zhang}, \binits{Z.}},
\oauthor{\bsnm{Zhou}, \binits{Z.}},
\oauthor{\bsnm{Qiu}, \binits{Z.}}:
Qwen3 {{Technical Report}}.
arXiv
(2025).
\doiurl{10.48550/arXiv.2505.09388}
\end{botherref}
\endbibitem

\bibitem[\protect\citeauthoryear{Comanici
  et~al.}{2025}]{comaniciGemini25Pushing2025}
\begin{botherref}
\oauthor{\bsnm{Comanici}, \binits{G.}},
\oauthor{\bsnm{Bieber}, \binits{E.}},
\oauthor{\bsnm{Schaekermann}, \binits{M.}},
\oauthor{\bsnm{Pasupat}, \binits{I.}},
\oauthor{\bsnm{Sachdeva}, \binits{N.}},
\oauthor{\bsnm{Dhillon}, \binits{I.}},
\oauthor{\bsnm{Blistein}, \binits{M.}},
\oauthor{\bsnm{Ram}, \binits{O.}},
\oauthor{\bsnm{Zhang}, \binits{D.}},
\oauthor{\bsnm{Rosen}, \binits{E.}}, et al.:
Gemini 2.5: Pushing the frontier with advanced reasoning, multimodality, long
  context, and next generation agentic capabilities.
arXiv preprint arXiv:2507.06261
(2025)
\end{botherref}
\endbibitem

\bibitem[\protect\citeauthoryear{{Anthropic}}{2025}]{anthropicClaude4System2025}
\begin{botherref}
\oauthor{\bsnm{{Anthropic}}}:
Claude 4 {{System Card}}.
https://www.anthropic.com/claude-4-system-card
(2025)
\end{botherref}
\endbibitem

\bibitem[\protect\citeauthoryear{{OpenAI}}{2025}]{openaiGPT5SystemCard2025}
\begin{botherref}
\oauthor{\bsnm{{OpenAI}}}:
{{GPT-5 System Card}}.
https://cdn.openai.com/gpt-5-system-card.pdf
(2025)
\end{botherref}
\endbibitem

\bibitem[\protect\citeauthoryear{Lowe}{2012}]{loweExtractionChemicalStructures2012}
\begin{botherref}
\oauthor{\bsnm{Lowe}, \binits{D.M.}}:
Extraction of chemical structures and reactions from the literature.
Thesis,
University of Cambridge
(2012)
\end{botherref}
\endbibitem

\bibitem[\protect\citeauthoryear{Schneider
  et~al.}{2016}]{schneiderWhatsWhatNearly2016}
\begin{barticle}
\bauthor{\bsnm{Schneider}, \binits{N.}},
\bauthor{\bsnm{Stiefl}, \binits{N.}},
\bauthor{\bsnm{Landrum}, \binits{G.A.}}:
\batitle{What's {{What}}: {{The}} ({{Nearly}}) {{Definitive Guide}} to
  {{Reaction Role Assignment}}}.
\bjtitle{Journal of Chemical Information and Modeling}
\bvolume{56}(\bissue{12}),
\bfpage{2336}--\blpage{2346}
(\byear{2016})
\doiurl{10.1021/acs.jcim.6b00564}
\end{barticle}
\endbibitem

\bibitem[\protect\citeauthoryear{Genheden and
  Bjerrum}{2022}]{genhedenPaRoutesFrameworkBenchmarking2022}
\begin{barticle}
\bauthor{\bsnm{Genheden}, \binits{S.}},
\bauthor{\bsnm{Bjerrum}, \binits{E.}}:
\batitle{{{PaRoutes}}: Towards a framework for benchmarking retrosynthesis
  route predictions}.
\bjtitle{Digital Discovery}
\bvolume{1}(\bissue{4}),
\bfpage{527}--\blpage{539}
(\byear{2022})
\doiurl{10.1039/D2DD00015F}
\end{barticle}
\endbibitem

\bibitem[\protect\citeauthoryear{Somnath
  et~al.}{2021}]{somnathLearningGraphModels2021}
\begin{bchapter}
\bauthor{\bsnm{Somnath}, \binits{V.R.}},
\bauthor{\bsnm{Bunne}, \binits{C.}},
\bauthor{\bsnm{Coley}, \binits{C.}},
\bauthor{\bsnm{Krause}, \binits{A.}},
\bauthor{\bsnm{Barzilay}, \binits{R.}}:
\bctitle{Learning {{Graph Models}} for {{Retrosynthesis Prediction}}}.
In: \bbtitle{Advances in {{Neural Information Processing Systems}}},
vol. \bseriesno{34},
pp. \bfpage{9405}--\blpage{9415}.
\bpublisher{Curran Associates, Inc.}, \blocation{???}
(\byear{2021})
\end{bchapter}
\endbibitem

\bibitem[\protect\citeauthoryear{Dobbelaere
  et~al.}{2024}]{dobbelaereRxnINSIGHTFastChemical2024c}
\begin{barticle}
\bauthor{\bsnm{Dobbelaere}, \binits{M.R.}},
\bauthor{\bsnm{Lengyel}, \binits{I.}},
\bauthor{\bsnm{Stevens}, \binits{C.V.}},
\bauthor{\bsnm{Van~Geem}, \binits{K.M.}}:
\batitle{Rxn-{{INSIGHT}}: Fast chemical reaction analysis using bond-electron
  matrices}.
\bjtitle{Journal of Cheminformatics}
\bvolume{16}(\bissue{1}),
\bfpage{37}
(\byear{2024})
\doiurl{10.1186/s13321-024-00834-z}
\end{barticle}
\endbibitem

\bibitem[\protect\citeauthoryear{Deng et~al.}{2017}]{dengTriazoleUreasAct2017}
\begin{barticle}
\bauthor{\bsnm{Deng}, \binits{H.}},
\bauthor{\bsnm{Kooijman}, \binits{S.}},
\bauthor{\bsnm{Van Den~Nieuwendijk}, \binits{A.M.C.H.}},
\bauthor{\bsnm{Ogasawara}, \binits{D.}},
\bauthor{\bsnm{Van Der~Wel}, \binits{T.}},
\bauthor{\bsnm{Van~Dalen}, \binits{F.}},
\bauthor{\bsnm{Baggelaar}, \binits{M.P.}},
\bauthor{\bsnm{Janssen}, \binits{F.J.}},
\bauthor{\bsnm{Van Den~Berg}, \binits{R.J.B.H.N.}},
\bauthor{\bsnm{Den~Dulk}, \binits{H.}},
\bauthor{\bsnm{Cravatt}, \binits{B.F.}},
\bauthor{\bsnm{Overkleeft}, \binits{H.S.}},
\bauthor{\bsnm{Rensen}, \binits{P.C.N.}},
\bauthor{\bsnm{Van Der~Stelt}, \binits{M.}}:
\batitle{Triazole {{Ureas Act}} as {{Diacylglycerol Lipase Inhibitors}} and
  {{Prevent Fasting-Induced Refeeding}}}.
\bjtitle{Journal of Medicinal Chemistry}
\bvolume{60}(\bissue{1}),
\bfpage{428}--\blpage{440}
(\byear{2017})
\doiurl{10.1021/acs.jmedchem.6b01482}
\end{barticle}
\endbibitem

\bibitem[\protect\citeauthoryear{Li
  et~al.}{2023}]{liStructuralBasisSelective2023}
\begin{barticle}
\bauthor{\bsnm{Li}, \binits{X.}},
\bauthor{\bsnm{Chang}, \binits{H.}},
\bauthor{\bsnm{Bouma}, \binits{J.}},
\bauthor{\bsnm{De~Paus}, \binits{L.V.}},
\bauthor{\bsnm{Mukhopadhyay}, \binits{P.}},
\bauthor{\bsnm{Paloczi}, \binits{J.}},
\bauthor{\bsnm{Mustafa}, \binits{M.}},
\bauthor{\bsnm{Van Der~Horst}, \binits{C.}},
\bauthor{\bsnm{Kumar}, \binits{S.S.}},
\bauthor{\bsnm{Wu}, \binits{L.}},
\bauthor{\bsnm{Yu}, \binits{Y.}},
\bauthor{\bsnm{Van Den~Berg}, \binits{R.J.B.H.N.}},
\bauthor{\bsnm{Janssen}, \binits{A.P.A.}},
\bauthor{\bsnm{Lichtman}, \binits{A.}},
\bauthor{\bsnm{Liu}, \binits{Z.-J.}},
\bauthor{\bsnm{Pacher}, \binits{P.}},
\bauthor{\bsnm{Van Der~Stelt}, \binits{M.}},
\bauthor{\bsnm{Heitman}, \binits{L.H.}},
\bauthor{\bsnm{Hua}, \binits{T.}}:
\batitle{Structural basis of selective cannabinoid {{CB2}} receptor
  activation}.
\bjtitle{Nature Communications}
\bvolume{14}(\bissue{1}),
\bfpage{1447}
(\byear{2023})
\doiurl{10.1038/s41467-023-37112-9}
\end{barticle}
\endbibitem

\bibitem[\protect\citeauthoryear{Baggelaar
  et~al.}{2015}]{baggelaarHighlySelectiveReversible2015}
\begin{barticle}
\bauthor{\bsnm{Baggelaar}, \binits{M.P.}},
\bauthor{\bsnm{Chameau}, \binits{P.J.P.}},
\bauthor{\bsnm{Kantae}, \binits{V.}},
\bauthor{\bsnm{Hummel}, \binits{J.}},
\bauthor{\bsnm{Hsu}, \binits{K.-L.}},
\bauthor{\bsnm{Janssen}, \binits{F.}},
\bauthor{\bsnm{Van Der~Wel}, \binits{T.}},
\bauthor{\bsnm{Soethoudt}, \binits{M.}},
\bauthor{\bsnm{Deng}, \binits{H.}},
\bauthor{\bsnm{Den~Dulk}, \binits{H.}},
\bauthor{\bsnm{Allar{\`a}}, \binits{M.}},
\bauthor{\bsnm{Florea}, \binits{B.I.}},
\bauthor{\bsnm{Di~Marzo}, \binits{V.}},
\bauthor{\bsnm{Wadman}, \binits{W.J.}},
\bauthor{\bsnm{Kruse}, \binits{C.G.}},
\bauthor{\bsnm{Overkleeft}, \binits{H.S.}},
\bauthor{\bsnm{Hankemeier}, \binits{T.}},
\bauthor{\bsnm{Werkman}, \binits{T.R.}},
\bauthor{\bsnm{Cravatt}, \binits{B.F.}},
\bauthor{\bsnm{Van Der~Stelt}, \binits{M.}}:
\batitle{Highly {{Selective}}, {{Reversible Inhibitor Identified}} by
  {{Comparative Chemoproteomics Modulates Diacylglycerol Lipase Activity}} in
  {{Neurons}}}.
\bjtitle{Journal of the American Chemical Society}
\bvolume{137}(\bissue{27}),
\bfpage{8851}--\blpage{8857}
(\byear{2015})
\doiurl{10.1021/jacs.5b04883}
\end{barticle}
\endbibitem

\bibitem[\protect\citeauthoryear{Mock
  et~al.}{2020}]{mockDiscoveryNAPEPLDInhibitor2020}
\begin{barticle}
\bauthor{\bsnm{Mock}, \binits{E.D.}},
\bauthor{\bsnm{Mustafa}, \binits{M.}},
\bauthor{\bsnm{{Gunduz-Cinar}}, \binits{O.}},
\bauthor{\bsnm{Cinar}, \binits{R.}},
\bauthor{\bsnm{Petrie}, \binits{G.N.}},
\bauthor{\bsnm{Kantae}, \binits{V.}},
\bauthor{\bsnm{Di}, \binits{X.}},
\bauthor{\bsnm{Ogasawara}, \binits{D.}},
\bauthor{\bsnm{Varga}, \binits{Z.V.}},
\bauthor{\bsnm{Paloczi}, \binits{J.}},
\bauthor{\bsnm{Miliano}, \binits{C.}},
\bauthor{\bsnm{Donvito}, \binits{G.}},
\bauthor{\bsnm{Van~Esbroeck}, \binits{A.C.M.}},
\bauthor{\bsnm{Van Der~Gracht}, \binits{A.M.F.}},
\bauthor{\bsnm{Kotsogianni}, \binits{I.}},
\bauthor{\bsnm{Park}, \binits{J.K.}},
\bauthor{\bsnm{Martella}, \binits{A.}},
\bauthor{\bsnm{Van Der~Wel}, \binits{T.}},
\bauthor{\bsnm{Soethoudt}, \binits{M.}},
\bauthor{\bsnm{Jiang}, \binits{M.}},
\bauthor{\bsnm{Wendel}, \binits{T.J.}},
\bauthor{\bsnm{Janssen}, \binits{A.P.A.}},
\bauthor{\bsnm{Bakker}, \binits{A.T.}},
\bauthor{\bsnm{Donovan}, \binits{C.M.}},
\bauthor{\bsnm{Castillo}, \binits{L.I.}},
\bauthor{\bsnm{Florea}, \binits{B.I.}},
\bauthor{\bsnm{Wat}, \binits{J.}},
\bauthor{\bsnm{Van Den~Hurk}, \binits{H.}},
\bauthor{\bsnm{Wittwer}, \binits{M.}},
\bauthor{\bsnm{Grether}, \binits{U.}},
\bauthor{\bsnm{Holmes}, \binits{A.}},
\bauthor{\bsnm{Van~Boeckel}, \binits{C.A.A.}},
\bauthor{\bsnm{Hankemeier}, \binits{T.}},
\bauthor{\bsnm{Cravatt}, \binits{B.F.}},
\bauthor{\bsnm{Buczynski}, \binits{M.W.}},
\bauthor{\bsnm{Hill}, \binits{M.N.}},
\bauthor{\bsnm{Pacher}, \binits{P.}},
\bauthor{\bsnm{Lichtman}, \binits{A.H.}},
\bauthor{\bsnm{Van Der~Stelt}, \binits{M.}}:
\batitle{Discovery of a {{NAPE-PLD}} inhibitor that modulates emotional
  behavior in mice}.
\bjtitle{Nature Chemical Biology}
\bvolume{16}(\bissue{6}),
\bfpage{667}--\blpage{675}
(\byear{2020})
\doiurl{10.1038/s41589-020-0528-7}
\end{barticle}
\endbibitem

\bibitem[\protect\citeauthoryear{Jiang
  et~al.}{2023}]{jiangMonoacylglycerolLipaseInhibitor2023}
\begin{barticle}
\bauthor{\bsnm{Jiang}, \binits{M.}},
\bauthor{\bsnm{Huizenga}, \binits{M.C.W.}},
\bauthor{\bsnm{Wirt}, \binits{J.L.}},
\bauthor{\bsnm{Paloczi}, \binits{J.}},
\bauthor{\bsnm{Amedi}, \binits{A.}},
\bauthor{\bsnm{Van Den~Berg}, \binits{R.J.B.H.N.}},
\bauthor{\bsnm{Benz}, \binits{J.}},
\bauthor{\bsnm{Collin}, \binits{L.}},
\bauthor{\bsnm{Deng}, \binits{H.}},
\bauthor{\bsnm{Di}, \binits{X.}},
\bauthor{\bsnm{Driever}, \binits{W.F.}},
\bauthor{\bsnm{Florea}, \binits{B.I.}},
\bauthor{\bsnm{Grether}, \binits{U.}},
\bauthor{\bsnm{Janssen}, \binits{A.P.A.}},
\bauthor{\bsnm{Hankemeier}, \binits{T.}},
\bauthor{\bsnm{Heitman}, \binits{L.H.}},
\bauthor{\bsnm{Lam}, \binits{T.-W.}},
\bauthor{\bsnm{Mohr}, \binits{F.}},
\bauthor{\bsnm{Pavlovic}, \binits{A.}},
\bauthor{\bsnm{Ruf}, \binits{I.}},
\bauthor{\bsnm{Van Den~Hurk}, \binits{H.}},
\bauthor{\bsnm{Stevens}, \binits{A.F.}},
\bauthor{\bsnm{Van Der~Vliet}, \binits{D.}},
\bauthor{\bsnm{Van Der~Wel}, \binits{T.}},
\bauthor{\bsnm{Wittwer}, \binits{M.B.}},
\bauthor{\bsnm{Van~Boeckel}, \binits{C.A.A.}},
\bauthor{\bsnm{Pacher}, \binits{P.}},
\bauthor{\bsnm{Hohmann}, \binits{A.G.}},
\bauthor{\bsnm{Van Der~Stelt}, \binits{M.}}:
\batitle{A monoacylglycerol lipase inhibitor showing therapeutic efficacy in
  mice without central side effects or dependence}.
\bjtitle{Nature Communications}
\bvolume{14}(\bissue{1}),
\bfpage{8039}
(\byear{2023})
\doiurl{10.1038/s41467-023-43606-3}
\end{barticle}
\endbibitem

\bibitem[\protect\citeauthoryear{Zhu
  et~al.}{2024}]{zhuActivatingP53Y220CMutantSpecific2024}
\begin{barticle}
\bauthor{\bsnm{Zhu}, \binits{X.}},
\bauthor{\bsnm{Byun}, \binits{W.S.}},
\bauthor{\bsnm{Pie{\'n}kowska}, \binits{D.E.}},
\bauthor{\bsnm{Nguyen}, \binits{K.T.}},
\bauthor{\bsnm{Gerhartz}, \binits{J.}},
\bauthor{\bsnm{Geng}, \binits{Q.}},
\bauthor{\bsnm{Qiu}, \binits{T.}},
\bauthor{\bsnm{Zhong}, \binits{J.}},
\bauthor{\bsnm{Jiang}, \binits{Z.}},
\bauthor{\bsnm{Wang}, \binits{M.}},
\bauthor{\bsnm{Sarott}, \binits{R.C.}},
\bauthor{\bsnm{Hinshaw}, \binits{S.M.}},
\bauthor{\bsnm{Zhang}, \binits{T.}},
\bauthor{\bsnm{Attardi}, \binits{L.D.}},
\bauthor{\bsnm{Nowak}, \binits{R.P.}},
\bauthor{\bsnm{Gray}, \binits{N.S.}}:
\batitle{Activating p53y220c with a mutant-specific small molecule}.
\bjtitle{bioRxiv}
(\byear{2024})
\doiurl{10.1101/2024.10.23.619961}
{\href{https://arxiv.org/abs/https://www.biorxiv.org/content/early/2024/10/28/2024.10.23.619961.full.pdf}{{https://www.biorxiv.org/content/early/2024/10/28/2024.10.23.619961.full.pdf}}}
\end{barticle}
\endbibitem

\bibitem[\protect\citeauthoryear{Tarr
  et~al.}{2025}]{tarrDiscoveryMacrocyclicMyeloid2025}
\begin{barticle}
\bauthor{\bsnm{Tarr}, \binits{J.C.}},
\bauthor{\bsnm{Jeon}, \binits{K.}},
\bauthor{\bsnm{Veerasamy}, \binits{N.}},
\bauthor{\bsnm{Aichinger}, \binits{M.}},
\bauthor{\bsnm{Salovich}, \binits{J.M.}},
\bauthor{\bsnm{Zhao}, \binits{B.}},
\bauthor{\bsnm{Sensintaffar}, \binits{J.L.}},
\bauthor{\bsnm{Arnhof}, \binits{H.}},
\bauthor{\bsnm{Wunberg}, \binits{T.}},
\bauthor{\bsnm{Sgubin}, \binits{D.}},
\bauthor{\bsnm{Arnold}, \binits{A.}},
\bauthor{\bsnm{Vekariya}, \binits{R.H.}},
\bauthor{\bsnm{Christov}, \binits{P.P.}},
\bauthor{\bsnm{Kim}, \binits{K.}},
\bauthor{\bsnm{Fuchs}, \binits{J.E.}},
\bauthor{\bsnm{Karier}, \binits{P.}},
\bauthor{\bsnm{Betzemeier}, \binits{B.}},
\bauthor{\bsnm{Van~Meveren}, \binits{M.}},
\bauthor{\bsnm{Miriyala}, \binits{N.}},
\bauthor{\bsnm{Olejniczak}, \binits{E.T.}},
\bauthor{\bsnm{Engelhardt}, \binits{H.}},
\bauthor{\bsnm{Lee}, \binits{T.}},
\bauthor{\bsnm{McConnell}, \binits{D.}},
\bauthor{\bsnm{Fesik}, \binits{S.W.}}:
\batitle{Discovery of macrocyclic myeloid cell leukemia 1 (mcl-1) inhibitors
  that demonstrate potent cellular efficacy and in vivo activity in a mouse
  solid tumor xenograft model}.
\bjtitle{Journal of Medicinal Chemistry}
\bvolume{68}(\bissue{17}),
\bfpage{18553}--\blpage{18578}
(\byear{2025})
\doiurl{10.1021/acs.jmedchem.5c01376}
{\href{https://arxiv.org/abs/https://doi.org/10.1021/acs.jmedchem.5c01376}{{https://doi.org/10.1021/acs.jmedchem.5c01376}}}.
\bcomment{PMID: 40864607}
\end{barticle}
\endbibitem

\bibitem[\protect\citeauthoryear{Hassen
  et~al.}{2025}]{hassenSynthesisPlanningReaction2025}
\begin{botherref}
\oauthor{\bsnm{Hassen}, \binits{A.K.}},
\oauthor{\bsnm{Lai}, \binits{H.}},
\oauthor{\bsnm{Genheden}, \binits{S.}},
\oauthor{\bsnm{Preuss}, \binits{M.}},
\oauthor{\bsnm{Clevert}, \binits{D.-A.}}:
Synthesis {{Planning}} in {{Reaction Space}}: {{A Study}} on {{Success}},
  {{Robustness}} and {{Diversity}}.
Chemistry
(2025).
\doiurl{10.26434/chemrxiv-2025-js7dt}
\end{botherref}
\endbibitem

\bibitem[\protect\citeauthoryear{Dombrowski
  et~al.}{2022}]{dombrowskiChosenFewParallel2022}
\begin{barticle}
\bauthor{\bsnm{Dombrowski}, \binits{A.W.}},
\bauthor{\bsnm{Aguirre}, \binits{A.L.}},
\bauthor{\bsnm{Shrestha}, \binits{A.}},
\bauthor{\bsnm{Sarris}, \binits{K.A.}},
\bauthor{\bsnm{Wang}, \binits{Y.}}:
\batitle{The {{Chosen Few}}: {{Parallel Library Reaction Methodologies}} for
  {{Drug Discovery}}}.
\bjtitle{The Journal of Organic Chemistry}
\bvolume{87}(\bissue{4}),
\bfpage{1880}--\blpage{1897}
(\byear{2022})
\doiurl{10.1021/acs.joc.1c01427}
\end{barticle}
\endbibitem

\bibitem[\protect\citeauthoryear{Saigiridharan
  et~al.}{2024}]{saigiridharanAiZynthFinder40Developments2024}
\begin{barticle}
\bauthor{\bsnm{Saigiridharan}, \binits{L.}},
\bauthor{\bsnm{Hassen}, \binits{A.K.}},
\bauthor{\bsnm{Lai}, \binits{H.}},
\bauthor{\bsnm{{Torren-Peraire}}, \binits{P.}},
\bauthor{\bsnm{Engkvist}, \binits{O.}},
\bauthor{\bsnm{Genheden}, \binits{S.}}:
\batitle{{{AiZynthFinder}} 4.0: Developments based on learnings from 3 years of
  industrial application}.
\bjtitle{Journal of Cheminformatics}
\bvolume{16}(\bissue{1}),
\bfpage{57}
(\byear{2024})
\doiurl{10.1186/s13321-024-00860-x}
\end{barticle}
\endbibitem

\end{thebibliography}

\end{document}